\documentclass{article}
\usepackage[preprint]{neurips_2026}

\usepackage[utf8]{inputenc}
\usepackage[T1]{fontenc}
\usepackage{hyperref}
\usepackage{url}
\usepackage{booktabs}
\usepackage{amsfonts}
\usepackage{amsmath}
\usepackage{amssymb}
\usepackage{nicefrac}
\usepackage{microtype}
\usepackage{xcolor}
\usepackage{graphicx}
\graphicspath{{figures/}}
\usepackage{subcaption}
\usepackage{multirow}
\usepackage{wrapfig}
\usepackage{algorithm}
\usepackage{algorithmic}
\usepackage{makecell}
\usepackage{relsize}
\usepackage{bm}
\usepackage{enumitem}
\usepackage{arydshln}
\usepackage{colortbl}
\usepackage{tikz}
\usetikzlibrary{positioning, decorations.pathreplacing, calc}

\usepackage{amsmath,amssymb}
\usepackage{bm}
 
\newcommand{\bh}{\bm{h}}
\newcommand{\bp}{\bm{p}}
\newcommand{\bg}{\bm{g}}
\newcommand{\be}{\bm{e}}
\newcommand{\bq}{\bm{q}}
\newcommand{\bk}{\bm{k}}
\newcommand{\bv}{\bm{v}}
\newcommand{\ba}{\bm{a}}
\newcommand{\bo}{\bm{o}}
\newcommand{\bw}{\bm{w}}
\newcommand{\bz}{\bm{z}}
\newcommand{\bR}{\bm{R}}
\newcommand{\bH}{\bm{H}}
\newcommand{\bA}{\bm{A}}
\newcommand{\bT}{\bm{T}}
\DeclareMathOperator{\softmax}{softmax}

\newcommand{\br}{\bm{r}}

\newcommand{\bmu}{\bm{\mu}}
\newcommand{\bsigma}{\bm{\sigma}}
\newcommand{\bepsilon}{\bm{\epsilon}}
\newcommand{\bzero}{\bm{0}}
\newcommand{\bI}{\bm{I}}

\newcommand{\method}{TopVAE}

\newcommand{\E}{\mathbb{E}}
\newcommand{\loss}{\mathcal{L}}
\newcommand{\dataset}{\mathcal{D}}

\DeclareMathOperator{\diag}{diag}

\newcommand{\ourmark}{\textsuperscript{$\dagger$}}

\newcommand{\placeholder}[1]{{\color{gray}\texttt{--}}}
\newcommand{\na}{\textcolor{gray}{--}}

\definecolor{rowhl}{gray}{0.93}

\title{Smoothing Dark Areas in Molecular Latent Diffusion}

\author{
Xi Wang$^{1}$ \quad
Jiahan Li$^{1}$ \quad
Yuxuan Xia$^{1}$ \quad
Yingcheng Wu$^{2}$ \quad
Shaoyi Zheng$^{1}$ \quad
Shengjie Wang$^{1}$ \\
$^{1}$New York University \\
$^{2}$Stanford University
}

\begin{document}
\maketitle

\begin{abstract}
Latent diffusion is a promising framework for scalable 3D molecular generation, 
but it requires a latent space that remains smooth, valid, and navigable beyond 
posterior samples. Existing molecular VAEs, however, are typically learned through 
reconstruction-based objectives, which do not guarantee such a latent space. 
We show that this leads to \textbf{dark areas}: regions of latent space that are reachable 
during diffusion sampling but decode to disconnected or chemically invalid molecules. 
Unlike in image generation, molecular decoding requires strict structural and chemical 
precision, so even small latent perturbations can produce catastrophic failures. 
We therefore propose \textbf{\method{}}, a topology-optimized VAE that reduces dark areas 
by making the decoder internalize structural and chemical constraints during training, 
eliminating the need for test-time chemical correction. \method{} greatly improves 
off-posterior robustness, and when paired with a standard DiT, achieves $77\%$ lower 
FCD\textsubscript{3D} on QM9, the highest V\&C, $52\%$ lower FCD\textsubscript{3D} on 
GEOM-Drugs, and $1.29{\times}$ more stable and connected molecules on zero-shot scaffold inpainting.
\end{abstract}

\section{Introduction}
\label{sec:intro}

Latent diffusion models, originally developed for image generation~\citep{rombach2022high,peebles2023scalable}, have also been extended to 3D molecular generation~\citep{xu2023geometric,chen2025manipulating}. 
This setting is more challenging because valid molecules must simultaneously satisfy structural, geometric, and chemical constraints~\citep{you2024latent,luo2025towards}. 
A latent-variable formulation is therefore especially appealing: if these complexities can be absorbed by the decoder, diffusion can operate in a cleaner latent space, simplifying modeling and sampling.

\begin{figure}[t]
\centering
\includegraphics[width=\linewidth]{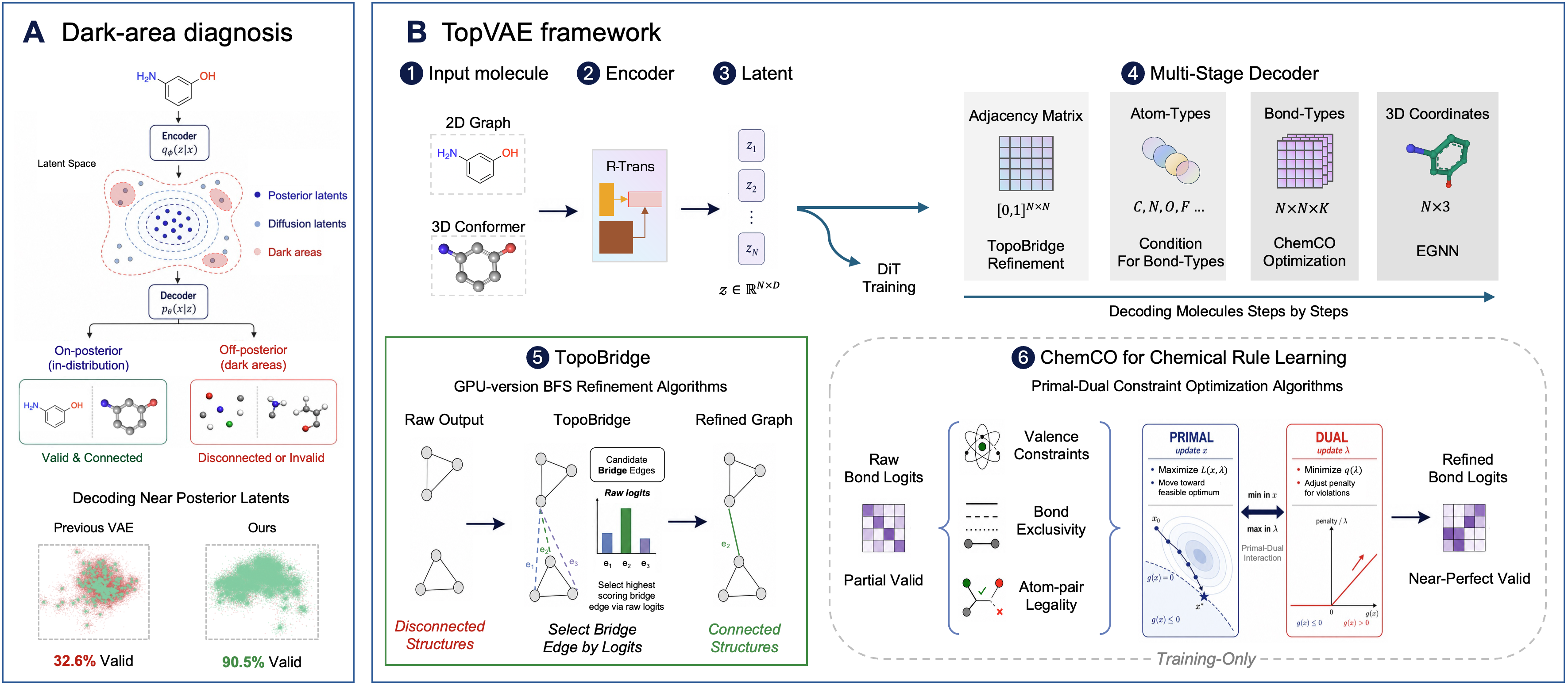}
\caption{\textbf{Overview of TopVAE.} (A) Dark-area diagnosis reveals that existing VAE decoders produce invalid molecules in off-posterior latent regions, which TopVAE closes by internalizing chemical rules. (B) TopVAE combines a topology-first multi-stage decoder with TopoBridge for guaranteed connectivity and ChemCO for chemical constraint learning during training.}
\label{fig:main-fig}
\end{figure}

Molecular latent spaces are typically learned with reconstruction objectives plus regularization, 
as in VAE or VQ-VAE frameworks~\citep{luo2025towards}. 
However, strong reconstruction does not guarantee a latent space suitable for diffusion. 
Recent work shows that reconstruction quality can poorly predict, and sometimes even conflict 
with, downstream generation performance~\citep{yao2025reconstruction,skorokhodov2025improving,xu2026making}, 
because diffusion requires a smooth, interpolatable latent manifold rather than correctness 
only at posterior samples. 
Although latent regularization is intended to enlarge the region of valid latents, this is 
particularly difficult for molecules: molecular decoding demands strict structural and chemical 
precision, and unlike images, it lacks strong architectural biases that make nearby latent 
points decode naturally. As a result, even small perturbations around posterior latents 
can produce severely corrupted molecules, including disconnected structures or graphs that 
fail chemical sanitization (Fig.~\ref{fig:main-fig}). 
We call these failure regions \textbf{dark areas}: latent neighborhoods where the decoder 
breaks, rendering the surrounding manifold non-interpolatable and difficult to navigate for diffusion.

Dark areas arise mainly from two sources: \emph{topological disconnection}, where the decoded graph fractures into isolated components, and \emph{chemical invalidity}, where predicted bonds violate valence or compatibility rules. 
Prior work often addresses these issues by enforcing chemical rules during generation~\citep{jin2018junction,liu2018constrained,ma2018constrained,krenn2022selfies}. 
While such constraints can repair outputs, they do not remove the underlying latent-space fragility and may also bias generation. 
Instead, we aim to internalize these constraints during VAE training so that valid molecules occupy a more robust latent neighborhood, without requiring chemical constraint optimization (ChemCO) at inference time. 
To this end, we propose \textbf{TopVAE} (\textbf{T}opology-\textbf{OP}timized \textbf{VAE}), which promotes connectivity through BFS-based adjacency refinement, enforces valence and bond-type constraints via unrolled primal--dual optimization during training, and selectively injects ChemCO's corrections into the decoder~\citep{self-imitation}, enabling ChemCO-free inference.

In summary, our contributions are:
\begin{enumerate}[leftmargin=1.8em,itemsep=2pt,topsep=2pt]

\item \textbf{Dark areas in molecular latent space.}
We identify and formalize dark areas: latent regions reachable by diffusion sampling but decoded as chemically invalid or disconnected molecules.

\item \textbf{TopVAE} with three components:
a) TopoBridge guarantees connectivity via adjacency refinement;
b) inherent ChemCO provides chemical constraints through unrolled primal--dual optimization; 
c) Advantage-Gated Constratint Learning (AGCL) that selectively feeds constraint-based correction signals into the decoder training.

\item \textbf{State-of-the-art generation with dark-area closure.}
TopVAE paired with a standard DiT achieves $77\%$ lower FCD\textsubscript{3D} on QM9, 
the highest V\&C and $52\%$ lower FCD\textsubscript{3D} on GEOM-Drugs, 
and $1.29{\times}$ more stable and connected molecules on zero-shot scaffold inpainting.

\end{enumerate}

\section{Related Work}
\label{sec:related}

\paragraph{3D molecular generation.}
Diffusion and flow models for de novo 3D molecules fall into two families.
\emph{Data-space} models generate directly in molecular data space: some diffuse atom types and coordinates and infer bonds post hoc~\citep{hoogeboom2022equivariant}, while later graph-aware or complete-molecule variants reduce atom--bond mismatch
by jointly modeling graph/topology and geometry, by explicitly predicting
bond variables, or by introducing bond-formation-aware training objectives
~\citep{peng2023moldiff,huang2023learning,vignac2023midi,le2023navigating,Megalodon,GFMDiff}.
\emph{Latent} models first compress molecules into continuous representations and then fit a diffusion model~\citep{xu2023geometric,you2024latent,luo2025towards,joshi2025all}, gaining scalability and controllability but requiring high quality latents for diffusion models.
Our dark-area diagnostic makes this assumption explicit and testable, motivating a decoder-centric complement to existing latent molecular diffusion models.

\paragraph{Constraint-aware molecular decoding.}
Chemical and structural validity has been enforced via
grammar-level constraints~\citep{kusner2017grammar,krenn2022selfies},
structured graph decoders~\citep{jin2018junction,liu2018constrained},
validity-oriented regularization~\citep{ma2018constrained},
and differentiable constraint-satisfaction layers~\citep{wang2023linsatnet,zeng2024glinsat}.
TopVAE takes a different route: rather than retaining a permanent constraint layer
at inference, we use constraint-guided corrections from ChemCO as a selective training signal using AGCL,
with the goal that the unconstrained decoder itself internalizes chemical rules
and supports ChemCO-free inference
(Sec.~\ref{sec:agcl}, Fig.~\ref{fig:chemco-agcl-effect}).
Extended related work is in Appendix~\ref{app:extended_related}.

\section{Prerequisites}
\label{sec:prelim}

\paragraph{3D molecular VAE.}
A 3D molecule can represented as $M=(\bA,\bT,\bm{B},\bR)$, where $\bA\in\{0,1\}^{N\times N}$ is the binary adjacency matrix, $\bT$ denotes atom types, $\bm{B}$ denotes bond types, and $\bR\in\mathbb{R}^{N\times 3}$ stores atomic coordinates. 
A molecular VAE encodes $M$ into a latent variable $\bz$ and reconstructs the molecule through a decoder $D_\theta$:
\begin{equation}
    q_\phi(\bz\mid M)
    =
    \mathcal{N}\!\left(\bmu_\phi(M),\diag(\bsigma_\phi^2(M))\right),
    \qquad
    \hat M = D_\theta(\bz),
    \quad
    \bz=\bmu_\phi(M)+\bsigma_\phi(M)\odot\bepsilon .
    \label{eq:vae-encoding}
\end{equation}
Here $q_\phi(\bz\mid M)$ is the encoder posterior, parameterized by encoder parameters $\phi$, which maps an input molecule to the mean and variance of a Gaussian latent distribution.
The VAE is trained by balancing reconstruction and latent regularization:
\begin{equation}
    \loss_{\mathrm{VAE}}
    =
    \E_{M\sim\dataset}
    \E_{\bz\sim q_\phi(\bz\mid M)}
    \big[-\log p_\theta(M\mid \bz)\big]
    +
    \beta\,\mathrm{KL}\!\left(q_\phi(\bz\mid M)\,\|\,p(\bz)\right).
    \label{eq:vae-objective}
\end{equation}

\paragraph{Latent diffusion over molecular latents.}
A diffusion model is trained on the VAE latent space to generate new latent codes. Starting from a clean latent $\bz_0$, the forward process adds Gaussian noise:
\begin{equation}
    \bz_t
    =
    \sqrt{\bar\alpha_t}\,\bz_0
    +
    \sqrt{1-\bar\alpha_t}\,\bepsilon,
    \qquad
    \bepsilon\sim\mathcal{N}(\bzero,\bI).
    \label{eq:latent-diffusion}
\end{equation}
A denoiser $\bepsilon_\psi(\bz_t,t)$ is trained to predict $\bepsilon$, typically with an MSE loss. 
At sampling time, iterative denoising produces a clean latent $\hat{\bz}_0$, 
which is decoded into a molecule: $\hat M = D_\theta(\hat{\bz}_0)$.
Therefore, the decoder must work not only on posterior latents, 
but also on noisy and prior-like latents visited during diffusion.




\paragraph{Dark areas in molecular latent space.}
At inference time, a latent diffusion model may sample latents that lie between or outside the posterior codes seen during training.
We call a decoded graph \emph{chemically valid} if it (i)~passes RDKit sanitization---requiring legal valences, consistent aromaticity, and recognized bond types---and (ii)~forms a single connected component.
Denoting the sampling-reachable region by $\mathcal{A}_{\mathrm{LDM}}$ and the valid molecule set by $\mathcal{M}_{\mathrm{valid}}$, we define the \textbf{dark areas} as
\begin{equation}
    \mathcal{D}_{\mathrm{dark}}
    =
    \{\bz\in\mathcal{A}_{\mathrm{LDM}}:
    D_\theta(\bz)\notin\mathcal{M}_{\mathrm{valid}}\}.
\end{equation}

\section{TopVAE: Topology-Optimized VAE for Chemical Constraint Learning}
\label{sec:method}
TopVAE targets dark areas in molecular latent space.
The goal is to make the VAE decoder learn chemical rules, closing dark-areas.
TopVAE therefore combines a structured molecular decoder with two constraint-aware modules.
TopoBridge refines the predicted adjacency into a guaranteed connected adjacency matrix.
ChemCO unrolls chemical constraint optimization and converts valence limits, bond exclusivity, atom-pair legality, and degree rules into differentiable training signals.

\subsection{Encoder}
\label{sec:encoder}
 
The encoder maps the molecular graph into latent features with a Relational Transformer~\citep{diao2022relational}.
We first embed atoms and atom pairs:
\begin{equation}
    \bh_i^{(0)} = f_{\mathrm{enc-node}}(T_i,\bR_i),
    \qquad
    \bp_{ij}=f_{\mathrm{enc-edge}}(B_{ij},\|\bR_i-\bR_j\|_2).
    \label{eq:encoder-embed}
\end{equation}
The Relational Transformer updates atom features by edge-aware attention, where pair features $\bp_{ij}$ are injected into the attention between atoms $i$ and $j$:
\begin{equation}
    \bH
    =
    \operatorname{RTrans}_{\phi}
    \left(
        \{\bh_i^{(0)}\}_{i=1}^{N},\;
        \{\bp_{ij}\}_{i,j=1}^{N}
    \right).
    \label{eq:encoder-rtrans}
\end{equation}
The resulting atom-level features $\bH=(\bh_1,\ldots,\bh_N)$ are used to parameterize the latent distribution in Eq.~\eqref{eq:vae-encoding}.
 
\subsection{Topology-aware multi-stage decoder}
\label{sec:decoder}
 
Using the notation in Sec.~\ref{sec:prelim}, TopVAE decodes each latent code
$\bz$ in a topology-first order:
\begin{equation}
    p_\theta(\bA,\bT,\bm{B},\bR\mid\bz)
    =
    p_\theta(\bA\mid\bz)\,
    p_\theta(\bT\mid\bA,\bz)\,
    p_\theta(\bm{B}\mid\bA,\bT,\bz)\,
    p_\theta(\bR\mid\bA,\bT,\bm{B},\bz).
    \label{eq:decoder-factorization}
\end{equation}
This factorization makes adjacency the first decoded object. In particular,
bond prediction is constrained by
\begin{equation}
    A_{ij}=0 \Rightarrow B_{ij}=0,
    \qquad
    A_{ij}=1 \Rightarrow B_{ij}\in\{1,\ldots,K\}.
    \label{eq:adj-bond-constraint}
\end{equation}
 
Concretely, a node projection head $f_{\mathrm{node}}$ first maps each latent
token $\bz_i$ to an initial decoder feature $\bg_i^{(0)}$. The adjacency head $f_A$
then takes the pair feature $[\bg_i^{(0)},\bg_j^{(0)}]$ and predicts the edge-existence
probability:
\begin{equation}
    \bg_i^{(0)}=f_{\mathrm{node}}(\bz_i),
    \qquad
    P^A_{ij}=\sigma\!\left(f_A([\bg_i^{(0)},\bg_j^{(0)}])\right).
\end{equation}
 
Because molecular bonds are undirected and self-bonds are invalid, we symmetrize
$P^A$ and set its diagonal to zero; TopoBridge, denoted by $\Pi_{\mathrm{TB}}$,
then refines it into a connected adjacency matrix,
\begin{equation}
    \widetilde{\bA}=\Pi_{\mathrm{TB}}(P^A).
\end{equation}
 
The refined adjacency is injected into TopoFormer as an attention bias:
\begin{equation}
    \alpha_{ij}
    =
    \softmax_j
    \left(
        \frac{(\bq_i)^\top\bk_j}{\sqrt d}
        +
        \be_A(\widetilde A_{ij})
    \right),
    \qquad
    \bg_i \leftarrow \sum_j \alpha_{ij}\bv_j ,
    \label{eq:topoformer-attn}
\end{equation}
where $\bq_i,\bk_j,\bv_j$ are projections of the current decoder features and
$\be_A(\widetilde A_{ij})$ is a learned adjacency bias.
 
Given the atom-type distribution $P_i^T=\softmax(f_T(\bg_i))$, we form an
atom-conditioned feature $\ba_i^T$ from $P_i^T$ and use it in the bond head.
For each pair $(i,j)$, the bond head predicts logits over real bond types:
\begin{equation}
    \bo^B_{ij}
    =
    f_B(\bg_i,\bg_j,\ba_i^T,\ba_j^T,\widetilde A_{ij}),
    \qquad
    \bo^B_{ij}\in\mathbb{R}^{K}.
\end{equation}
The full bond distribution is then obtained by adjacency gating:
\begin{equation}
    P^B_{ij,0}=1-\widetilde A_{ij},
    \qquad
    P^B_{ij,c}
    =
    \widetilde A_{ij}\softmax(\bo^B_{ij})_c,
    \quad c=1,\ldots,K .
    \label{eq:bond-gating}
\end{equation}
Thus bond prediction is both adjacency-conditioned and atom-conditioned: non-adjacent
pairs are assigned no bond, while adjacent pairs are classified among real bond
types using the predicted atom information.
 
For coordinate reconstruction, the coordinate head first predicts an initial
coordinate matrix from the final decoder features,
\begin{equation}
    \bR^{(0)}=f_R(\bg).
\end{equation}
The bond distribution is first mapped to a continuous edge feature
$\be_{ij}^B=\sum_{c=0}^{K}P^B_{ij,c}\,\bw_c$ via learned bond
embeddings $\{\bw_c\}$. The EGNN then takes $\bR^{(0)}$, the decoder features
$\bg$, and $\be^B$ as inputs, and iteratively updates features and
coordinates:
\begin{equation}
    (\bg^{(\ell+1)},\bR^{(\ell+1)})
    =
    \mathrm{EGNN}_\ell(\bg^{(\ell)},\bR^{(\ell)},\be^B),
    \qquad
    \widehat{\bR}=\bR^{(L)} .
\end{equation}

\subsection{TopoBridge: connected adjacency refinement}
\label{sec:topobridge}

TopoBridge converts the soft adjacency probabilities $P^A$ into a binary
connected adjacency matrix $\widetilde{\bA}$.  It first builds an initial
undirected graph by keeping the top-$k$ most confident neighbours for each atom,
then symmetrises and removes self-loops.

Disconnected graphs are repaired by greedily adding bridge edges.  Choose a root
in the largest component and let $S$ be the set of atoms reachable by BFS.  If
$S\neq V$, add the most likely cross-component edge
\begin{equation}
    (u^\star,v^\star)
    =
    \arg\max_{u\in S,\;v\notin S}
    P^A_{uv},
    \label{eq:topobridge-bridge}
\end{equation}
and re-run BFS.  Each insertion strictly grows $|S|$, so the procedure
terminates and returns
$\widetilde{\bA}=\Pi_{\mathrm{TB}}(P^A)$ with
$\mathcal{G}(\widetilde{\bA})$ connected.

Because $\Pi_{\mathrm{TB}}$ is discrete, TopVAE applies a straight-through
surrogate~\citep{bengio2013estimating} for downstream decoder stages:
\begin{equation}
    \bA^{\mathrm{ST}}
    =
    P^A+\mathrm{stopgrad}\!\left(\widetilde{\bA}-P^A\right).
    \label{eq:topobridge-ste}
\end{equation}
The forward value is $\widetilde{\bA}$; gradients flow through $P^A$.  The
adjacency head is supervised from its logits against the ground-truth adjacency.

\paragraph{Over-connection risk.}
Because TopoBridge forces edges, atoms connected by a bridge necessarily receive
a nonzero bond type, which could bias the decoded graph toward
over-connectedness.  Empirically, however, atom-degree and
ring-size distributions of TopVAE are comparable to UDM-3D (degree EMD: $0.568$ vs.\ $0.558$; see Appendix~\ref{app:bridge_audit} and
Tables~\ref{tab:bridge_audit}--\ref{tab:degree_ring_dist}).
This is because bridge edges are selected by descending adjacency logit
(Eq.~\ref{eq:topobridge-bridge}), inserting only edges the model already
considers most likely, and the multi-stage decoder conditions downstream
bond-type and coordinate heads on the refined adjacency, integrating forced
edges into a globally consistent graph.

\paragraph{Computational overhead.}
TopoBridge's refinement adds only $2.8$\,ms per batch (${\sim}2.8\%$ of forward
time on an H100 NVL, batch size 64) for a converged model. Scaling is governed by the number of
bridge insertions rather than atom count $N$. Full profiling and scaling curves are in Appendix~\ref{app:computational_cost}.

\subsection{ChemCO: Unrolled Chemical Constraint Optimization}
\label{sec:chemco}

TopoBridge guarantees connectivity but does not constrain bond \emph{types}:
atom pairs may still receive chemically illegal bonds or violate valence caps.
ChemCO closes this gap by solving a constrained optimization problem over
bond-type assignments, producing a chemically improved distribution that
AGCL~(Sec.~\ref{sec:agcl}) selectively injects into training.

Let $k\!=\!1,\ldots,K$ index bond types with bond orders $o_k$, and
$k\!=\!0$ denote no bond. From the decoder's raw logits $U_{ij}^{(k)}$
over all $K\!+\!1$ bond classes (including no-bond $k\!=\!0$),
ChemCO takes the relative neural utility:
\begin{equation}
    \bar U_{ij}^{(k)}
    =
    U_{ij}^{(k)}-U_{ij}^{(0)},
    \qquad k=1,\ldots,K,
    \label{eq:chemco-utility-main}
\end{equation}
as a fixed input encoding the decoder's preference for each bond type.
ChemCO introduces free primal variables $\Phi^{(k)}$,
initialized as $\Phi_0^{(k)}\!=\!\bar U^{(k)}$, and maps them to a continuous
assignment score via
\begin{equation}
    Y^{(k)}
    =
    T_k(\Phi)
    :=
    \tfrac{1}{2}\!\bigl(g(\Phi^{(k)})+g(\Phi^{(k)})^\top\bigr)
    \odot\widetilde A
    \odot\Omega^{(k)},
    \quad
    g(x)=\sigma(x)^2,
    \label{eq:chemco-transform-main}
\end{equation}
where $\widetilde A$ is the TopoBridge adjacency support restricting
optimization to candidate edges, and
$\Omega^{(k)}$ is a binary mask that zeros out chemically invalid atom--bond
combinations (e.g., a triple bond between two oxygen atoms).
During training $\Omega^{(k)}$ is constructed from predicted atom types.
From $Y$, ChemCO derives
\begin{equation}
    s_{ij}=\textstyle\sum_{k}Y_{ij}^{(k)},
    \qquad
    \operatorname{val}_i=\textstyle\sum_{j\neq i}\sum_{k}o_k Y_{ij}^{(k)},
    \qquad
    \deg_i=\textstyle\sum_{j\neq i}s_{ij},
    \label{eq:chemco-stats}
\end{equation}
subject to three chemical constraints:
$s_{ij}\!\le\!1$ (each atom pair carries at most one bond type),
$\operatorname{val}_i\!\le\!c_i$ (total bond order at each atom does not exceed its valence capacity), and
$\deg_i\!\ge\!d_{\min}$ (every valid atom participates in at least one bond).

\paragraph{Objective.}
ChemCO maximizes expected bond utility while penalizing chemical
constraint violations through an adaptive-penalty scheme.  Each
violation is smoothly approximated by
$\psi_\gamma(r)=\gamma^{-1}\log(1+e^{\gamma r})$, a soft surrogate for
$[r]_+$.  The per-step score is
\begin{equation}
\begin{aligned}
    \mathcal J_t(\Phi)
    &=
    \textstyle\sum_{i<j}\sum_{k=1}^{K}
    \bar U_{ij}^{(k)}Y_{ij}^{(k)}
    -\sum_{i<j}
    \mu_{ij,t}\psi_\gamma(s_{ij}-1)
    \\[-2pt]
    &\quad
    -\textstyle\sum_i
    \lambda_{i,t}\psi_\gamma(\operatorname{val}_i-c_i)
    -\sum_i
    \nu_{i,t}\psi_\gamma(d_{\min}-\deg_i),
\end{aligned}
\label{eq:chemco-penalty-score}
\end{equation}
where $Y=T(\Phi)$.  The three penalty terms enforce bond exclusivity,
valence limits, and minimum-degree requirements respectively.  After
each gradient step on $\Phi$, multipliers are raised for still-violated
constraints, automatically concentrating pressure where it is most
needed.  Full update rules are in Appendix~\ref{app:chempo}.
We also provide analysis of the computational overhead; see Appendix~\ref{app:computational_cost}.

\subsection{Advantage-Gated Constraint Learning}
\label{sec:agcl}

ChemCO provides a constraint-optimization signal for learning chemical rules.
However, this signal should remain conservative: if constraint-induced
corrections are imposed indiscriminately, they may conflict with the
reconstruction objective. Following the
positive-advantage principle~\cite{self-imitation, advantage-learning}, AGCL
selectively injects the ChemCO signal only when it improves ground-truth bond
NLL on bonded pairs.

\begin{figure}[t]
    \centering
    \includegraphics[width=\linewidth]{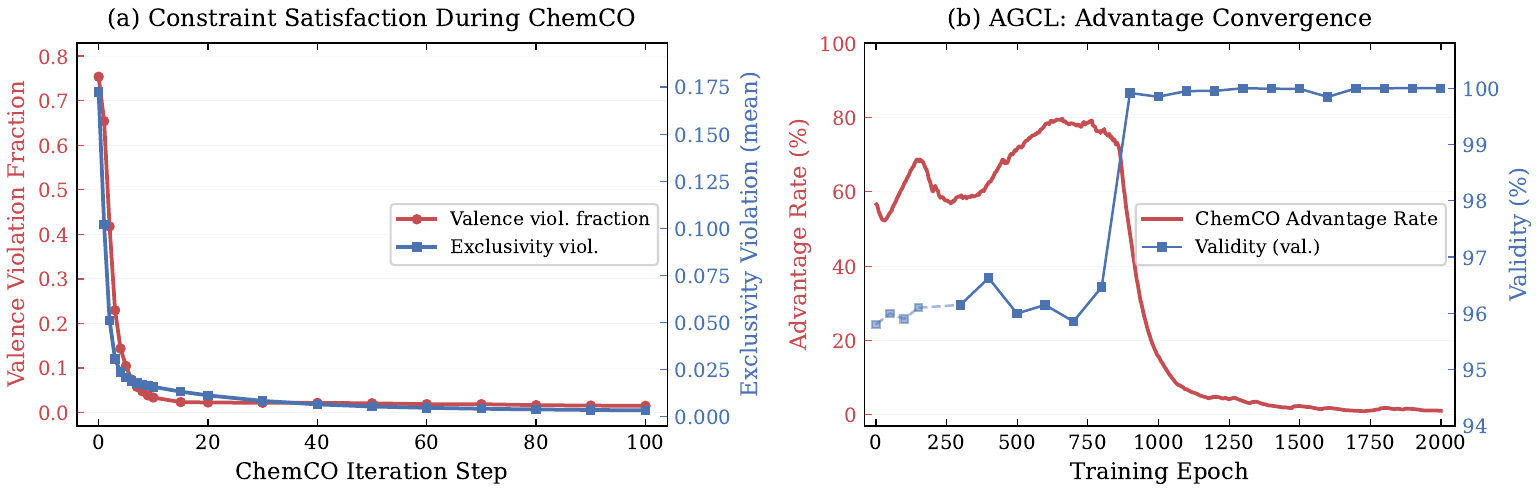}
    \caption{
        \textbf{ChemCO and AGCL dynamics.}
        (a) Averaged ChemCO optimization curves: red reports over-valent atoms, and blue reports mean pairwise bond-type exclusivity excess.
        (b) AGCL training curves: red reports the fraction of samples with
        positive advantage ($a_b\!>\!0$), and blue reports molecule validity.
        }
    \label{fig:chemco-agcl-effect}
\end{figure}

Let $P_{\mathrm{raw}}$ and $P_{\mathrm{chem}}$ be the bond distributions
before and after ChemCO. For molecule~$b$ with bonded pair set $\mathcal S_b$
and ground-truth labels $y_{ij}^{\star}$, the per-molecule advantage is
\begin{equation}
    \mathcal E_b(P)
    =
    \frac{1}{|\mathcal S_b|}
    \sum_{(i,j)\in\mathcal S_b}
    -\log P_{ij,y_{ij}^{\star}},
    \qquad
    a_b
    =
    \left[
        \mathcal E_b(P_{\mathrm{raw}})
        -
        \mathcal E_b(P_{\mathrm{chem}})
    \right]_+ .
    \label{eq:agcl-advantage}
\end{equation}
Thus $a_b\!>\!0$ only when ChemCO lowers the ground-truth bond NLL.
AGCL uses this advantage to weight a consistency loss:
\begin{equation}
    \mathcal L_{\mathrm{AGCL}}
    =
    \frac{
        \sum_b a_b
        \sum_{(i,j)\in\mathcal S_b}
        \left\|
            P_{\mathrm{raw},ij}
            -
            \operatorname{sg}(P_{\mathrm{chem},ij})
        \right\|_2^2
    }{
        (K+1)\sum_b a_b|\mathcal S_b|+\epsilon
    } ,
    \label{eq:agcl-loss}
\end{equation}
where $\operatorname{sg}(\cdot)$ detaches the ChemCO output so that
$\mathcal{L}_{\mathrm{AGCL}}$ steers only the raw decoder without
back-propagating through the unrolled solver.
As training progresses, the raw decoder internalizes the chemical constraints:
the fraction of molecules with positive advantage $a_b\!>\!0$ approaches zero
(Figure~\ref{fig:chemco-agcl-effect}) and ChemCO can be removed at inference
time (Table~\ref{tab:ablation}).

\subsection{Training Objective}
\label{sec:training-objective}

TopVAE is trained end-to-end with reconstruction losses for each decoded
modality, KL regularization, and the AGCL selective teacher:
\begin{equation}
    \mathcal{L}
    =
    \mathcal{L}_{\mathrm{topo}}
    +\mathcal{L}_{\mathrm{atom}}
    +\lambda_{3\mathrm{D}}\,\mathcal{L}_{\mathrm{3D}}
    +\beta\,\mathcal{L}_{\mathrm{KL}}
    +\lambda_{\mathrm{AGCL}}\,\mathcal{L}_{\mathrm{AGCL}}.
    \label{eq:total-loss}
\end{equation}
Here $\mathcal{L}_{\mathrm{topo}}$ groups focal binary cross-entropy for the
adjacency matrix and cross-entropy for bond types,
$\mathcal{L}_{\mathrm{atom}}$ is cross-entropy over atom types,
$\mathcal{L}_{\mathrm{3D}}$ combines coordinate regression and pairwise
distance supervision, and $\mathcal{L}_{\mathrm{KL}}$ is the standard VAE
KL divergence.
All individual loss terms, weights, architectural details, ChemCO
hyperparameters, and warmup schedules are reported in
Appendix~\ref{app:training}.


\section{Experiments}
\label{sec:results}

\subsection{Setup}

\textbf{Datasets.}
QM9~\citep{ramakrishnan2014quantum} contains ${\sim}$134k small organic molecules with up to 9 heavy atoms (C, N, O, F) and their equilibrium geometries computed at the DFT level.
GEOM-Drugs~\citep{axelrod2022geom} contains ${\sim}$304k drug-like molecules with up to 90 heavy atoms and multiple conformers per molecule.

\textbf{Metrics.}
Atom stability and molecular stability (MolStab) follow the definitions in 
EDM~\citep{hoogeboom2022equivariant}: an atom is stable if its valence equals the 
reference value, and a molecule is stable if all its atoms are stable (not necessarily connected).
Validity requires successful RDKit sanitization.
Connectivity requires the molecular graph to be a single connected component.
We report joint metrics \textbf{V\&C} (validity $\wedge$ connectivity), \textbf{V\&U} 
(validity $\wedge$ uniqueness), and \textbf{S$\wedge$C} (MolStable $\wedge$ connectivity).
\textbf{FCD} measures distributional similarity to the reference set in chemical descriptor 
space~\citep{preuer2018frechet}.
\textbf{iFID} adapts the interpolated FID to molecular VAE latent spaces~\citep{xu2026making}, 
decoding SLERP midpoints between posterior means and measuring the Fr\'{e}chet distance of 
ChemNet activations (details in Appendix~\ref{app:metrics}).



\subsection{VAE Dark-Area Diagnosis}
\label{sec:dark-area-diagnosis}

To quantify dark areas, we evaluate each VAE decoder under latent perturbations of increasing magnitude.
For each validation molecule $M$, we encode its posterior mean $\bz = \bmu_\phi(M)$ and decode the perturbed latent
\begin{equation}
    \tilde{\bz} = \bz + \sigma \cdot s_\mathrm{emp} \cdot \bepsilon, \qquad \bepsilon \sim \mathcal{N}(\bzero, \bI),
    \label{eq:noise-perturb}
\end{equation}
where $s_\mathrm{emp} = \mathrm{std}(\{\bmu_\phi(M)\}_{M \in \mathcal{D}})$ is the empirical standard deviation of the encoded latents and $\sigma$ controls the perturbation strength relative to the latent spread.

Table~\ref{tab:vae_diag} reports results at $\sigma{=}0$ and $0.5$.
Existing VAEs collapse under perturbation: S$\wedge$C drops by over an order of 
magnitude on QM9 and falls to zero on GEOM-Drugs, whereas \method{} achieves higher S$\wedge$C.
PCA projection (Figure~\ref{fig:dark-area-embeddings}) confirms the gap---\method{} 
decodes $90.5\%$ of perturbed latents as valid connected molecules versus $44.4\%$ (UAE) and $32.6\%$ (ADiT).
This robustness extends to the diffusion trajectory: TopVAE+DiT reaches $>$90\% validity 
by denoising progress 0.25 while UDM-3D does not saturate until progress ${\sim}$0.90
(Figures~\ref{fig:dark-area-diffusion}--\ref{fig:dark-area-showcase}, Appendix~\ref{app:dark_area}).

\begin{table}[t]
\centering
\caption{\textbf{VAE dark-area diagnosis} on QM9 and GEOM-Drugs ($n{=}10\text{k}$).
\method{} maintains substantially higher decodability under noise perturbation.}
\label{tab:vae_diag}
\footnotesize
\setlength{\tabcolsep}{4pt}
\begin{tabular}{@{}l cccc@{}}
\toprule
& \makecell{Post.\\MolStab$\,\uparrow$}
& \makecell{Post.\\S$\wedge$C$\,\uparrow$}
& \makecell{$\sigma{=}0.5$\\S$\wedge$C$\,\uparrow$}
& \makecell{iFID\\$\downarrow$} \\
\midrule
\multicolumn{5}{@{}l}{\emph{QM9}} \\[2pt]
UAE            & 1.000 & 0.998 & 0.023 & 2.88 \\
ADiT-VAE      & 0.951 & 0.951 & 0.000 & 7.08 \\
\rowcolor{rowhl}
\textbf{\method{}} & \textbf{1.000} & \textbf{1.000} & \textbf{0.662} & \textbf{0.395} \\
\midrule
\multicolumn{5}{@{}l}{\emph{GEOM-Drugs}} \\[2pt]
UAE            & 0.998 & 0.000 & 0.000 & 62.00 \\
\rowcolor{rowhl}
\textbf{\method{}} & \textbf{0.960} & \textbf{0.960} & \textbf{0.033} & \textbf{42.77} \\
\bottomrule
\end{tabular}
\end{table}

\subsection{Latent Diffusion Generation on QM9 and GEOM-Drugs}

\paragraph{QM9.}
Table~\ref{tab:qm9} compares de novo generation on QM9.
\method{}+DiT achieves perfect V\&C and the best FCD\textsubscript{3D} ($77\%$ lower over previous SOTA), indicating that dark-area closure translates directly into higher-quality 3D generation.
QM9 is near-saturated for MolStab---all competitive methods exceed $0.97$---so the principal gains are in topology-consistent generation and 3D distributional fidelity.

\paragraph{GEOM-Drugs.}
Table~\ref{tab:geom} reports results on the larger GEOM-Drugs benchmark, where dark areas are more severe.
\method{}+DiT achieves the highest V\&C ($0.951$, $+8\%$), V\&U ($1.000$), and FCD\textsubscript{3D} ($8.25$, $-52\%$), while remaining competitive on 2D distributional metrics.

\paragraph{Summary.}
The V\&C improvements on both benchmarks are direct consequences of TopoBridge (guaranteeing connectivity) and ChemCO (enforcing chemical validity), which together mitigate the dark areas diagnosed in Section~\ref{sec:dark-area-diagnosis}.
The FCD\textsubscript{3D} gains further suggest that a smoother, more navigable latent space enables the diffusion prior to produce latent codes that decode into higher-fidelity 3D geometries. 
We provide additional analysis of 3D conformations, chemical properties, and diversity distributions (Appendices~\ref{app:3d_conformer}--\ref{app:energy_relaxation}).

\begin{table*}[t]
\centering
\caption{\textbf{QM9 de novo 3D molecule generation} ($n{=}10{,}000$).
Baseline results are from respective papers; \method{}+DiT is our evaluation.}
\label{tab:qm9}
\footnotesize
\setlength{\tabcolsep}{3.2pt}
\begin{tabular*}{\textwidth}{@{}l@{\extracolsep{\fill}}*{8}{c}@{}}
\toprule
& \multicolumn{5}{c}{\textbf{2D Metrics}} & \multicolumn{3}{c}{\textbf{3D Metrics}} \\
\cmidrule(lr){2-6} \cmidrule(l){7-9}
Method
  & FCD$\downarrow$
  & \makecell{Atom\\Stab.$\uparrow$}
  & \makecell{Mol\\Stab.$\uparrow$}
  & V\&C$\uparrow$
  & V\&U$\uparrow$
  & FCD\textsubscript{3D}$\downarrow$
  & \makecell{Atom\\Stab.$\uparrow$}
  & \makecell{Mol\\Stab.$\uparrow$} \\
\midrule
EDM               & 1.285 & 0.986 & 0.817 & 0.934 & 0.907 & 1.285 & 0.986 & 0.817 \\
CDGS              & 0.798 & 0.997 & 0.951 & 0.936 & 0.860 & \na   & \na   & \na  \\
GeoLDM            & 1.030 & 0.989 & 0.894 & 0.951 & 0.927 & 1.030 & 0.989 & 0.897 \\
LDM-3DG           & 0.559 & 0.976 & 0.869 & 1.000 & 0.953 & \na   & \na   & \na  \\
MolFLAE           & \na   & 0.994 & 0.920 & \na   & 0.889 & \na   & \na   & \na  \\
GFMDiff           & \na   & 0.989 & 0.877 & \na   & 0.951 & \na   & 0.989 & 0.877 \\
MiDi              & 0.187 & 0.998 & 0.976 & 0.980 & 0.954 & 1.100 & 0.983 & 0.842 \\
JODO              & \underline{0.138} & 0.999 & \underline{0.988} & \underline{0.990} & \underline{0.960} & 0.885 & \underline{0.992} & 0.934 \\
EQGAT-diff        & 2.088 & 0.999 & 0.971 & 0.965 & 0.950 & 1.520 & 0.988 & 0.888 \\
SemlaFlow         & 0.863 & 0.995 & 0.949 & 0.857 & 0.821 & 1.127 & 0.971 & 0.787 \\
UDM-3D               & \textbf{0.130} & 0.999 & \underline{0.988} & 0.983 & \textbf{0.973} & \underline{0.881} & \textbf{0.993} & \textbf{0.935} \\
\hdashline\noalign{\vskip 3pt}
\rowcolor{rowhl}
\textbf{\method{}+DiT} & 0.185 & 0.996 & \textbf{1.000} & \textbf{1.000} & 0.959 & \textbf{0.207}\textsuperscript{\colorbox{gray!15}{\textcolor{cyan!70!black}{\scriptsize$\downarrow$77\%}}} & 0.987 & \underline{0.925} \\
\bottomrule
\end{tabular*}
\end{table*}

\begin{table*}[t]
\centering
\caption{\textbf{GEOM-Drugs de novo 3D molecule generation} ($n{=}10{,}000$).
Baseline results are from respective papers; \method{}+DiT is our evaluation.}
\label{tab:geom}
\footnotesize
\setlength{\tabcolsep}{3.2pt}
\begin{tabular*}{\textwidth}{@{}l@{\extracolsep{\fill}}*{8}{c}@{}}
\toprule
& \multicolumn{5}{c}{\textbf{2D Metrics}} & \multicolumn{3}{c}{\textbf{3D Metrics}} \\
\cmidrule(lr){2-6} \cmidrule(l){7-9}
Method
  & FCD$\downarrow$
  & \makecell{Atom\\Stab.$\uparrow$}
  & \makecell{Mol\\Stab.$\uparrow$}
  & V\&C$\uparrow$
  & V\&U$\uparrow$
  & FCD\textsubscript{3D}$\downarrow$
  & \makecell{Atom\\Stab.$\uparrow$}
  & \makecell{Mol\\Stab.$\uparrow$} \\
\midrule
EDM               & 40.14 & 0.991 & 0.914 & 0.359 & 0.991 & 31.29 & 0.831 & 0.002 \\
CDGS              & 22.05 & 0.991 & 0.706 & 0.285 & 0.285 & \na   & \na   & \na  \\
GeoLDM            & 39.81 & 0.996 & 0.909 & 0.482 & 0.998 & 30.68 & 0.843 & 0.008 \\
MiDi              & 7.054 & 0.968 & 0.822 & 0.633 & 0.654 & 23.14 & 0.750 & 0.003 \\
JODO              & \underline{2.523} & \textbf{1.000} & \underline{0.981} & 0.874 & 0.902 & 19.99 & \underline{0.845} & 0.010 \\
EQGAT-diff        & 5.898 & \textbf{1.000} & \textbf{0.989} & 0.845 & 0.863 & 26.33 & 0.825 & 0.007 \\
UDM-3D            & \textbf{0.692} & \textbf{1.000} & 0.925 & \underline{0.879} & \underline{0.907} & \underline{17.36} & \textbf{0.852} & \underline{0.014} \\
\hdashline\noalign{\vskip 3pt}
\rowcolor{rowhl}
\textbf{\method{}+DiT} & 2.68 & \textbf{1.000} & \underline{0.981} & \textbf{0.951}\textsuperscript{\colorbox{gray!15}{\textcolor{cyan!70!black}{\scriptsize$\uparrow$8\%}}} & \textbf{1.000}\textsuperscript{\colorbox{gray!15}{\textcolor{cyan!70!black}{\scriptsize$\uparrow$10\%}}} & \textbf{8.25}\textsuperscript{\colorbox{gray!15}{\textcolor{cyan!70!black}{\scriptsize$\downarrow$52\%}}} & 0.832 & \textbf{0.020}\textsuperscript{\colorbox{gray!15}{\textcolor{cyan!70!black}{\scriptsize$\uparrow$43\%}}} \\
\bottomrule
\end{tabular*}
\end{table*}

\subsection{Zero-Shot Scaffold Inpainting}
\label{sec:inpainting}

Scaffold inpainting stress-tests dark areas under OOD conditions: given a scaffold with $k$ atoms, we encode it into $k$ latent tokens and fill the remaining $n{-}k$ slots (heavy atoms) via reverse diffusion.
As the expansion size grows, a larger fraction of the latent code is sampled off-posterior, increasing exposure to $\mathcal{D}_{\mathrm{dark}}$.

Table~\ref{tab:geom_inpaint} reports results on GEOM-Drugs across five scaffolds (benzene, pyridine, naphthalene, indole, cyclohexane; 1k samples each; per-scaffold breakdown in Appendix~\ref{app:inpaint}).
\method{}+DiT consistently outperforms UDM-3D, with the gap widening at larger expansions: at $+60$ atoms, UAE collapses to $5.6\%$ S$\wedge$C while \method{} maintains $51.4\%$; at $+80$ and beyond, UAE produces near-zero valid molecules while \method{} continues to generate scaffold-preserving structures.

\begin{table}[t]
\centering
\caption{\textbf{Zero-shot scaffold inpainting on GEOM-Drugs} (mean over scaffolds).
Per-scaffold breakdown in Appendix Table~\ref{tab:geom_inpaint_detail}.}
\label{tab:geom_inpaint}
\small
\setlength{\tabcolsep}{4pt}
\begin{tabular}{@{}r cc cc c r cc cc@{}}
\toprule
& \multicolumn{2}{c}{\textbf{Scaff.\ Pres.}\,$\uparrow$}
& \multicolumn{2}{c}{\textbf{S$\wedge$C}\,$\uparrow$}
& \phantom{a}
& & \multicolumn{2}{c}{\textbf{Scaff.\ Pres.}\,$\uparrow$}
& \multicolumn{2}{c}{\textbf{S$\wedge$C}\,$\uparrow$} \\
\cmidrule(lr){2-3} \cmidrule(lr){4-5}
\cmidrule(lr){8-9} \cmidrule(l){10-11}
$+n$ & UAE & \method{} & UAE & \method{}
& & $+n$ & UAE & \method{} & UAE & \method{} \\
\midrule
5  & 10.0 & \textbf{11.3} & 89.8 & \textbf{95.9}
&& 60  & 5.0  & \textbf{28.9} & 5.6  & \textbf{51.4} \\
10 & 14.8 & \textbf{15.4} & 86.1 & \textbf{91.1}
&& 80  & 0.0  & \textbf{25.2} & 0.1  & \textbf{26.4} \\
20 & 15.2 & \textbf{15.9} & \textbf{86.9} & 84.8
&& 100 & 0.0  & \textbf{19.8} & 0.0  & \textbf{8.6}  \\
40 & 17.9 & \textbf{27.8} & 63.9 & \textbf{69.6}
&& \\
\midrule
\rowcolor{rowhl}
\multicolumn{6}{@{}r}{} &
\textbf{Mean} & 9.0 & \textbf{20.6} & 47.5 & \textbf{61.1} \\
\bottomrule
\end{tabular}
\end{table}

\subsection{Ablation}
\label{sec:ablation}

\paragraph{Component ablation.}
Table~\ref{tab:ablation} isolates the contribution of each module.
Without ChemCO, $\sigma{=}0.5$ Stab$\wedge$Conn drops from $0.662$ to $0.02$: the decoder loses chemical validity off-manifold.
Without TopoBridge, posterior MolStab remains high ($0.951$) but Stab$\wedge$Conn falls to $0.302$: molecules pass valence checks yet fragment into disconnected components.

\paragraph{Inference-time ablation.}
To verify that the decoder has learned chemical rules rather than relying on ChemCO as a post-processing step, Table~\ref{tab:ablation} varies the number of ChemCO iterations at inference from $0$ (fully disabled) to $100$.
All metrics remain stable, confirming that AGCL has transferred the constraint knowledge into the decoder weights during training.

\paragraph{Hyperparameter sensitivity.}
Appendix~\ref{app:sensitivity} reports sensitivity analysis for
TopoBridge's adjacency sparsity,
ChemCO's minimum-degree penalty ($d_{\min}\in\{0,1\}$), and the $\Omega^{(k)}$
chemical mask. TopVAE is robust to adjacency sparsity and hyperparameters.


\begin{table}[t]
\centering
\caption{\textbf{Ablation studies.}
\textbf{(a)}~Component ablation on QM9 ($n{=}10k$).
\textbf{(b)}~Inference-time ablation: varying ChemCO iterations at inference on GEOM-Drugs ($n{=}10k$). The decoder produces similar results with or without ChemCO.}
\label{tab:ablation}

\vspace{4pt}
\footnotesize

\begin{minipage}[t]{0.58\textwidth}
\centering
\textbf{(a) Component ablation}\\[4pt]
\setlength{\tabcolsep}{3pt}
\begin{tabular}{@{}l cccc cc@{}}
\toprule
& \multicolumn{4}{c}{Posterior ($\sigma{=}0$)}
& \multicolumn{2}{c}{Noise $\sigma{=}0.5$} \\
\cmidrule(lr){2-5}\cmidrule(lr){6-7}
Configuration
  & MolStab & Conn. & S$\wedge$C
  & $\Delta$S$\wedge$C
  & MolStab & S$\wedge$C \\
\midrule
\rowcolor{rowhl}
\textbf{Full TopVAE}
  & \textbf{1.000} & \textbf{1.000} & \textbf{1.000} & --
  & \textbf{0.682} & \textbf{0.662} \\
\midrule
w/o ChemCO
  & 0.941 & 1.000 & 0.941 & --5.9\%
  & 0.410 & 0.020 \\
w/o TopoBridge
  & 0.951 & 0.318 & 0.302 & --69.8\%
  & 0.405 & 0.010 \\
\bottomrule
\end{tabular}
\end{minipage}%
\hfill
\begin{minipage}[t]{0.38\textwidth}
\centering
\textbf{(b) Inference-time ablation}\\[4pt]
\setlength{\tabcolsep}{3pt}
\begin{tabular}{@{}r ccc cc@{}}
\toprule
\makecell{ChemCO\\Steps}
  & \makecell{Mol\\Stab.$\uparrow$}
  & S$\wedge$C$\uparrow$
  & Conn.$\uparrow$
  & FCD$\downarrow$
  & FCD\textsubscript{3D}$\downarrow$ \\
\midrule
0   & \textbf{0.981} & \textbf{0.933} & \textbf{0.951} & 2.68 & 8.25 \\
10  & 0.980 & 0.931 & \textbf{0.951} & 2.68 & 8.25 \\
\rowcolor{rowhl}
100 & 0.979 & 0.928 & 0.948 & 2.68 & \textbf{8.24} \\
\bottomrule
\end{tabular}
\end{minipage}

\end{table}

\section{Conclusion}
\label{sec:conclusion}

We identified \emph{dark areas} in molecular latent spaces, i.e.,
regions where existing VAE decoders produce chemically invalid or
disconnected structures despite near-perfect posterior reconstruction,
and showed that per-atom stability metrics mask this failure by
accepting fragmented atom clouds as stable.
TopVAE closes dark areas through a selective-teacher paradigm:
TopoBridge guarantees connectivity, ChemCO formulates chemical rules
as differentiable constrained optimization, and AGCL gates this signal
by positive advantage so the decoder progressively internalizes the
constraints and ChemCO can be removed at inference.
The resulting smoother latent manifold yields $77\%$ lower
FCD\textsubscript{3D} on QM9, the highest V\&C and $52\%$ lower
FCD\textsubscript{3D} on GEOM-Drugs, and $1.29{\times}$ higher
molecular stability on zero-shot scaffold inpainting, demonstrating
that dark-area closure translates directly into generation quality
across both in-distribution and out-of-distribution regimes.

\section{Limitations}
\label{sec:limitations}

The ChemCO--AGCL framework currently enforces only graph-level valence
rules. In principle, any differentiable constraint admitting a
primal--dual formulation can serve as a selective training signal: 3D
physical priors such as van der Waals clash penalties, force-field
energy bounds, and steric strain limits can be unrolled in the same
manner, enabling decoders to internalize geometric feasibility
alongside chemical validity and produce physically plausible molecules
without post-hoc relaxation; incorporating such priors remains future
work. Likewise, extending TopVAE to conditional generation targeting
quantum properties, bioactivity profiles, or protein-pocket
constraints is a natural next step toward latent molecular design.

\begin{ack}
Funding disclosure will be added in the camera-ready version.
\end{ack}

\clearpage
\bibliographystyle{plainnat}
\bibliography{references}

\clearpage
\appendix

\section{Extended Related Work}
\label{app:extended_related}

This appendix expands on the related work discussion in
Sec.~\ref{sec:related}, providing detailed comparisons with prior methods and
clarifying where \method{} fits relative to data-space molecular generators,
molecular latent diffusion models, VAE latent-space analyses, and
constraint-aware decoding methods.

\paragraph{Data-space 3D molecular generation.}
Early 3D molecular diffusion models generate molecules directly in data
space. EDM introduced E(3)-equivariant diffusion over atom coordinates and
categorical atom features~\citep{hoogeboom2022equivariant}. Because EDM does
not generate bond variables explicitly, molecular bonds are typically inferred
after generation through distance-based or chemistry-based post-processing.
This post-hoc bond recovery can create graph--geometry mismatch: the generated
atom types and coordinates may not admit a chemically consistent bond graph.
MolDiff formalized this issue as atom--bond inconsistency and proposed to
generate atom and bond information jointly~\citep{peng2023moldiff}.

Subsequent data-space methods make molecular topology, bond variables, or
bond-formation information part of the generative process. JODO jointly models
atom types, formal charges, bond information, and 3D coordinates through a
joint 2D--3D diffusion formulation~\citep{huang2023learning}. MUDiff combines
discrete graph diffusion with continuous coordinate diffusion for molecular
graphs and conformations~\citep{hua2024mudiff}. MiDi performs denoising over
both molecular graphs and 3D atom arrangements~\citep{vignac2023midi}.
EQGAT-diff studies equivariant diffusion with mixed categorical and continuous
variables, including atom, bond, and coordinate channels~\citep{le2023navigating}.
Recent scalable or graph-aware molecular generators further improve
continuous--discrete denoising architectures and bond-formation-aware training
objectives~\citep{Megalodon,GFMDiff}.

These data-space methods reduce the mismatch between generated geometry and
molecular graph structure, but they run the generative dynamics over the full
molecular representation. They therefore do not isolate the molecular decoder
as an object of study. In contrast, \method{} focuses on the decoder used by a
latent generative pipeline: given a compact continuous latent code, the decoder
must produce a connected and chemically valid molecule not only on posterior
latents, but also on interpolated, prior-like, and diffusion-perturbed latents.

\paragraph{Molecular latent diffusion and autoencoding.}
Latent diffusion separates representation learning from prior learning: a
encoder first maps data into a lower-dimensional latent space, and a
diffusion model is then trained in that latent space~\citep{rombach2022high}.
Transformer-based diffusion priors such as DiT further improve the scalability
of latent-space generation~\citep{peebles2023scalable}. This paradigm is
especially attractive for 3D molecules, where the raw data contain mixed
discrete--continuous variables, permutation structure, chemical constraints,
and geometric symmetries.

Several recent methods adapt latent diffusion or latent autoencoding to 3D
molecular generation. GeoLDM constructs a point-structured molecular latent
space with invariant scalar and equivariant vector components, and trains a
latent diffusion model over this representation~\citep{xu2023geometric}.
Latent 3D Graph Diffusion provides theoretical motivation for molecular latent
diffusion, emphasizing that useful latents should be low-dimensional,
reconstructive, and symmetry-preserving~\citep{you2024latent}. UAE-3D
compresses atom types, bonds, and coordinates into a unified latent sequence
and applies a standard DiT prior without molecule-specific inductive bias in
the latent generator~\citep{luo2025towards}. ADiT extends latent diffusion to a
shared framework for all-atom molecules and periodic materials
~\citep{joshi2025all}. Related latent autoencoding work such as MolFLAE learns
fixed-dimensional E(3)-equivariant molecular latents for zero-shot molecular
manipulation, without being limited to the standard ``autoencoder plus
diffusion prior'' pipeline~\citep{chen2025manipulating}.

These works primarily address how to construct expressive, compact, and
symmetry-aware molecular latent spaces. \method{} addresses a complementary
failure mode: even when reconstruction quality is high, a decoder can fail on
the latent regions actually queried by a diffusion prior. Our dark-area
diagnostic evaluates whether prior samples, interpolation paths, and
diffusion-perturbed latents remain chemically decodable. This shifts attention
from reconstruction-only autoencoding to the operational robustness required
for latent molecular generation.

\paragraph{Latent-space quality in VAEs.}
A VAE decoder is trained on samples from the encoder posterior, but downstream
latent generation may query regions closer to the aggregated posterior,
interpolation paths between encoded molecules, or trajectories produced by a
learned latent prior. This creates a gap between reconstruction-time decoding
and generation-time decoding. Prior analyses relate this gap to
aggregate-posterior mismatch and to the hypothesized, empirically debated
phenomenon of low-density latent ``holes''~\citep{hoffman2017beta,li2021low}.
For molecular generation, such holes are especially consequential: an
off-posterior latent may decode not merely to a low-quality sample, but to a
disconnected graph or a molecule that violates chemical valence and
sanitization constraints.

Earlier molecular VAEs already showed that continuous molecular latent spaces
are useful for optimization, while also exposing the importance of validity and
representation bias in molecular decoding. The original chemical VAE maps
SMILES strings into a continuous latent space for molecular design
~\citep{gomez2018automatic}. Grammar VAE uses grammar constraints to improve
syntactic validity~\citep{kusner2017grammar}, and Syntax-Directed VAE further
incorporates syntax and semantic constraints into the decoder
~\citep{dai2018syntax}. JT-VAE generates a junction tree over chemical
substructures before assembling the molecular graph, improving chemical
validity through a substructure-level generative process
~\citep{jin2018junction}. SELFIES provides a semantically robust molecular
string representation designed to improve molecular validity at the
representation level~\citep{krenn2022selfies}. NP-VAE constructs latent spaces
for large natural products and complex molecular structures
~\citep{ochiai2023variational}. Beyond molecules, SRL-VAE shows that
adversarial smoothing can improve latent robustness in image VAEs
~\citep{lee2025enhancing}.

These methods motivate the importance of smooth and valid latent decoding, but
they do not directly test a 3D molecular VAE under the latent distribution
induced by modern latent diffusion. \method{} makes this test explicit through
dark-area analysis: we evaluate whether the decoder remains valid under
posterior perturbations, interpolation, prior sampling, and diffusion sampling,
and we identify topological disconnection and chemical invalidity as two major
failure modes.

\paragraph{Constraint-aware graph and molecule decoding.}
Validity-aware molecular generation has been addressed at multiple levels.
At the representation level, grammar-based methods and SELFIES restrict the
output language so that decoded strings better respect syntactic or semantic
molecular validity~\citep{kusner2017grammar,krenn2022selfies}. At the
substructure or graph-decoder level, JT-VAE and constrained graph VAEs impose
chemical validity through the generative procedure itself: JT-VAE builds
molecules from chemical substructure trees~\citep{jin2018junction}, while
CGVAE uses constrained graph extension to guide molecular graph generation
~\citep{liu2018constrained}. At the graph-regularization level, regularized
graph VAEs penalize invalid matrix-valued outputs to encourage properties such
as connectivity and valence consistency~\citep{ma2018constrained}.

Another line of work introduces differentiable optimization or satisfiability
layers inside neural networks. OptNet embeds a differentiable quadratic
programming solver as a neural network layer~\citep{amos2017optnet}, while
LinSATNet and GLinSAT project neural outputs toward feasible sets defined by
linear satisfiability constraints~\citep{wang2023linsatnet,zeng2024glinsat}.
These methods provide useful templates for differentiable constrained
prediction, but molecular validity is not simply a generic linear feasibility
problem. It involves discrete support constraints, atom-type-dependent valence,
bond-type compatibility, global graph connectivity, aromaticity and charge
consistency, and consistency between topology and 3D geometry.

Constrained diffusion methods impose hard structural constraints along the
generative trajectory. For example, ConStruct maintains graph constraints such
as planarity or acyclicity by combining edge-absorbing noise with projection
operators~\citep{madeira2024generative}. These structural graph constraints
are related to molecular validity, but they are not identical to chemical
sanitization constraints. In particular, molecular decoding must handle both
global graph structure and local chemistry-specific rules.

\method{} takes a different route from permanent projection layers and
post-hoc sanitization. A permanent projection layer can create decoder
dependency: the raw neural decoder may never internalize the constraints and
may require the projection module at inference time. Post-hoc sanitization has
the opposite limitation: it can repair invalid outputs after decoding, but it
does not provide a training signal that teaches the decoder why the output was
invalid. \method{} uses ChemCO+AGCL as a selective training-time teacher.
ChemCO computes chemically constrained bond corrections through unrolled
optimization, and AGCL injects these corrections only when they improve
ground-truth bond likelihood. As a result, the default decoder progressively
internalizes chemical rules and can be used without ChemCO during inference.

\paragraph{Positioning.}
\method{} sits between complete-molecule data-space generation and molecular
latent diffusion. Like complete-molecule data-space models, it treats molecular
topology and bond types as first-class objects rather than recovering bonds
only after coordinate generation. Like latent diffusion methods, it preserves
the efficiency and controllability of a compact continuous latent space. Its
central contribution is decoder-centric: instead of asking only whether the
encoder reconstructs posterior latents, \method{} asks whether the latent
space is chemically navigable in the regions that a diffusion prior will
actually visit.

Concretely, \method{} addresses the two dominant dark-area failure modes.
TopoBridge guarantees connected candidate topology through adjacency
refinement, preventing disconnected molecular fragments. ChemCO enforces
chemistry-specific bond and valence constraints through unrolled primal--dual
optimization. AGCL then distills ChemCO's corrections into the raw decoder
during training, enabling ChemCO-free inference. This makes \method{} a
complement to prior molecular latent diffusion models: it improves the
decodability and chemical robustness of the latent space on which such priors
depend.

\clearpage
\section{Extended Dark-Area Analysis}
\label{app:dark_area}

VAE decoders can produce invalid
outputs when the latent code falls outside the support of the
aggregate posterior---so-called \emph{dark areas} of the latent space.
The diffusion prior is designed to keep generated latent codes on the
data manifold, but the degree to which it succeeds depends on both the
prior's expressiveness and the decoder's robustness.  We analyse dark
areas from two complementary perspectives: the static geometry of the
VAE latent space (Section~\ref{app:dark_area_vae}) and the dynamic
denoising trajectory of the diffusion prior
(Section~\ref{app:dark_area_diffusion}).

\subsection{VAE Latent Space Geometry}
\label{app:dark_area_vae}

\begin{figure}[t]
\centering
\includegraphics[width=\linewidth]{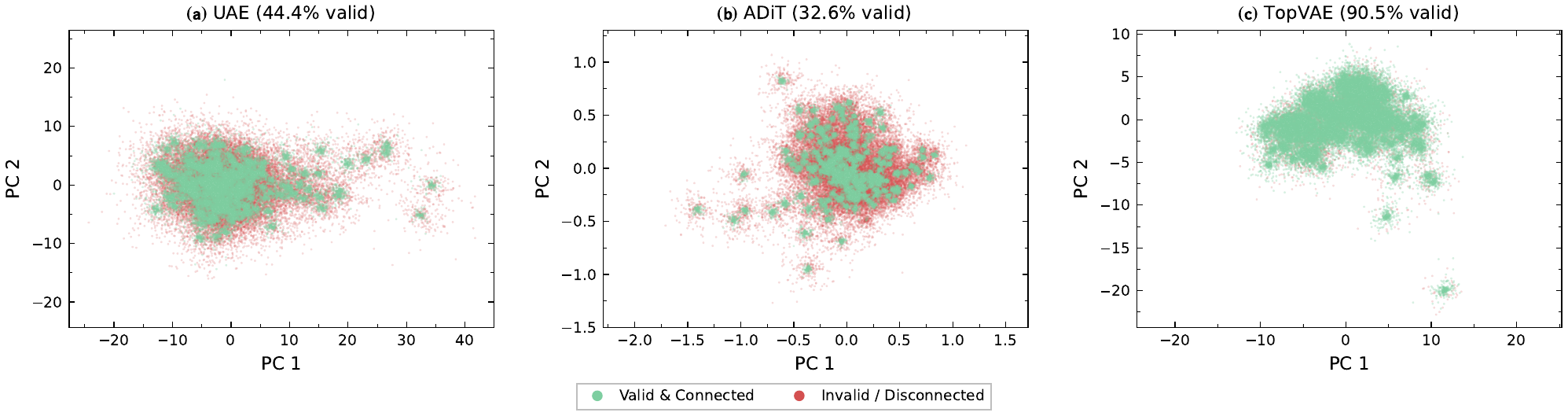}
\caption{\textbf{VAE dark-area embeddings.}  2-D PCA of per-molecule
mean latent vectors ($n{=}2{,}000$, QM9 test set).  Green
points decode to valid molecules; red points decode to invalid ones.
Invalid molecules concentrate in low-density peripheral regions,
delineating the dark areas of each latent space.}
\label{fig:dark-area-embeddings}
\end{figure}

To probe the extent of dark areas, we randomly sample 2{,}000 QM9
molecules and encode each through three VAEs (UAE, TopVAE, ADiT).
For each latent code $z = \mu_\phi(M)$, we add Gaussian perturbations
$z' = z + \sigma \cdot s_{\mathrm{emp}} \cdot \varepsilon$ at noise
levels $\sigma \in \{0, 0.05, 0.1, 0.2, 0.3, 0.5\}$ (50 samples per
nonzero $\sigma$; $s_{\mathrm{emp}}$ is the empirical latent standard
deviation), decode each $z'$, and check validity via RDKit
sanitisation and single-fragment connectivity.
Figure~\ref{fig:dark-area-embeddings} shows a 2-D PCA projection of
the pooled per-molecule mean latents, coloured by decoded validity.
TopVAE retains 90.5\% validity across all perturbation levels,
compared to 44.4\% for UAE and 32.6\% for ADiT, indicating that
TopoBridge and ChemCO substantially shrink the dark areas of the
latent space.

\subsection{Diffusion Denoising Trajectory}
\label{app:dark_area_diffusion}

To understand how the diffusion prior navigates the latent space
during generation, we track molecules through the full denoising
process.  At each diffusion timestep we decode the intermediate latent
code, check RDKit validity, and record the first principal component
of the per-molecule mean latent.

\begin{figure}[t]
\centering
\includegraphics[width=\linewidth]{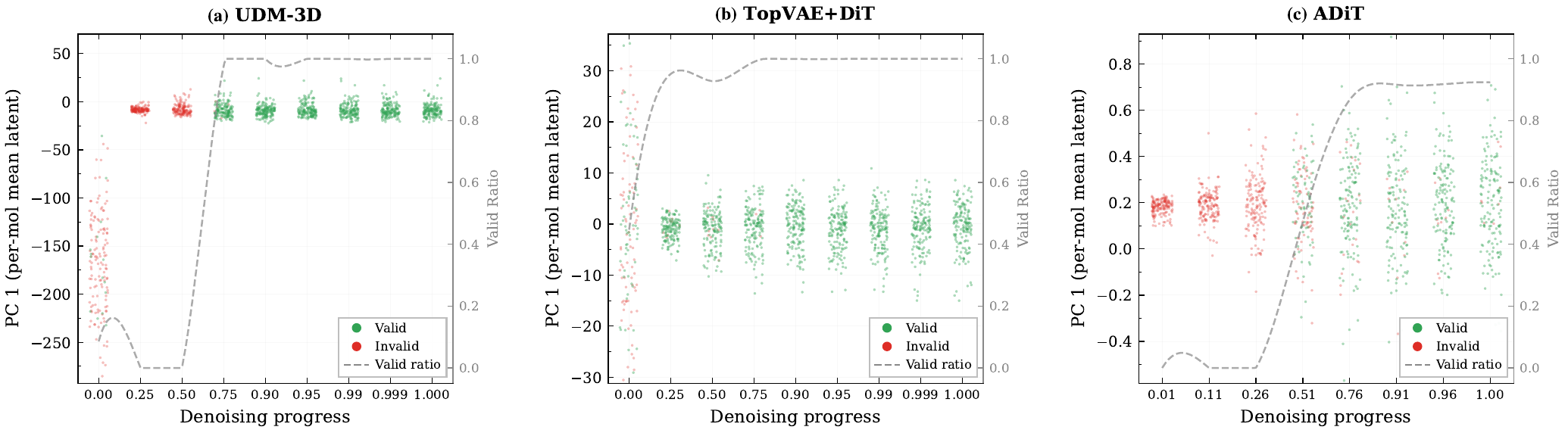}
\caption{\textbf{Latent-space trajectories during diffusion
denoising.}  Each dot represents a molecule at a specific denoising
step; colour indicates validity (green = valid, red = invalid).  The
dashed grey curve shows the batch validity ratio.
\textbf{(e)}~UDM-3D exhibits wide latent spread and delayed validity.
\textbf{(f)}~TopVAE+DiT shows compact trajectories with early
validity onset.
\textbf{(g)}~ADiT operates in a narrow latent range with a gradual
validity ramp.}
\label{fig:dark-area-diffusion}
\end{figure}

Figure~\ref{fig:dark-area-diffusion} plots these trajectories for
three models: UDM-3D, TopVAE+DiT, and ADiT.  The dashed grey curve
shows the instantaneous validity ratio across the batch.  Two
observations stand out:

\begin{enumerate}
\item \textbf{Latent spread.}  UDM-3D's intermediate latents span a
  PC\,1 range of ${\sim}300$ units early in denoising, reflecting
  high variance in the unconstrainted latent space.  TopVAE+DiT
  operates within a ${\sim}60$-unit range, and ADiT within
  ${\sim}1$ unit, indicating progressively tighter concentration
  on the data manifold.

\item \textbf{Validity transition.}  UDM-3D's validity ratio rises
  slowly and does not reach 80\% until denoising progress
  ${\sim}$0.75.  TopVAE+DiT crosses 90\% validity by progress 0.25,
  and its intermediate molecules are already chemically plausible at
  early denoising stages.  This early validity is a direct
  consequence of the TopoBridge connectivity guarantee and ChemCO
  valence constraints, which make the decoder robust to partially
  denoised (i.e., off-manifold) latent codes.
\end{enumerate}

\subsection{Molecule Trajectory Showcase}
\label{app:dark_area_showcase}

\begin{figure}[t]
\centering
\includegraphics[width=\linewidth]{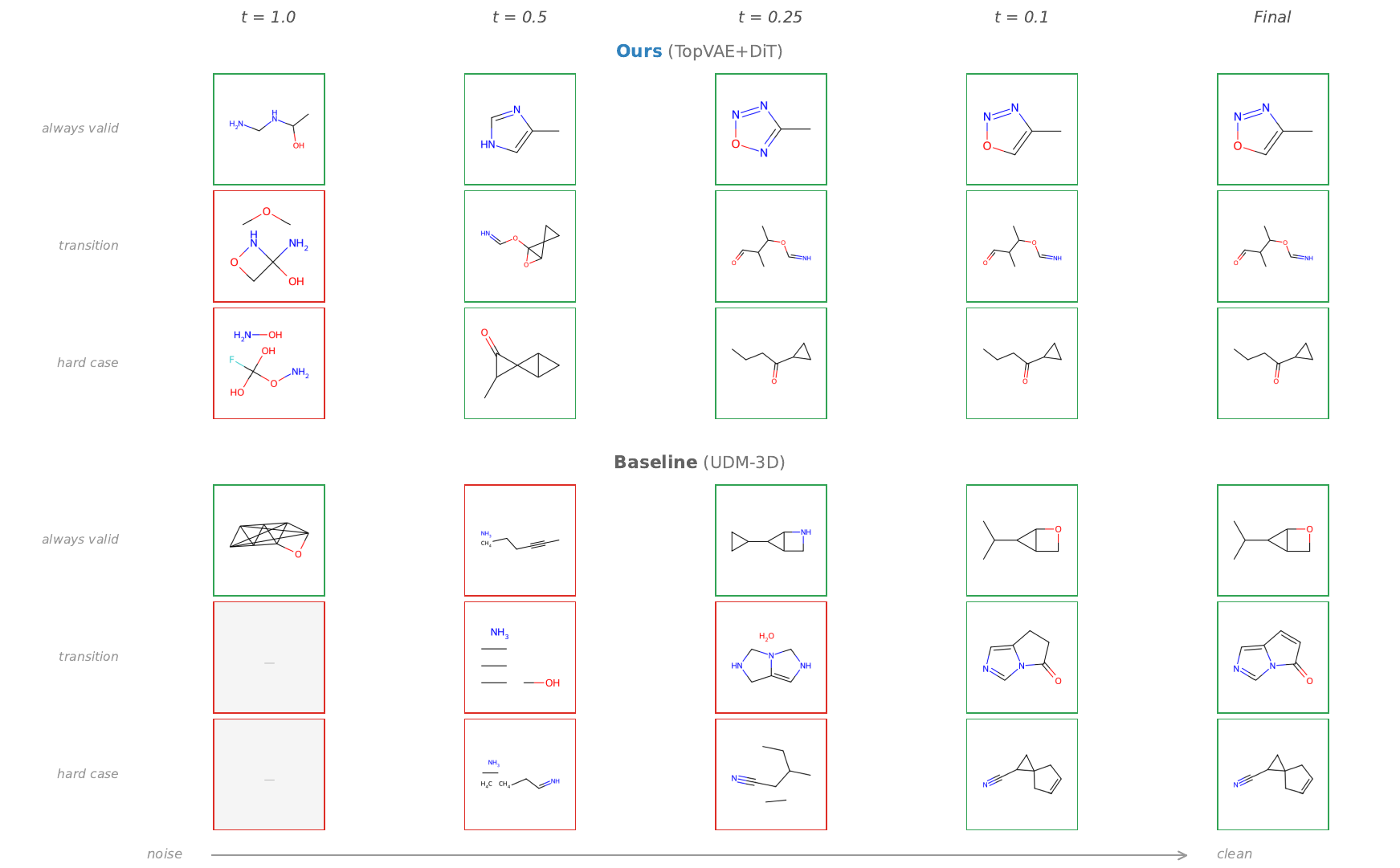}
\caption{\textbf{Denoising trajectories of representative molecules.}
Green/red borders indicate valid/invalid decoded molecules at each
timestep.  \textbf{Top:} TopVAE+DiT produces valid intermediates
earlier and more consistently.  \textbf{Bottom:} UDM-3D often
remains invalid until late denoising stages, and hard cases may
retain valence errors in the final output.}
\label{fig:dark-area-showcase}
\end{figure}

Figure~\ref{fig:dark-area-showcase} visualises representative
denoising trajectories for TopVAE+DiT and UDM-3D at five diffusion
timesteps ($t = 1.0, 0.5, 0.25, 0.1$, and the final output).
Molecules are grouped into three categories based on their validity
profile across the trajectory:

\begin{itemize}
\item \textbf{Always valid} (top rows): molecules that decode to
  valid structures at every sampled timestep.  TopVAE+DiT produces
  such trajectories more frequently, reflecting the decoder's
  built-in chemical constraints.

\item \textbf{Transition} (middle rows): molecules that start invalid
  at high noise ($t{=}1.0$) and become valid during denoising.
  TopVAE+DiT transitions earlier ($t{\sim}0.5$) while UDM-3D
  transitions later ($t{\sim}0.25$ or $0.1$), consistent with the
  validity curves in Figure~\ref{fig:dark-area-diffusion}.

\item \textbf{Hard case} (bottom rows): molecules that remain invalid
  for most of the trajectory before recovering at the final step,
  or that oscillate between valid and invalid states.  Even in hard
  cases, TopVAE+DiT recovers valid structures by the final step,
  whereas UDM-3D sometimes produces final molecules with residual
  valence violations or disconnected fragments (red borders).
\end{itemize}

Green borders indicate valid molecules; red borders indicate invalid
ones.  The visualisation confirms that TopVAE's decoder-level
constraints (TopoBridge + ChemCO) provide a robustness buffer that
allows the diffusion prior to produce valid molecules even from
partially denoised, off-manifold latent codes.

\clearpage
\section{3D Conformer Quality}\label{app:3d_conformer}

Tables~\ref{tab:3d_mmd_qm9} and~\ref{tab:3d_mmd_geom} report Maximum Mean Discrepancy (MMD) scores for bond lengths, bond angles, and dihedral angles between generated and reference molecules, following the \texttt{geom\_predictor} evaluation protocol used by JODO and EDM (bonds inferred from 3D coordinates).

\begin{table}[t]
\centering
\caption{\textbf{3D geometry MMD on QM9} ($n$=10{,}000). Lower is better. Baseline numbers are from respective papers.}
\label{tab:3d_mmd_qm9}
\begin{tabular}{lccc}
\toprule
\textbf{Method} & \textbf{Bond Len.\,$\downarrow$} & \textbf{Bond Ang.\,$\downarrow$} & \textbf{Dihedral\,$\downarrow$} \\
\midrule
EDM       & 1.30e-1 & 1.82e-2 & 6.64e-4 \\
GeoLDM   & 2.40e-1 & 1.00e-2 & 6.59e-4 \\
JODO     & 1.48e-1 & 1.21e-2 & 6.29e-4 \\
ADiT      & 9.98e-1 & 3.38e-2 & 1.46e-3 \\
UDM-3D    & 7.04e-2 & 9.84e-3 & \textbf{3.47e-4} \\
\midrule
TopVAE+DiT          & \textbf{6.65e-2} & \textbf{5.74e-3} & 4.83e-3 \\
\bottomrule
\end{tabular}
\end{table}

\begin{table}[t]
\centering
\caption{\textbf{3D geometry MMD on GEOM-Drugs} ($n$=10{,}000). Lower is better. Baseline numbers from respective papers.}
\label{tab:3d_mmd_geom}
\begin{tabular}{lccc}
\toprule
\textbf{Method} & \textbf{Bond Len.\,$\downarrow$} & \textbf{Bond Ang.\,$\downarrow$} & \textbf{Dihedral\,$\downarrow$} \\
\midrule
EDM       & 4.29e-1 & 4.96e-1 & 1.46e-2 \\
GeoLDM    & 3.91e-1 & 4.22e-1 & 1.69e-2 \\
JODO      & 8.49e-2 & 1.15e-2 & 6.68e-4 \\
UDM-3D    & \textbf{9.89e-3} & \textbf{5.11e-3} & \textbf{1.74e-4} \\
\midrule
TopVAE+DiT & 1.78e-2 & 1.58e-2 & 4.86e-3 \\
\bottomrule
\end{tabular}
\end{table}

\paragraph{Analysis.}
On QM9, TopVAE+DiT achieves the best bond-length MMD (6.65e-2 vs.\ 7.04e-2 for the next-best method UDM-3D, a 6\% reduction) and the best bond-angle MMD (5.74e-3 vs.\ 9.84e-3 for UDM-3D, a 42\% reduction). The dihedral MMD (4.83e-3) is weaker than UDM-3D's 3.47e-4; we attribute this to the topology-first decoder factorization, which prioritizes graph correctness over torsional accuracy---a trade-off that future work on coordinate refinement may address.

On GEOM-Drugs, TopVAE+DiT achieves bond-length MMD of 1.78e-2, on the same order of magnitude as UDM-3D (9.89e-3) and substantially better than EDM (24$\times$), GeoLDM (22$\times$), and JODO (4.8$\times$). This demonstrates that the topology-first decoder, despite lacking explicit E(3)-equivariant denoising in data space, produces competitive 3D geometries on drug-sized molecules.

\clearpage
\section{Molecular Property Distributions}\label{app:property_dist}

To assess whether TopVAE+DiT generates molecules that faithfully reproduce the training distribution beyond aggregate FCD scores, 
we compute 8 physicochemical descriptors on the generated set ($n$=10{,}000) and compare them to the reference.
\textit{UDM-3D's checkpoint was not released when we completed the manuscript; the results in this section are from our own re-training. They are for reference only and may not reflect UDM-3D's true performance.
}

\begin{table}[t]
\centering
\caption{\textbf{Property statistics on QM9} ($n$=10{,}000). Reference values computed on the QM9 test set. Bold indicates closer to reference.}
\label{tab:property_qm9}
\begin{tabular}{lccc}
\toprule
\textbf{Property} & \textbf{Dataset Ref.} & \textbf{UDM-3D} & \textbf{TopVAE+DiT} \\
\midrule
MW (Da)          & 122.6 & 122.4          & \textbf{122.6} \\
logP             & 0.35  & \textbf{0.43}  & 0.45 \\
QED              & 0.466 & 0.467          & \textbf{0.468} \\
HeavyAtoms       & 8.79  & 8.78           & \textbf{8.79} \\
Rings            & 1.74  & \textbf{1.87}  & 2.07 \\
Fsp3             & 0.709 & \textbf{0.750} & 0.778 \\
TPSA (\AA$^2$)   & 35.6  & \textbf{32.9}  & 31.3 \\
RotBonds         & 0.94  & \textbf{0.88}  & 0.70 \\
\midrule
Valid molecules  & ---   & 9{,}986        & \textbf{10{,}000} \\
\bottomrule
\end{tabular}
\end{table}

\begin{table}[t]
\centering
\caption{\textbf{Property statistics on GEOM-Drugs} ($n$=10{,}000). Reference values computed on the GEOM-Drugs training set. Bold indicates closer to reference.}
\label{tab:property_geom}
\begin{tabular}{lccc}
\toprule
\textbf{Property} & \textbf{Dataset Ref.} & \textbf{UDM-3D} & \textbf{TopVAE+DiT} \\
\midrule
MW (Da)          & 355.5 & 321.5          & \textbf{351.5} \\
logP             & 2.86  & 2.41           & \textbf{2.91} \\
QED              & 0.646 & 0.530          & \textbf{0.668} \\
HeavyAtoms       & 24.9  & 22.2           & \textbf{24.9} \\
Rings            & 3.00  & 2.22           & \textbf{3.27} \\
Fsp3             & 0.305 & 0.537          & \textbf{0.334} \\
TPSA (\AA$^2$)   & 73.9  & 67.2           & \textbf{68.2} \\
RotBonds         & 5.05  & \textbf{4.88}  & 4.75 \\
\midrule
Valid molecules  & ---   & 1{,}823 / 10{,}000 & \textbf{10{,}000} / 10{,}000 \\
\bottomrule
\end{tabular}
\end{table}

\paragraph{Analysis.}
On GEOM-Drugs, TopVAE+DiT matches the reference distribution more closely than UDM-3D on six of eight properties (all except rotatable bonds, where UDM-3D is slightly closer: 4.88 vs.\ 4.75 against a reference of 5.05). The improvement is particularly pronounced for molecular weight (1.1\% deviation vs.\ 9.6\%), heavy-atom count (exact match vs.\ 10.8\% undercount), and ring count (9\% deviation vs.\ 26\%), indicating that UDM-3D's UAE backbone tends to generate structurally simpler molecules on this larger-molecule benchmark. TopVAE+DiT also achieves 100\% validity (10{,}000/10{,}000), compared to only 1{,}823/10{,}000 for UDM-3D---a direct consequence of dark-area closure.

\begin{figure}[t]
\centering
\includegraphics[width=\textwidth]{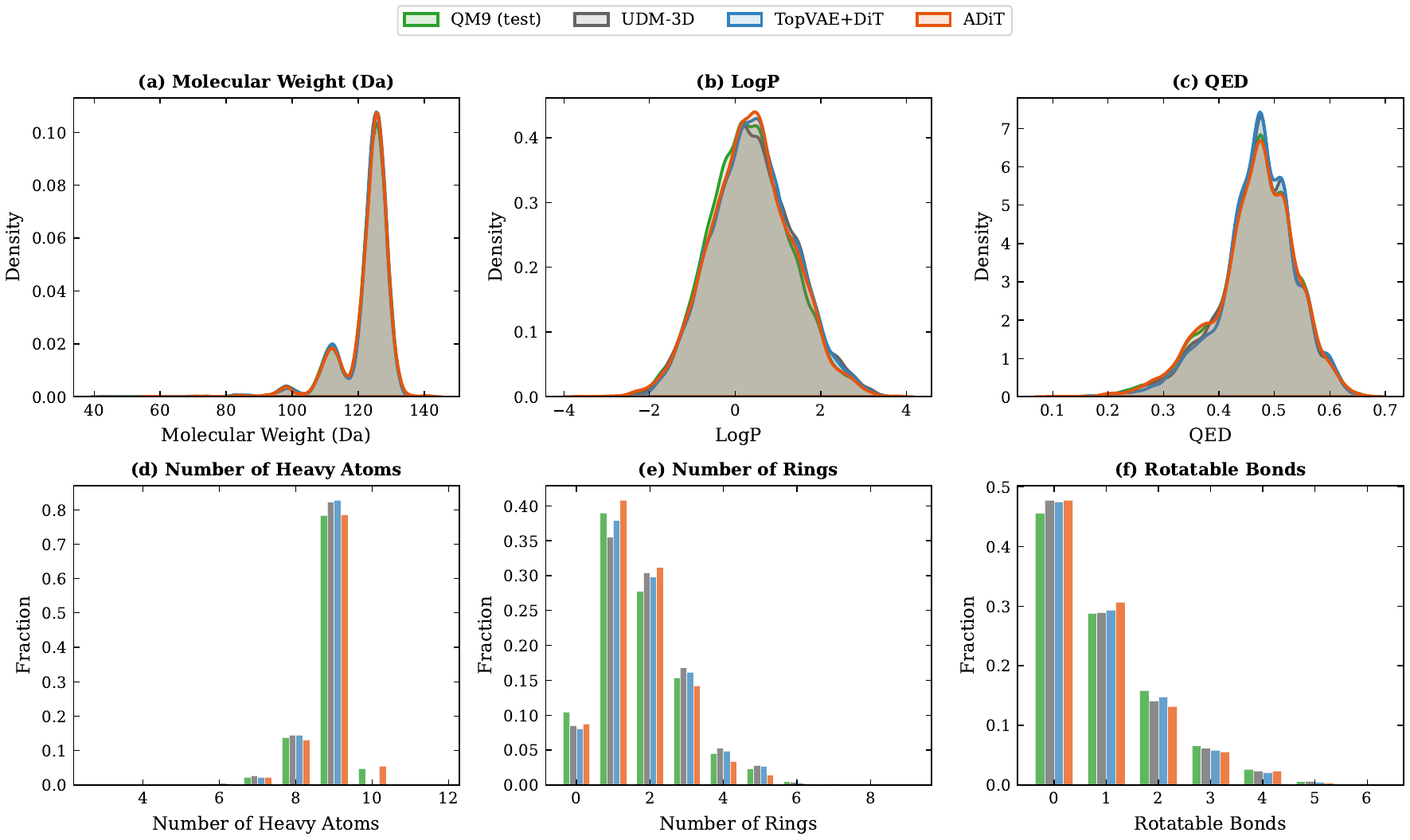}
\caption{\textbf{Property distributions on QM9} ($n$=10{,}000). Green: test-set reference; blue: TopVAE+DiT; gray: UDM-3D (re-trained); orange: ADiT. All three models closely match the reference distribution, consistent with QM9 being near-saturated for small molecules ($\leq$9 heavy atoms). Minor differences are visible in ring count and rotatable bonds, where TopVAE+DiT slightly overrepresents 2-ring structures.}
\label{fig:property_dist_qm9}
\end{figure}

\begin{figure}[t]
\centering
\includegraphics[width=\textwidth]{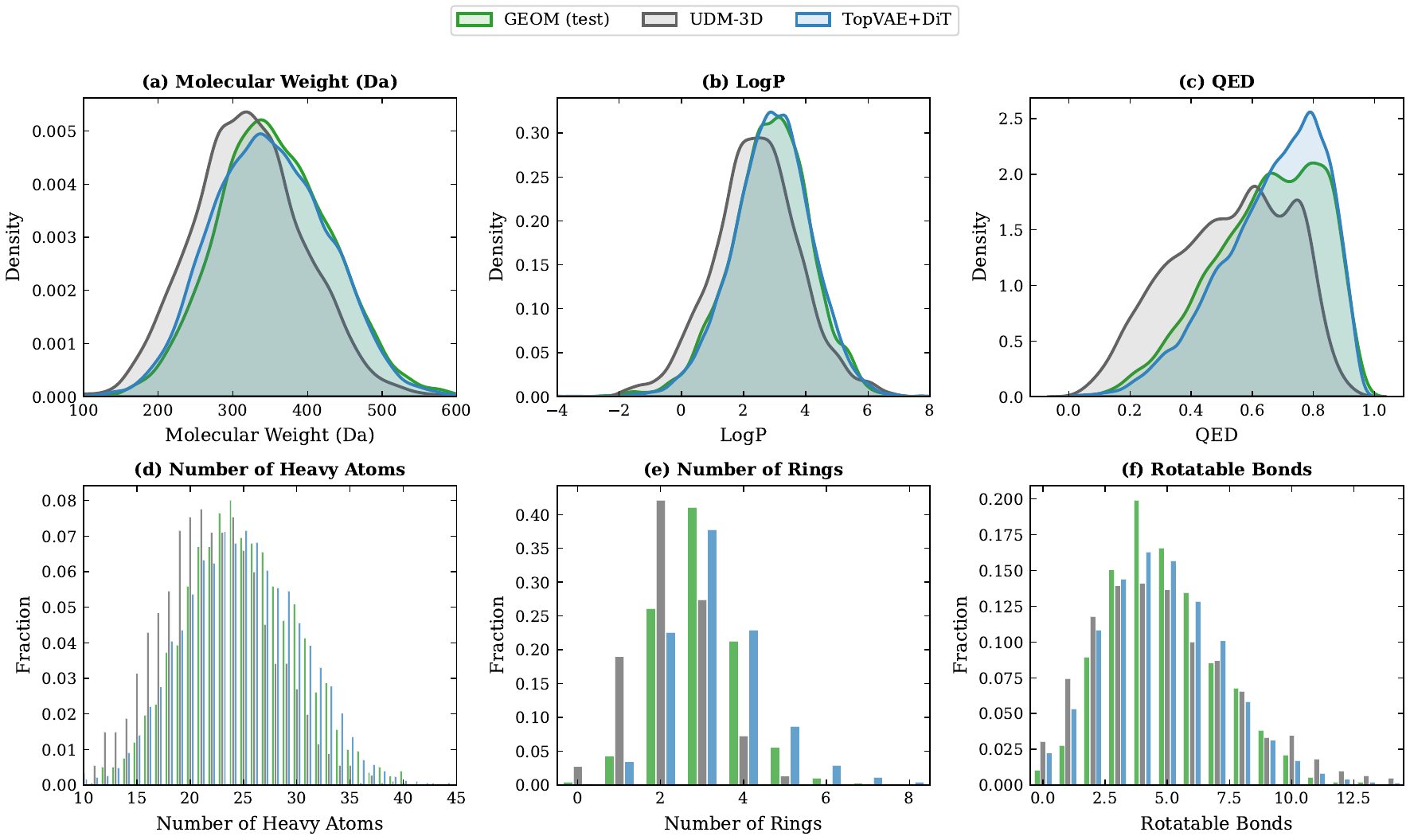}
\caption{\textbf{Property distributions on GEOM-Drugs} ($n$=10{,}000). Green: test-set reference; blue: TopVAE+DiT; gray: UDM-3D (re-trained). TopVAE+DiT closely tracks the reference across all eight properties. UDM-3D exhibits a systematic leftward shift in molecular weight and heavy-atom count, underrepresentation of 3--4 ring systems, and a bimodal QED distribution---consistent with its tendency to decode structurally simpler molecules due to dark-area failures.}
\label{fig:property_dist_geom}
\end{figure}

On QM9, both models produce near-identical property statistics, consistent with the benchmark being saturated for small molecules ($\leq$9 heavy atoms). UDM-3D is closer to the reference on four properties (logP, Rings, Fsp3, TPSA, RotBonds), while TopVAE+DiT matches the reference more closely on MW, QED, and HeavyAtoms. TopVAE+DiT achieves perfect validity (10{,}000 vs.\ 9{,}986).
Figures~\ref{fig:property_dist_qm9} and~\ref{fig:property_dist_geom} visualize the full property distributions underlying the summary statistics in Tables~\ref{tab:property_geom} and~\ref{tab:property_qm9}.

\clearpage
\section{Diversity and Novelty}\label{app:diversity}

To verify that dark-area closure does not reduce structural diversity, Tables~\ref{tab:diversity_geom} and~\ref{tab:diversity_qm9} report diversity and novelty metrics on both datasets. Internal diversity is computed as $1 - \overline{\mathrm{Tanimoto}}$ over Morgan fingerprints (radius 2, 2048 bits, up to 500k sampled pairs). 
Novelty is the fraction of unique SMILES absent from the training set. 
Unique scaffolds are Murcko scaffolds of unique molecules. 
Since UDM-3D has not released pretrained checkpoints, \textbf{all UDM-3D results in this section are re-trained using the official UDM-3D GitHub repository and default configurations.} The results of UDM-3D are for reference only.

\begin{table}[t]
\centering
\caption{\textbf{Diversity and novelty on QM9} ($n$=10{,}000). \textbf{UDM-3D is re-trained using the official UDM-3D repository} (for reference only). QM9 contains $\leq$9 heavy atoms; novelty and unique scaffold rates are naturally lower than on GEOM-Drugs because the training set already covers most of the reachable chemical space at this size.}
\label{tab:diversity_qm9}
\begin{tabular}{lcc}
\toprule
\textbf{Metric} & \textbf{TopVAE+DiT} & \textbf{UDM-3D} \\
\midrule
Valid molecules                      & \textbf{10{,}000} (100\%) & 9{,}986 (99.9\%) \\
FCD (2D) $\downarrow$                & \textbf{0.185}            & 0.207 \\
Internal diversity $\uparrow$        & 0.914            & \textbf{0.916} \\
Novelty $\uparrow$                   & \textbf{45.1\%}  & 42.3\% \\
Unique Murcko scaffolds $\uparrow$   & \textbf{36.3\%}  & 29.6\% \\
Uniqueness $\uparrow$                & 95.7\%           & \textbf{96.6\%} \\
\midrule
Unique SMILES                        & 9{,}571          & 9{,}660 \\
Unique scaffold count                & \textbf{3{,}477} & 2{,}864 \\
\bottomrule
\end{tabular}
\end{table}

\begin{table}[t]
\centering
\caption{\textbf{Diversity and novelty on GEOM-Drugs} ($n$=10{,}000 sampled). \textbf{UDM-3D is re-trained using the official UDM-3D repository} (for reference only). UDM-3D produces only 1{,}823 valid molecules (18.2\%); its diversity and novelty metrics are computed on this smaller set, which naturally inflates per-molecule diversity and novelty.}
\label{tab:diversity_geom}
\begin{tabular}{lcc}
\toprule
\textbf{Metric} & \textbf{TopVAE+DiT} & \textbf{UDM-3D} \\
\midrule
Valid molecules                       & \textbf{10{,}000} (100\%)  & 1{,}823 (18.2\%) \\
FCD (2D) $\downarrow$                & \textbf{2.68}              & 25.73 \\
Internal diversity $\uparrow$        & 0.874                      & 0.898$^*$ \\
Novelty $\uparrow$                   & 99.95\%                    & 100\%$^*$ \\
Unique Murcko scaffolds $\uparrow$   & \textbf{88.7\%}            & 85.4\%$^*$ \\
Uniqueness $\uparrow$                & 99.99\%                    & 100\%$^*$ \\
\bottomrule
\multicolumn{3}{l}{\footnotesize $^*$Computed on 1{,}823 valid molecules; small-sample sizes inflate these metrics.}
\end{tabular}
\end{table}

\paragraph{QM9 analysis.}
On QM9, internal diversity is effectively tied (0.914 vs.\ 0.916), consistent with the ${\sim}$0.91 range reported by EDM and GeoLDM on this dataset. TopVAE+DiT explores a substantially broader scaffold space, producing 3{,}477 unique Murcko scaffolds versus 2{,}864 for UDM-3D---a 21\% increase---and achieves higher novelty (45.1\% vs.\ 42.3\%). TopVAE+DiT also attains a lower FCD (0.185 vs.\ 0.207), indicating a generated distribution that better matches the training set. UDM-3D has slightly higher uniqueness (96.6\% vs.\ 95.7\%), but TopVAE+DiT's advantage in scaffold coverage and novelty suggests it explores a broader region of chemical space.

\paragraph{GEOM-Drugs analysis.}
TopVAE+DiT achieves high internal diversity (0.874), near-perfect novelty (99.95\%), and 88.7\% unique Murcko scaffolds across all 10{,}000 generated molecules, confirming the absence of mode collapse. UDM-3D shows slightly higher internal diversity (0.898) and novelty (100\%), but these metrics are computed on only 1{,}823 valid molecules---the remaining 81.8\% fail RDKit sanitization due to dark-area decoding failures. The 10$\times$ difference in FCD (2.68 vs.\ 25.73) reflects the severe distributional shift caused by dark areas in the UAE latent space: the few valid molecules that survive sanitization are not representative of the target distribution. TopVAE+DiT achieves a higher unique scaffold rate (88.7\% vs.\ 85.4\%) despite evaluating over $5\times$ more molecules, indicating genuine structural diversity rather than small-sample artifacts.

\clearpage
\section{Energy Relaxation Evaluation}
\label{app:energy_relaxation}
 
To assess whether TopVAE's topological improvements translate into
more physically plausible 3D geometries, we conduct an MMFF94
energy-relaxation study on GEOM-Drugs. For each generated molecule,
we compute the MMFF94 single-point energy of the generated conformer,
run a local geometry optimization to the nearest force-field minimum,
and report (i) the relaxation energy
$\Delta E = E_{\text{generated}} - E_{\text{relaxed}}$ (lower indicates
the generated geometry is closer to equilibrium) and (ii) the
heavy-atom RMSD between the generated and relaxed conformers.
 
\paragraph{Results.}
Table~\ref{tab:mmff_energy} reports MMFF94 relaxation statistics for
TopVAE+DiT, UDM-3D, and reference molecules from the GEOM-Drugs
test set ($n{=}2{,}000$ per source). TopVAE+DiT generates 3D
geometries substantially closer to force-field minima than UDM-3D:
the median relaxation energy is $3.9{\times}$ lower
(19.8 vs.\ 76.8\,kcal/mol), and the median RMSD is 12\% lower
(0.99 vs.\ 1.12\,\AA). Both generative models are expectedly farther
from equilibrium than the reference dataset (median
$\Delta E = 13.9$\,kcal/mol), but TopVAE is considerably closer.
 
These results suggest that enforcing topological correctness via
TopoBridge and ChemCO provides an indirect benefit to 3D geometric
quality: a chemically consistent bond graph constrains the coordinate
head and EGNN to produce conformers in more physically reasonable
regions of configuration space.
 
\begin{table}[t]
\centering
\caption{MMFF94 energy relaxation on GEOM-Drugs ($n{=}2{,}000$).
$\Delta E$ is the energy difference between the generated and
locally optimized conformer (lower is better). RMSD is the
heavy-atom root-mean-square deviation between the two conformers.}
\label{tab:mmff_energy}
\begin{tabular}{lccc}
\toprule
\textbf{Metric}
  & \textbf{Reference}
  & \textbf{TopVAE+DiT}
  & \textbf{UDM-3D} \\
\midrule
$\Delta E$ mean (kcal/mol) $\downarrow$
  & 16.5 & \textbf{40.6} & 104.4 \\
$\Delta E$ median (kcal/mol) $\downarrow$
  & 13.9 & \textbf{19.8} & 76.8 \\
RMSD mean (\AA) $\downarrow$
  & 0.65 & \textbf{1.14} & 1.24 \\
RMSD median (\AA) $\downarrow$
  & 0.51 & \textbf{0.99} & 1.12 \\
\bottomrule
\end{tabular}
\end{table}

\clearpage
\section{Hyperparameter Sensitivity Analysis}
\label{app:sensitivity}

We study the sensitivity of TopVAE to two key design choices:
TopoBridge hyperparameters (adjacency sparsity, minimum degree) and
the $\Omega^{(k)}$ chemical mask.
All runs use QM9 with 500 training epochs and posterior evaluation
($n{=}10\text{k}$); absolute numbers may shift with full convergence
(2000 epochs), but cross-run rankings are expected to hold.

\subsection{TopoBridge Hyperparameters}
\label{app:topobridge_sensitivity}

Table~\ref{tab:topobridge_sensitivity} varies the adjacency sparsity
(\texttt{adj\_max\_deg}, i.e., the top-$k$ neighbor count) and the
minimum-degree penalty ($d_{\min}$) in ChemCO.

\begin{table}[t]
\centering
\caption{\textbf{TopoBridge hyperparameter sensitivity} on QM9
(500 epochs, $n{=}10\text{k}$ posterior).
Default: \texttt{adj\_max\_deg}$=4$, $d_{\min}{=}1$.}
\label{tab:topobridge_sensitivity}
\footnotesize
\begin{tabular}{@{}llcccc@{}}
\toprule
Run & Change vs.\ default
  & Valid.$\uparrow$
  & Uniq.$\uparrow$
  & \makecell{3D\\S$\wedge$C$\uparrow$}
  & \makecell{3D Atom\\Stab.$\uparrow$} \\
\midrule
\texttt{topk\_3}
  & \texttt{adj\_max\_deg}$=3$ (sparse)
  & 1.000 & 0.916 & 0.0146 & 0.689 \\
\texttt{topk\_8}
  & \texttt{adj\_max\_deg}$=8$ (dense)
  & 1.000 & 0.925 & 0.0160 & 0.682 \\
\texttt{dmin\_0}
  & $d_{\min}{=}0$ (no anti-degeneracy)
  & 1.000 & 0.922 & \textbf{0.0435} & \textbf{0.722} \\
\bottomrule
\end{tabular}
\end{table}

\paragraph{Findings.}
\begin{enumerate}[leftmargin=1.5em,itemsep=2pt]
\item \textbf{Robust to adjacency sparsity.}
    Varying \texttt{adj\_max\_deg} from 3 to 8 produces nearly identical
    metrics (3D S$\wedge$C: 0.0146 vs.\ 0.0160). ChemCO compensates for
    sparser adjacency by reallocating bond mass under valence constraints.
\item \textbf{Per-dataset tuning of $d_{\min}$.}
    Removing the minimum-degree penalty ($d_{\min}{=}0$) yields the highest
    3D S$\wedge$C on QM9 ($2.7{\times}$ over dense $topk\_8$ setting), suggesting that the
    $d_{\min}$ term over-regularizes on small molecules (${\leq}9$ heavy atoms).
    The term was designed for GEOM-Drugs, where bond collapse is the dominant
    failure mode; this result justifies per-dataset hyperparameter selection.
\end{enumerate}

\subsection{\texorpdfstring{$\Omega^{(k)}$ Chemical Mask Sensitivity}{Omega(k) Chemical Mask Sensitivity}}
\label{app:omega_sensitivity}

Table~\ref{tab:omega_sensitivity} compares the default QM9 configuration
(valence caps only, $\Omega^{(k)}\!=\!\bm{1}$) with the GEOM-style explicit
pair mask (Appendix~\ref{app:omega_mask}).

\begin{table}[t]
\centering
\caption{\textbf{$\Omega^{(k)}$ mask sensitivity} on QM9
(500 epochs, $n{=}10\text{k}$ posterior).}
\label{tab:omega_sensitivity}
\footnotesize
\begin{tabular}{@{}lcccc@{}}
\toprule
$\Omega^{(k)}$ config
  & Valid.$\uparrow$
  & Uniq.$\uparrow$
  & \makecell{3D\\S$\wedge$C$\uparrow$}
  & \makecell{3D Atom\\Stab.$\uparrow$} \\
\midrule
Valence caps only (default)
  & 1.000 & 0.925 & 0.0160 & 0.682 \\
+ Explicit pair restrictions
  & 1.000 & 0.920 & \textbf{0.0191} & \textbf{0.688} \\
\bottomrule
\end{tabular}
\end{table}

Adding the explicit chemical pair mask yields a 19\% relative gain in
3D S$\wedge$C (0.0160$\to$0.0191) with no loss in validity or uniqueness.
The gain is mild on QM9 (only 4 atom types) and is expected to be larger
on datasets with more diverse atom-type combinations.
This validates that $\Omega^{(k)}$ captures pair-specific chemical knowledge
that valence caps alone cannot encode.

\clearpage
\section{VAE Posterior Reconstruction Quality}
\label{app:vae-posterior}

Tables~\ref{tab:qm9-posterior} and~\ref{tab:geom-posterior} report VAE posterior reconstruction quality on QM9 and GEOM-Drugs respectively.
These results complement the dark-area diagnosis in Section~\ref{sec:dark-area-diagnosis} by showing that high posterior MolStab does not guarantee high Stab$\wedge$Conn: on GEOM-Drugs, UAE achieves MolStab${=}0.998$ but Stab$\wedge$Conn${=}0.000$, as every reconstructed molecule fragments into disconnected atoms.

\begin{table}[ht]
\centering
\caption{\textbf{VAE posterior reconstruction on QM9} ($n{=}10\text{k}$ validation set, \ourmark).}
\label{tab:qm9-posterior}
\footnotesize
\setlength{\tabcolsep}{3.5pt}
\begin{tabular}{@{}l*{6}{c}@{}}
\toprule
Method
  & FCD$\downarrow$
  & \makecell{Atom\\Stab.$\uparrow$}
  & \makecell{Mol\\Stab.$\uparrow$}
  & V\&C$\uparrow$
  & FCD\textsubscript{3D}$\downarrow$
  & \makecell{Mol\\Stab.\textsubscript{3D}$\uparrow$} \\
\midrule
UAE               & 0.100 & 1.000 & 1.000 & 0.999 & 24.17 & 0.000 \\
ADiT-VAE         & 0.795 & 1.000 & 0.951 & 0.904 & \textbf{0.898} & \textbf{0.900} \\
\rowcolor{rowhl}
\textbf{\method{}} & \textbf{0.089} & 1.000 & \textbf{1.000} & \textbf{1.000} & 15.39 & 0.002 \\
\bottomrule
\end{tabular}
\end{table}

\begin{table}[ht]
  \centering
  \caption{\textbf{VAE posterior reconstruction on GEOM-Drugs} ($n{=}1997$ validation set, \ourmark).
  UAE achieves near-perfect MolStab but zero connectivity---the atom-cloud failure mode discussed in Section~\ref{sec:dark-area-diagnosis}.
  \method{} numbers reported with the same greedy valence repair used by \method{}+DiT.}
  \label{tab:geom-posterior}
  \footnotesize
  \setlength{\tabcolsep}{3.5pt}
  \begin{tabular}{@{}l*{6}{c}@{}}
  \toprule
  Method
    & FCD$\downarrow$
    & \makecell{Atom\\Stab.$\uparrow$}
    & \makecell{Mol\\Stab.$\uparrow$}
    & V\&C$\uparrow$
    & FCD\textsubscript{3D}$\downarrow$
    & \makecell{Mol\\Stab.\textsubscript{3D}$\uparrow$} \\
  \midrule
  UAE               & 60.67 & 1.000 & \textbf{0.998} & 0.000 & 44.28 & \textbf{0.002} \\
  \rowcolor{rowhl}
  \textbf{\method{}} & \textbf{46.17} & \textbf{0.999} & 0.960 & \textbf{0.998} & 44.28 & 0.000 \\
  \bottomrule
  \end{tabular}
  \end{table}

\section{Scaffold Inpainting: Per-Scaffold Results}
\label{app:inpaint}

\begin{figure}[t]
\centering
\includegraphics[width=\linewidth, page=1]{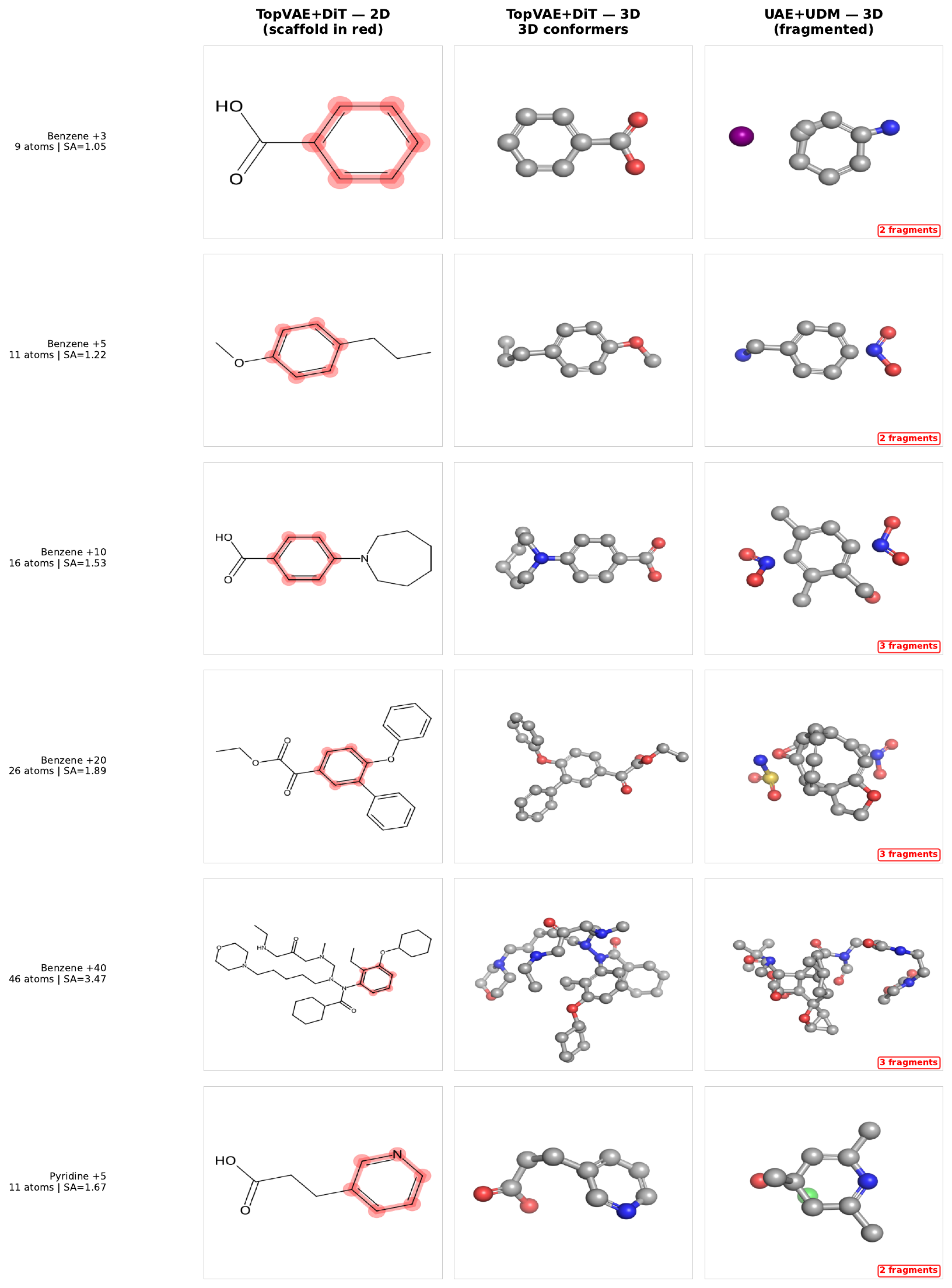}
\caption{Scaffold Inpainting (Part 1).}
\label{fig:scaffold_showcase_1}
\end{figure}

Table~\ref{tab:geom_inpaint_detail} provides the per-scaffold breakdown of the GEOM-Drugs inpainting results summarized in Table~\ref{tab:geom_inpaint}.
All experiments use differential inpainting with noise scale${=}0.3$, 1k samples per cell.
The advantage of \method{} is consistent across all five scaffolds, with cyclohexane (saturated, non-aromatic) being 
the easiest and naphthalene/indole (fused aromatic rings) the most challenging for scaffold preservation (Figure ~\ref{fig:scaffold_showcase_1}, ~\ref{fig:scaffold_showcase_2}).

\begin{table}[t]
\centering
\caption{\textbf{GEOM-Drugs scaffold inpainting: per-scaffold breakdown.}
1k samples per cell, differential inpainting (noise scale${=}0.3$).
Values in \%. Bold = better per cell.}
\label{tab:geom_inpaint_detail}
\scriptsize
\setlength{\tabcolsep}{3pt}
\begin{tabular}{@{} c l r cc cc @{}}
\toprule
& & & \multicolumn{2}{c}{\textbf{Scaff.\ Pres.}\,$\uparrow$}
& \multicolumn{2}{c}{\textbf{S$\wedge$C}\,$\uparrow$} \\
\cmidrule(lr){4-5} \cmidrule(l){6-7}
Scaffold & $+n$ & $N$ & UAE & \method{} & UAE & \method{} \\
\midrule
\multirow{7}{*}{\fbox{\includegraphics[height=36pt]{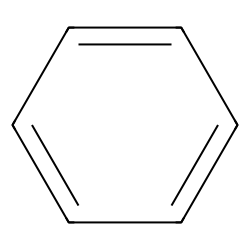}}}
 & 5 & 11 & \textbf{7.0} & 4.0 & 95.6 & \textbf{97.2} \\
 & 10 & 16 & 8.4 & \textbf{9.2} & 89.0 & \textbf{94.8} \\
 & 20 & 26 & 7.4 & \textbf{21.2} & 86.6 & \textbf{87.6} \\
 & 40 & 46 & 28.4 & \textbf{60.8} & 62.2 & \textbf{70.2} \\
 & 60 & 66 & 0.6 & \textbf{50.2} & 1.6 & \textbf{51.2} \\
 & 80 & 86 & 0.0 & \textbf{34.8} & 0.0 & \textbf{25.2} \\
 & 100 & 106 & 0.0 & \textbf{16.4} & 0.0 & \textbf{9.6} \\
\midrule
\multirow{7}{*}{\fbox{\includegraphics[height=36pt]{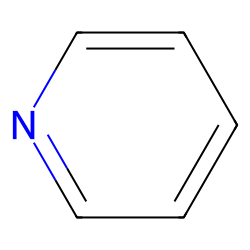}}}
 & 5 & 11 & 3.4 & 3.4 & 95.6 & \textbf{97.6} \\
 & 10 & 16 & 4.0 & \textbf{5.2} & 90.2 & \textbf{93.6} \\
 & 20 & 26 & 2.6 & \textbf{7.8} & \textbf{87.8} & 87.2 \\
 & 40 & 46 & 2.8 & \textbf{11.0} & 69.4 & \textbf{76.0} \\
 & 60 & 66 & 0.0 & \textbf{11.2} & 2.0 & \textbf{53.6} \\
 & 80 & 86 & 0.0 & \textbf{6.8} & 0.0 & \textbf{29.4} \\
 & 100 & 106 & 0.0 & \textbf{2.8} & 0.0 & \textbf{8.8} \\
\midrule
\multirow{7}{*}{\fbox{\includegraphics[height=36pt]{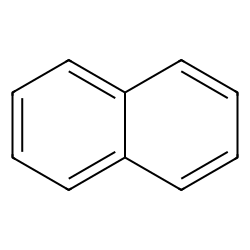}}}
 & 5 & 15 & \textbf{1.8} & 1.6 & 81.2 & \textbf{92.0} \\
 & 10 & 20 & 1.0 & \textbf{1.2} & 79.6 & \textbf{89.4} \\
 & 20 & 30 & 0.0 & \textbf{1.6} & 85.0 & \textbf{85.8} \\
 & 40 & 50 & 0.0 & \textbf{3.0} & 46.2 & \textbf{68.2} \\
 & 60 & 70 & 0.0 & \textbf{3.0} & 0.4 & \textbf{52.6} \\
 & 80 & 90 & 0.0 & \textbf{2.4} & 0.0 & \textbf{31.0} \\
 & 100 & 110 & 0.0 & \textbf{0.4} & 0.0 & \textbf{8.2} \\
\midrule
\multirow{7}{*}{\fbox{\includegraphics[height=36pt]{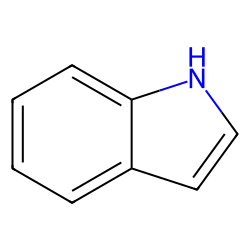}}}
 & 5 & 14 & \textbf{2.4} & 2.0 & 79.6 & \textbf{95.0} \\
 & 10 & 19 & 3.2 & 3.2 & 76.4 & \textbf{85.4} \\
 & 20 & 29 & 0.2 & \textbf{2.4} & \textbf{84.6} & 84.4 \\
 & 40 & 49 & 0.0 & \textbf{6.4} & 66.6 & \textbf{69.0} \\
 & 60 & 69 & 0.0 & \textbf{9.0} & 1.8 & \textbf{48.4} \\
 & 80 & 89 & 0.0 & \textbf{6.4} & 0.0 & \textbf{25.0} \\
 & 100 & 109 & 0.0 & \textbf{2.8} & 0.0 & \textbf{9.4} \\
\midrule
\multirow{7}{*}{\fbox{\includegraphics[height=36pt]{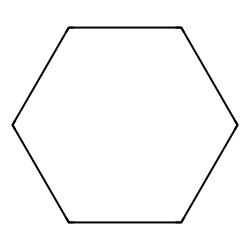}}}
 & 5 & 11 & 35.2 & \textbf{45.6} & 97.0 & \textbf{97.8} \\
 & 10 & 16 & 57.2 & \textbf{58.0} & \textbf{95.4} & 92.2 \\
 & 20 & 26 & \textbf{65.6} & 46.6 & \textbf{90.4} & 79.2 \\
 & 40 & 46 & \textbf{58.4} & 57.8 & \textbf{75.0} & 64.8 \\
 & 60 & 66 & 24.2 & \textbf{71.2} & 22.2 & \textbf{51.2} \\
 & 80 & 86 & 0.0 & \textbf{75.8} & 0.4 & \textbf{21.2} \\
 & 100 & 106 & 0.0 & \textbf{76.6} & 0.0 & \textbf{7.2} \\
\midrule
\rowcolor{rowhl}
& \textbf{Mean} & & 9.0 & \textbf{20.6} & 47.5 & \textbf{61.1} \\
\bottomrule
\end{tabular}
\end{table}

\begin{figure}[t]
\centering
\includegraphics[width=0.9\linewidth]{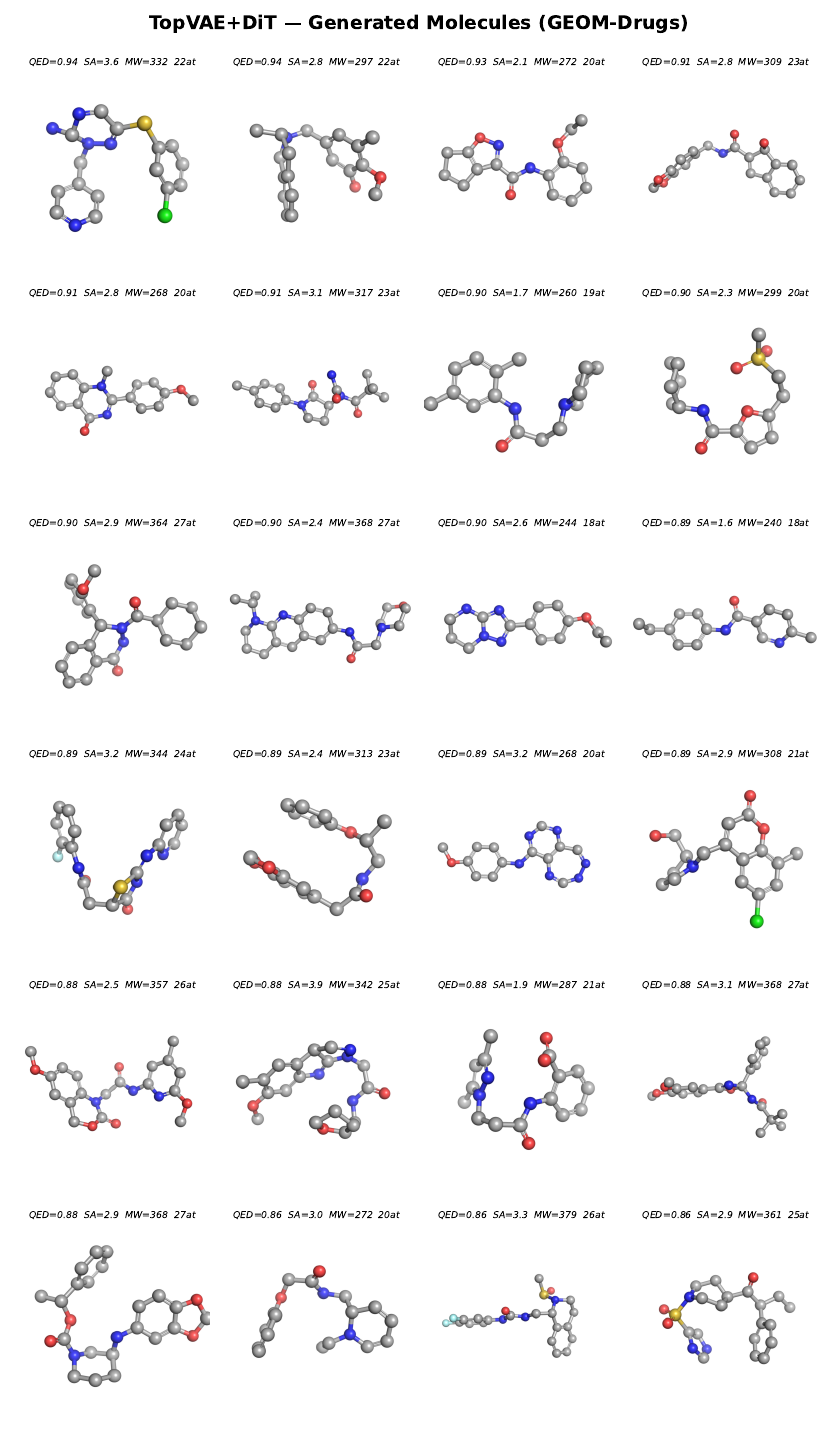}
\caption{3D Generation Molecules.}
\label{fig:showcase}
\end{figure}

\begin{figure}[t]
\centering
\includegraphics[width=\linewidth, page=2]{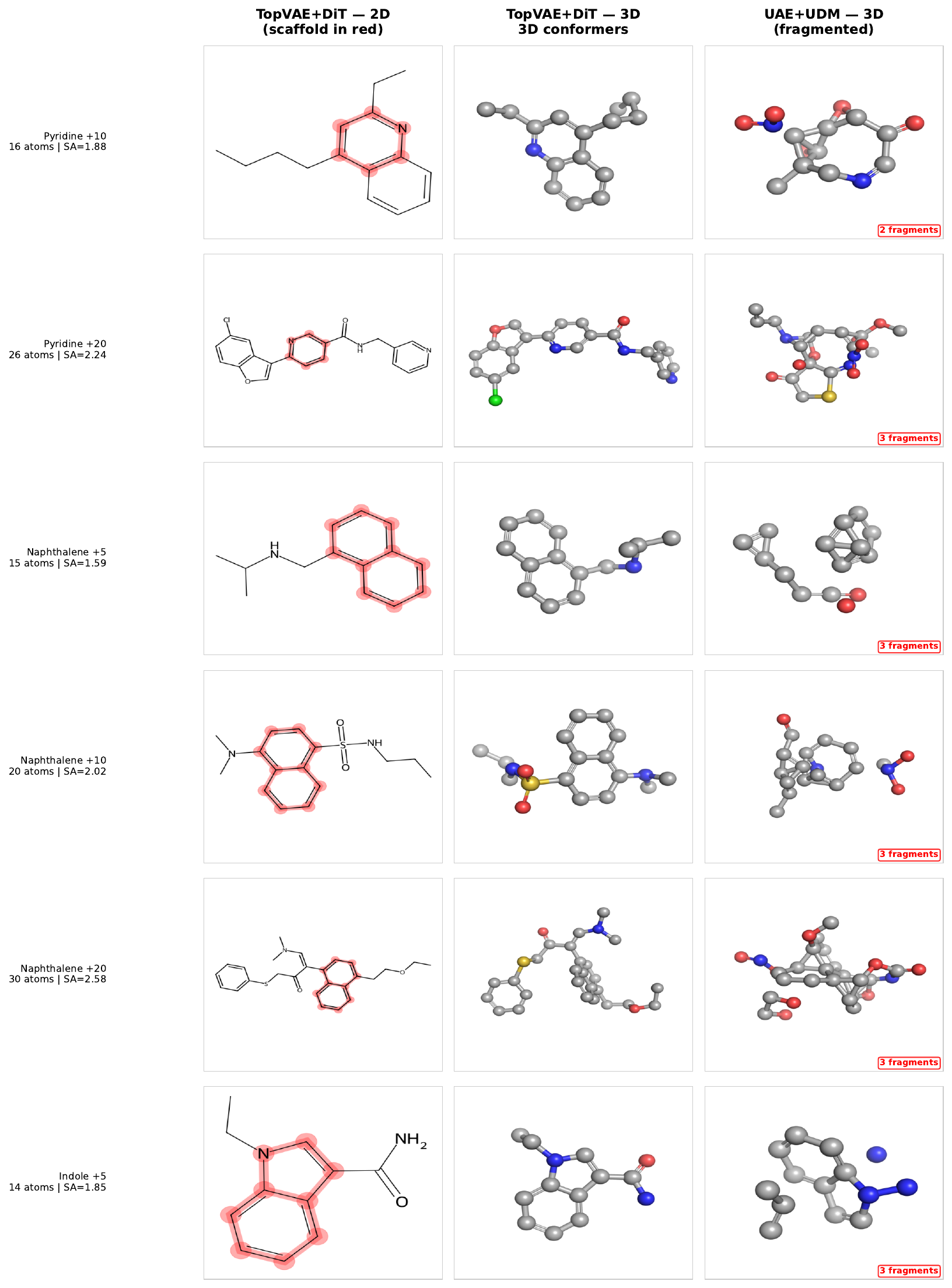}
\caption{Scaffold Inpainting (Part 2).}
\label{fig:scaffold_showcase_2}
\end{figure}

\begin{figure}[t]
\centering
\includegraphics[width=\linewidth, page=3]{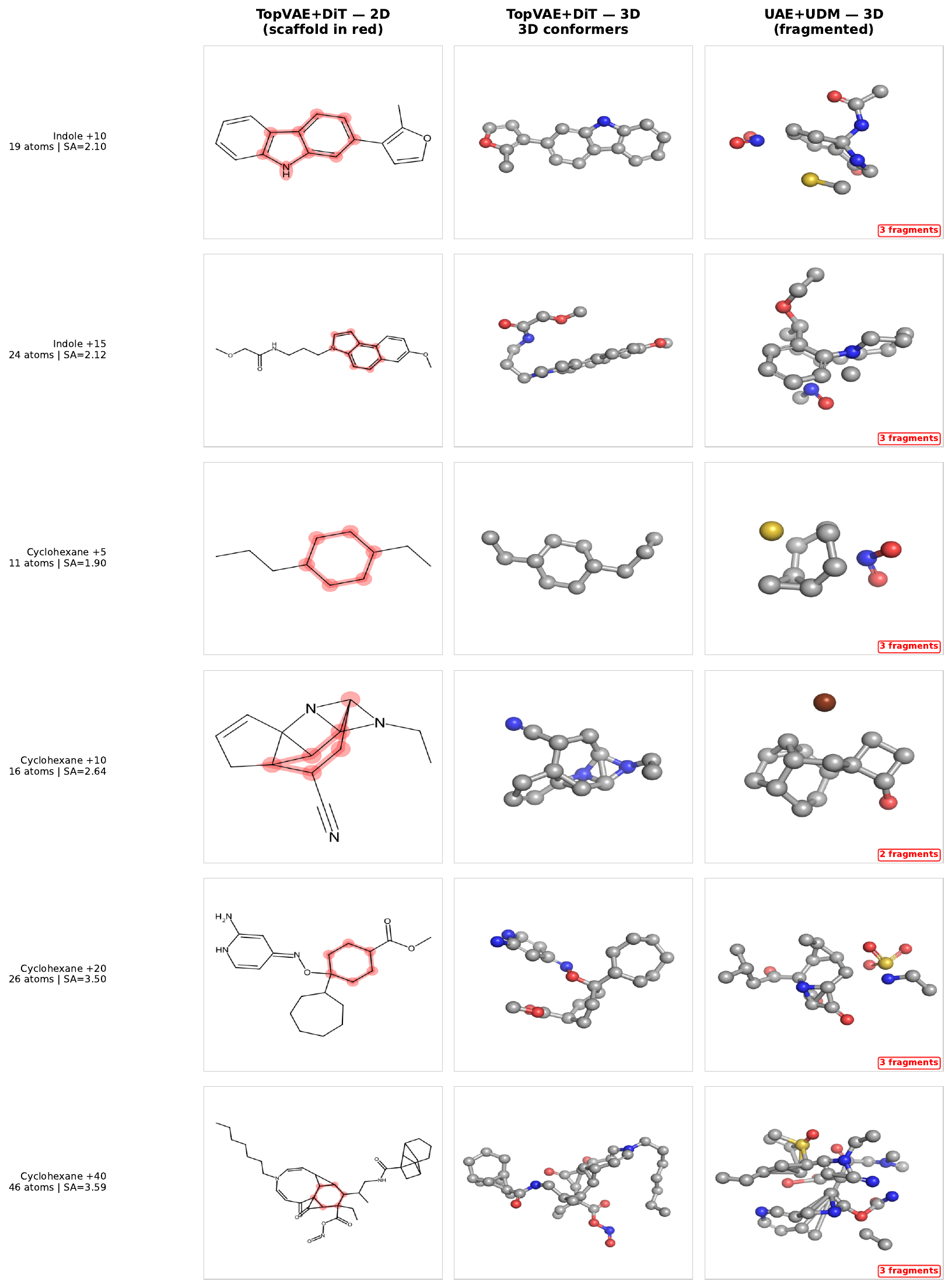}
\caption{Scaffold Inpainting (Part 3).}
\label{fig:scaffold_showcase_2}
\end{figure}


\clearpage
\section{TopoBridge Chemical Plausibility Analysis}
\label{app:bridge_audit}

TopoBridge guarantees graph connectivity by adding repair edges between
disconnected components through greedy BFS refinement
(Eq.~\ref{eq:topobridge-bridge}--\ref{eq:topobridge-ste}).  To isolate the
effect of BFS connectivity repair from the decoder's own predictions, we
instrument the inference pipeline to track \emph{BFS-inserted edges}: edges
that did not exist after initial thresholding/top-$k$ of the adjacency logits
and were explicitly added by the BFS repair procedure (isolated-node fix or
component-merging) to ensure connectivity.  Concretely, we snapshot the
adjacency matrix $\mathbf{A}_{\mathrm{init}}$ immediately after
thresholding and before any repair, then identify inserted edges via
$\mathbf{M}_{\mathrm{ins}} = \mathrm{clamp}(\mathbf{A}_{\mathrm{final}}
- \mathbf{A}_{\mathrm{init}},\, \min{=}0)$.
This differs from the broader set of graph-theoretic bridge bonds (cut edges
found by Tarjan's algorithm), most of which originate from the decoder's
adjacency head and are not artifacts of repair.

\paragraph{BFS-inserted edge chemistry.}
Table~\ref{tab:bridge_audit} summarizes the chemical properties of
BFS-inserted edges in TopVAE+DiT generations on GEOM-Drugs.  Among 10{,}000
generated molecules, only 2.1\% (210 molecules) receive any BFS-inserted edge,
totaling 211 inserted edges out of approximately 270{,}000 total edges ---
i.e., 0.078\% of all bonds.  This shows that BFS repair is nearly idle at
inference time: the decoder has learned to produce connected graphs during
training via the BFS-STE straight-through estimator, and the repair mechanism
serves as a safety net that rarely fires.

Among the 211 BFS-inserted edges, 98.1\% connect atom pairs for which the
assigned bond type is chemically allowed under the $\Omega^{(k)}$ atom-pair
mask.  This is expected because the repair selects the highest-probability edge
predicted by the adjacency head, so even forced connections involve plausible
atom pairs.  The mean bond length of inserted edges is 2.55\,\AA{}
($\sigma{=}0.41$), longer than typical covalent bonds, with 40.8\% falling
within the standard covalent bonding range of 0.8--2.5\,\AA.  This is
consistent with their role as last-resort cross-fragment connections between
components that were nearly but not quite linked by the decoder. 

\begin{table}[t]
\centering
\caption{BFS-inserted edge audit on GEOM-Drugs ($n{=}10{,}000$ generated
molecules).  A BFS-inserted edge is an edge absent after initial adjacency
thresholding that was explicitly added by BFS connectivity repair.  For
comparison, the corresponding statistics for all graph-theoretic bridge bonds
(cut edges) are shown in the right column; note that the vast majority of
bridge bonds are predicted by the decoder, not inserted by repair.}
\label{tab:bridge_audit}
\begin{tabular}{lcc}
\toprule
\textbf{Statistic} & \textbf{BFS-inserted edges} & \textbf{All bridge bonds} \\
\midrule
Molecules affected
    & 2.1\% (210/10{,}000)   & 99.9\% (9{,}989/10{,}000) \\
Total edges
    & 211                     & 96{,}791 \\
Mean per molecule
    & 0.02                    & 9.68 \\
Mean fraction of all edges
    & 0.078\%                  & 35.8\% \\
Chemically allowed atom pairs
    & 98.1\%                  & 99.97\% \\
Bond length in covalent range [0.8, 2.5]\,\AA
    & 40.8\%                  & 99.87\% \\
Mean bond length
    & 2.55\,\AA{} ($\sigma{=}0.41$) & 1.43\,\AA{} ($\sigma{=}0.15$) \\
Bond type: Single / Double / Triple
    & 98.1\% / 1.5\% / ${<}1$\%     & 81\% / 18\% / ${<}1$\% \\
\bottomrule
\end{tabular}
\end{table}

\paragraph{Degree and ring distributions.}
To assess whether TopoBridge systematically increases graph density, we compare
atom-degree, ring-size, and bridge-bond distributions against the reference
test set and UDM-3D, which shares the same diffusion backbone but has no
TopoBridge module (Table~\ref{tab:degree_ring_dist}).

\begin{table}[t]
\centering
\caption{Degree, ring, and graph-theoretic bridge-bond statistics on
GEOM-Drugs ($n{=}10{,}000$).  Lower EMD\,$\downarrow$ indicates closer match
to the reference distribution.}
\label{tab:degree_ring_dist}
\begin{tabular}{lccc}
\toprule
Metric & Reference & TopVAE+DiT & UDM-3D \\
\midrule
Mean atom degree                         & 2.089 & 2.185 & 2.146 \\
Graph-theoretic bridge-bond fraction     & 0.643 & 0.356 & 0.384 \\
Degree EMD vs.\ ref.\ $\downarrow$       & ---   & 0.568 & \textbf{0.558} \\
Ring-size EMD vs.\ ref.\ $\downarrow$    & ---   & \textbf{0.561} & 0.706 \\
\bottomrule
\end{tabular}
\end{table}

The mean degree of TopVAE-generated molecules (2.185) is within 5\% of the
reference value (2.089) and comparable to UDM-3D (2.146), which uses no
TopoBridge connectivity refinement.  The two models trade wins across the EMD
metrics: TopVAE+DiT better reproduces ring sizes ($0.561$ vs.\ $0.706$), while
the degree EMDs are nearly tied ($0.568$ vs.\ $0.558$).  The graph-theoretic
bridge-bond fraction of TopVAE+DiT (0.356) is lower than the reference value
(0.643) and close to UDM-3D (0.384), showing that TopoBridge does not inflate
the fraction of final bonds occupying graph-theoretic bridge positions.

\paragraph{Summary.}
Taken together, these results support three conclusions.
First, TopoBridge's primary contribution is at \emph{training} time: the
BFS-STE gradient signal teaches the decoder to predict connected graphs, so
that BFS repair at inference is nearly idle (2.1\% of molecules, 0.078\% of
edges).
Second, when repair does fire, it inserts a single edge connecting chemically
valid atom pairs (98.1\% allowed), though with atypically long bond lengths
(mean 2.55\,\AA) reflecting their last-resort nature.  This affects a
negligible fraction of all edges and does not measurably perturb overall
geometric quality.
Third, TopoBridge does not systematically increase graph density: generated
molecules exhibit near-reference mean degree and competitive degree EMD. Since
BFS repair fires on only 2.1\% of generated molecules, the remaining connectivity
gain is attributable to learned decoder behavior rather than chemically
implausible forced edges.

\subsection{Design Justification: Why Not Soft Connectivity Penalties?}
\label{app:connectivity_design}

A natural alternative to TopoBridge is a differentiable soft
connectivity penalty---for example, maximising the Fiedler value
(second-smallest eigenvalue of the graph Laplacian) or penalising the
number of connected components via a smooth surrogate.
We explored soft penalty formulations and encountered several
practical difficulties based on our experiments:
\begin{enumerate}
  \item \textbf{No guarantee.} Soft penalties reduce but do not
    eliminate disconnected outputs; molecules can still fragment when
    the penalty weight is insufficient or when the penalty landscape
    has local minima.
  \item \textbf{Loss balancing.} The appropriate penalty weight varies
    with molecule size, batch composition, and training stage,
    requiring extensive hyperparameter search that hard TopoBridge
    avoids entirely.
  \item \textbf{Gradient quality.} Computing eigenvalues of the
    $N{\times}N$ Laplacian at each training step is $O(N^3)$ and
    produces gradients that are numerically unstable near degenerate
    eigenvalues (exactly the regime of near-disconnected graphs).
    TopoBridge's BFS procedure is $O(N{+}E)$ and uses a
    straight-through estimator that avoids these issues.
\end{enumerate}
TopoBridge provides a hard connectivity guarantee at modest
computational cost that self-anneals during training
(Appendix~\ref{app:bfs_annealing}), and the chemical plausibility
audit (Appendix~\ref{app:bridge_audit}) confirms that this guarantee
does not come at the expense of chemical validity.  We therefore adopt
TopoBridge as the preferred connectivity mechanism.

\clearpage
\section{Computational Cost Analysis}
\label{app:computational_cost}

We profile TopVAE's training cost on an NVIDIA H100 NVL GPU with
batch size 64 on GEOM-Drugs. Component timings are forward-only and averaged
over 50 forward passes after 10 warmup iterations. Training-step timings are
averaged over 50 forward+backward passes after the same warmup.

\paragraph{Component-level breakdown.}
Table~\ref{tab:cost_breakdown} reports the per-component cost of a
single TopVAE forward pass.  The encoder accounts for 36.7\% of
forward time.  ChemCO contributes 24.0\% but is \textbf{removed
entirely at inference},
making it a training-only investment.  TopoBridge's BFS-STE adds
only 2.8\,ms per batch (2.8\% of forward time) when measured at
convergence ($\alpha{=}1.0$); as we show in
Section~\ref{app:bfs_annealing}, this cost is self-annealing and
remains negligible throughout the majority of training.
Note that Tarjan bridge detection is a standard $O(V{+}E)$ graph
algorithm independent of TopoBridge and is folded into the ``Other''
category.

\begin{table}[t]
\centering
\caption{Per-component forward-pass cost of TopVAE on GEOM-Drugs
(H100 NVL, batch size 64).  TopoBridge timing uses the BFS-only cost
from a converged model ($\alpha{=}1.0$), which is representative of
the vast majority of training.}
\label{tab:cost_breakdown}
\begin{tabular}{lrc}
\toprule
\textbf{Component} & \textbf{Time (ms)} & \textbf{\% of Forward} \\
\midrule
Encoder                              & 36.8 & 36.7\% \\
AdjHead + BFS-STE (converged)        &  2.8 &  2.8\% \\
TopoFormer                           &  2.8 &  2.8\% \\
AtomHead                             &  0.2 &  0.2\% \\
BondTypeHead                         &  1.1 &  1.1\% \\
Bond Distribution Assembly           &  0.5 &  0.5\% \\
\textbf{ChemCO (40 Steps)}           & \textbf{24.1} & \textbf{24.0\%} \\
CoordHead + EGNN                     &  7.8 &  7.8\% \\
Loss computation                     &  2.7 &  2.7\% \\
Other (Tarjan, data transfer, etc.)  & 21.8 & 21.7\% \\
\midrule
\textbf{Total forward}               & \textbf{100.3} & \textbf{100\%} \\
\bottomrule
\end{tabular}
\end{table}

\paragraph{Comparison with UAE.}
Table~\ref{tab:cost_comparison} compares the total training-step cost
between TopVAE and UAE.  TopVAE is $4.3{\times}$ slower per
forward+backward step and uses $5.1{\times}$ more peak GPU memory.
Under 4-GPU DDP (H100 NVL), communication partially overlaps with
computation and the effective per-step ratio drops to $3.11{\times}$
($295$\,ms/step vs.\ $95$\,ms/step from training logs).  This
overhead is a training-time investment: at inference, ChemCO is
disabled (saving 24\% of forward time), and the remaining modules
produce guaranteed-connected, chemically valid molecules-.

\begin{table}[t]
\centering
\caption{Training cost comparison between TopVAE and UAE on
GEOM-Drugs (H100 NVL, batch size 64).}
\label{tab:cost_comparison}
\begin{tabular}{lccc}
\toprule
\textbf{Metric} & \textbf{TopVAE} & \textbf{UAE} & \textbf{Ratio} \\
\midrule
Forward only (ms)       & 100.3 & 41.1 & $2.4{\times}$ \\
Forward + backward (ms) & 180.9 & 41.4 & $4.3{\times}$ \\
Peak GPU memory (MB)    & 4{,}546 & 895 & $5.1{\times}$ \\
\bottomrule
\end{tabular}
\end{table}

\subsection{TopoBridge Cost is Self-Annealing}
\label{app:bfs_annealing}

The TopoBridge's BFS repair cost depends on the number of disconnected components
in the initial top-$k$ graph, which in turn depends on the quality of
the adjacency predictor.  Early in training, the predictor outputs
near-random probabilities, producing highly fragmented graphs that
require many bridge insertions; as training progresses, the predictor
sharpens and the top-$k$ graph is already connected for most
molecules, so the BFS loop fires rarely.

\begin{figure}[t]
  \centering
  \includegraphics[width=\linewidth]{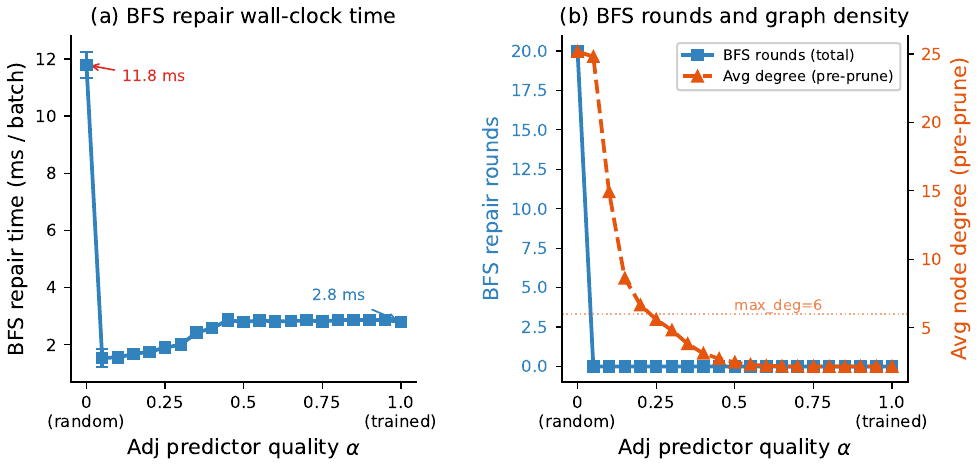}
  \caption{BFS repair cost as a function of adjacency predictor
  quality $\alpha$ (GEOM-Drugs, H100 NVL, batch 64).
  \textbf{(a)}~Wall-clock time drops $4.2{\times}$ from random
  ($11.8$\,ms) to converged ($2.8$\,ms).
  \textbf{(b)}~The number of BFS repair rounds (blue) and the
  pre-prune average node degree (orange) both decrease as the
  predictor sharpens.}
  \label{fig:bfs_vs_alpha}
\end{figure}

Figure~\ref{fig:bfs_vs_alpha} quantifies this relationship by
interpolating between a random adjacency matrix and the converged
predictor output via a mixing coefficient
$\alpha \in [0,1]$: $P^A_\alpha = \alpha \cdot P^A_{\text{trained}}
+ (1{-}\alpha) \cdot P^A_{\text{random}}$. \textit{The data is simulated which might different from real training data.}
At $\alpha{=}0$ (random), the BFS repair takes $11.8$\,ms/batch and
requires ${\sim}20$ bridge insertions per molecule; by $\alpha{=}0.25$
the cost has already collapsed to ${\sim}3$\,ms, and at $\alpha{=}1.0$
(converged) it stabilises at $2.8$\,ms (${\sim}0$ repair rounds for
most molecules).  Panel (b) shows that the average node degree
(pre-prune) drops from ${\sim}25$ to the \texttt{max\_deg}{$=$}6
ceiling as the predictor improves, explaining why the initial graph
becomes well-connected and repair becomes unnecessary.

\begin{figure}[t]
  \centering
  \includegraphics[width=0.55\linewidth]{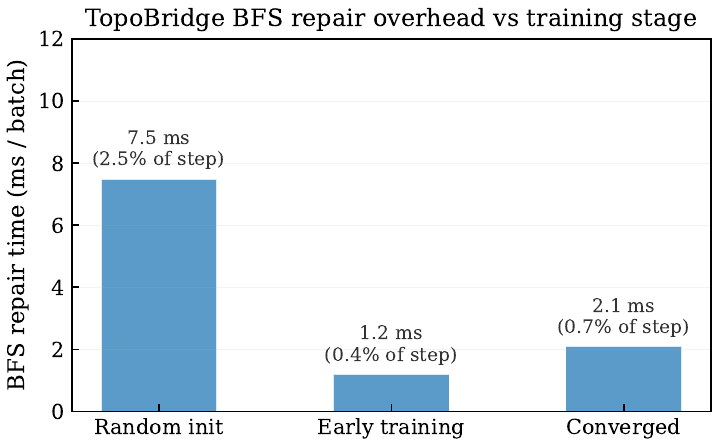}
  \caption{TopoBridge BFS repair overhead at three training stages.
  The cost drops from 2.5\% to ${<}1$\% of total step time as the
  adjacency predictor improves.}
  \label{fig:bfs_progression}
\end{figure}

Figure~\ref{fig:bfs_progression} shows the same effect measured at
three discrete training stages (random initialisation, early training,
converged).  The BFS repair overhead drops from $7.5$\,ms
(2.5\% of step) at random initialisation to $1.2$\,ms (0.4\%) during
early training and $2.1$\,ms (0.7\%) at convergence, confirming that
the cost self-anneals and remains negligible throughout the majority
of training. The data is collected during real training process.

\subsection{Scaling with Molecule Size and Batch Size}
\label{app:scaling}

\begin{figure}[t]
  \centering
  \includegraphics[width=\linewidth]{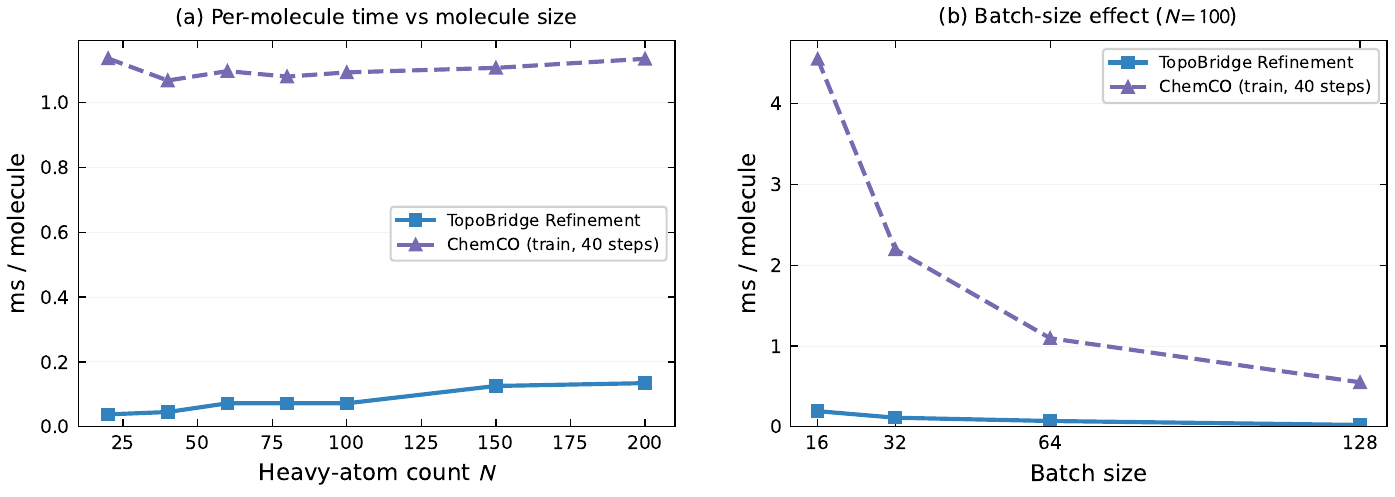}
  \caption{Per-molecule wall-clock time for TopoBridge refinement
  and ChemCO (40 steps).
  \textbf{(a)}~Scaling with heavy-atom count $N$ (batch 64).
  TopoBridge grows slowly ($0.05 \to 0.13$\,ms/mol from $N{=}25$ to
  $200$), while ChemCO remains the dominant cost at
  ${\sim}1.1$\,ms/mol across all sizes.
  \textbf{(b)}~Batch-size effect at $N{=}100$.  TopoBridge is
  batch-size-insensitive; ChemCO benefits from GPU parallelism at
  larger batches.}
  \label{fig:scalability}
\end{figure}

A natural concern is whether TopoBridge and ChemCO scale to
molecules larger than those in GEOM-Drugs (${\leq}90$ heavy atoms).
We profile both components on synthetic inputs with $N$ ranging from
25 to 200 heavy atoms (Figure~\ref{fig:scalability}).

\paragraph{TopoBridge refinement} scales nearly flat with molecule
size: per-molecule cost increases only from ${\sim}0.05$\,ms at
$N{=}25$ to ${\sim}0.13$\,ms at $N{=}200$ (Figure~\ref{fig:scalability}a).
This is expected because the BFS repair cost is governed by the number
of disconnected components (and hence bridge insertions), not directly
by atom count.  For a molecule with $c$ components the greedy loop
executes exactly $c{-}1$ insertions, each requiring an $O(N)$ BFS.
The worst case is $O(N^2)$ (all atoms isolated), but the converged
adjacency predictor produces less components compared to early stage, keeping repair cost near-constant.  TopoBridge
is also insensitive to batch size
(Figure~\ref{fig:scalability}b), remaining at
${\sim}0.1$\,ms/mol across batch sizes 16--128.

\paragraph{ChemCO} is the true scalability bottleneck.  Its
per-molecule cost (${\sim}1.1$\,ms, 40 steps) is roughly
$10{\times}$ that of TopoBridge across all molecule sizes and dominates
at small batch sizes due to lower GPU utilisation
(Figure~\ref{fig:scalability}b).  Because ChemCO is removed at
inference with no quality loss (Table~6), the scalability concern
applies only to training.  For future work on molecules with
$>$100 heavy atoms, ChemCO's iterative projection is the natural
target for amortised or approximate alternatives; TopoBridge itself is
expected to remain negligible.

Table~\ref{tab:scaling_size} summarises the end-to-end forward-pass
scaling on real GEOM-Drugs molecules, confirming modest overhead
growth.

\begin{table}[t]
\centering
\caption{Forward-pass time by molecule size on GEOM-Drugs
(single GPU, batch 64).}
\label{tab:scaling_size}
\begin{tabular}{lrrr}
\toprule
Atom count & TopVAE (ms) & UAE (ms) & Ratio \\
\midrule
$N \in [15, 25)$ & 88.7  & 41.1 & $2.2{\times}$ \\
$N \in [25, 35)$ & 101.1 & 41.4 & $2.4{\times}$ \\
\bottomrule
\end{tabular}
\end{table}

\clearpage
\section{Training Protocol}
\label{app:training}

This section provides complete training details for reproducing TopVAE.

\subsection{Data Preprocessing}
\label{app:data}

\paragraph{Kekulization.}
All molecules are kekulized via
\texttt{Chem.Kekulize(mol, clearAromaticFlags=True)}, converting aromatic bonds
to alternating single/double bonds and reducing the bond vocabulary from
$\{\text{single, double, triple, aromatic}\}$ to $K\!=\!3$ kekulized types
$\{\text{single, double, triple}\}$. Molecules in datasets that fail kekulization are
discarded.

\paragraph{Atom and bond vocabularies.}
QM9 (heavy-only): 4 atom types $\{\text{C, N, O, F}\}$, max 9 heavy atoms.
GEOM-Drugs (heavy-only): 12 atom types
$\{\text{C, N, O, F, S, Cl, Br, P, I, B, Si, Bi}\}$, max 90 heavy atoms.

\paragraph{Input features.}
The encoder receives a fully-connected edge graph ($N\!\times\!N$).
Bond edges carry one-hot bond type; non-bond edges carry zeros; self-loops
receive a dedicated flag. Node features concatenate atom-type one-hot with
39 GeoMol-style features (atomic number, aromaticity, degree, hybridization,
implicit valence, formal charge, ring membership), plus 3D coordinates.

\paragraph{Augmentation.}
All coordinates are mean-centered. GEOM-Drugs training applies random SO(3)
rotation and Gaussian translation (scale $0.1$); QM9 does not.

\subsection{Model Architectures}
\label{app:arch}

\paragraph{Encoder.}
Both UAE and TopVAE share the same encoder based on DMTBlock
transformer layers, with Gaussian Basis Function distance
embedding and message-passing attention. The encoder projects per-atom hidden
states to $\bmu$ and $\log\bsigma^2$ via linear heads, with posterior
log-variance clamped to $[-10,4]$ on GEOM-Drugs.

\paragraph{TopoFormer blocks.}
The topology-conditioned transformer (Sec.~\ref{sec:decoder}) uses
Graphormer-style additive pair bias from the binary adjacency:
$\text{bias}\!=\!\texttt{nn.Embedding}(2,n_\mathrm{heads})[\widetilde A]$.
Each block applies pre-LayerNorm, multi-head self-attention with additive
bias, and a GELU FFN ($4\times$ expansion), with residual connections.

\paragraph{EGNN coordinate head.}
The E($n$)-equivariant coordinate prediction uses normalized directions with
tanh-bounded magnitude:
\begin{equation}
    \Delta\br_i
    =
    \sum_j\frac{\br_i-\br_j}{\|\br_i-\br_j\|+\epsilon}\;
    \tanh\!\bigl(\phi_x(m_{ij})\bigr),
    \qquad
    \br_i\leftarrow\br_i+\Delta\br_i,
\end{equation}
where $\epsilon\!=\!10^{-6}$ and $\phi_x$ is a learned scalar MLP. Coordinates
are re-centered after each EGNN layer.

\paragraph{Bond type head.}
For each pair $(i,j)$ on the support $\widetilde A$, the bond head
concatenates node pair features, learned atom-type embeddings, adjacency
embedding, and optionally RBF distance features (16 Gaussian centers spanning
$[0.8,3.0]$\,\AA{} on GEOM-Drugs).

\paragraph{Adjacency head.}
Edge-existence logits are produced by a 2-layer MLP on concatenated node pairs,
symmetrized and zero-diagonal before sigmoid.

\paragraph{Architecture summary.}
\begin{table}[t]
\centering\small
\caption{Architecture configurations.}
\label{tab:arch_summary}
\begin{tabular}{@{}lcc@{}}
\toprule
\textbf{Parameter} & \textbf{QM9} & \textbf{GEOM-Drugs} \\
\midrule
Latent dim $d$ & 16 & 32 \\
Encoder hidden / heads / blocks & 64 / 4 / 3 & 128 / 8 / 4 \\
Decoder (Topo / Geo layers) & 3 / 3 & 4 / 4 \\
Decoder hidden / heads & 64 / 4 & 128 / 8 \\
Edge dim & 64 & 128 \\
Dropout & 0.1 & 0.1 \\
RBF distance features & No & Yes (16 centers) \\
\bottomrule
\end{tabular}
\end{table}

\subsection{Training Hyperparameters}
\label{app:hparams}

\begin{table}[t]
\centering\small
\caption{Training hyperparameters for TopVAE (Stage~1: VAE).}
\label{tab:train_hparams}
\begin{tabular}{@{}lcc@{}}
\toprule
\textbf{Parameter} & \textbf{QM9} & \textbf{GEOM-Drugs} \\
\midrule
Optimizer & AdamW & AdamW \\
Learning rate & $10^{-3}$ & $3\times10^{-4}$ \\
Weight decay & $10^{-5}$ & $10^{-5}$ \\
LR scheduler & Constant & Cosine ($T_\mathrm{max}\!=\!1200$) \\
LR warmup & -- & Linear, 50 epochs \\
Max epochs & 2000 & 1200 \\
Batch size $\times$ GPUs & $256\times4$ & $128\times4$ \\
Precision & fp32 & fp32 \\
Gradient clipping & -- & 1.0 \\
\midrule
\multicolumn{3}{@{}l}{\emph{Loss weights (Eq.~\ref{eq:total-loss})}} \\
$w_\mathrm{adj}$ & 1.0 & 10.0 \\
$w_\mathrm{atom}$, $w_\mathrm{bond}$, $\lambda_R$ & 1, 1, 1 & 1, 1, 1 \\
$\lambda_D$ (all-pairs + bonded-pair distance) & 0 & 1.0 \\
$\lambda_\mathrm{AGCL}$ & 1.0 & 0.1 \\
$\beta$ (KL weight) & $10^{-6}$ & $5\times10^{-6}$ \\
\midrule
\multicolumn{3}{@{}l}{\emph{TopoBridge}} \\
Threshold $\tau$ & 0.5 & 0.3 \\
Max degree cap & 4 & 6 (Considering charges) \\
Adjacency loss & BCE & Focal ($\alpha\!=\!0.75$, $\gamma\!=\!1.0$) \\
\midrule
\multicolumn{3}{@{}l}{\emph{ChemCO (Sec.~\ref{sec:chemco},
                              Appendix~\ref{app:chempo})}} \\
$T_\mathrm{CO}$ (train / eval) & 20 / 40 & 40 / 100 \\
Primal base rate $\alpha$ / decay $\gamma_\alpha$ & 0.05 / 0.99 & 0.05 / 0.99 \\
Dual base rates $\eta_\mu$, $\eta_+$, $\eta_-$ & 0.1, 0.2, 0.2 & 0.1, 0.2, 0.2 \\
Active-set sharpness $\gamma$ & 10.0 & 10.0 \\
Chemical mask $\Omega$ & No & Yes \\
\bottomrule
\end{tabular}
\end{table}

\subsection{Loss Function Details}
\label{app:losses}

The full training objective (Eq.~\ref{eq:total-loss}) consists of the terms
defined below. Let $\bm{M}_\mathrm{node}\!\in\!\{0,1\}^{B\times N}$ denote the
valid-atom mask and
$\bm{M}_\mathrm{pair}=\bm{M}_\mathrm{node}^{(i)}\wedge
\bm{M}_\mathrm{node}^{(j)}\wedge\mathbb{1}[i\!<\!j]$ the upper-triangular
valid-pair mask.

\paragraph{(i) Adjacency loss $\mathcal{L}_\mathrm{adj}$.}
On GEOM-Drugs we use focal binary cross-entropy:
\begin{equation}
    \mathcal{L}_\mathrm{adj}
    =
    \frac{1}{|\bm{M}_\mathrm{pair}|}
    \sum_{(i,j)\in\bm{M}_\mathrm{pair}}
    \alpha_t\,(1-p_t)^\gamma\,\mathrm{BCE}(s_{ij},t_{ij}),
\end{equation}
with $\alpha\!=\!0.75$ (up-weight bond edges) and $\gamma\!=\!1.0$. QM9 uses
standard BCE ($\gamma\!=\!0$). We found $\gamma\!=\!2.0$ caused negative-logit
drift; reducing to $\gamma\!=\!1.0$ eliminated this without additional
regularization.

\paragraph{(ii) Atom type loss $\mathcal{L}_\mathrm{atom}$.}
Cross-entropy over predicted atom types, averaged over valid atoms:
\begin{equation}
    \mathcal{L}_\mathrm{atom}
    =
    \frac{1}{|\bm{M}_\mathrm{node}|}
    \sum_{i\in\bm{M}_\mathrm{node}}
    \mathrm{CE}\!\bigl(A_{\mathrm{logits},i},\;a_i^\mathrm{gt}\bigr).
\end{equation}

\paragraph{(iii) Bond type loss $\mathcal{L}_\mathrm{bond}$.}
Conditional cross-entropy computed only on ground-truth bond edges within the
TopoBridge support $\widetilde A$:
\begin{equation}
    \mathcal{L}_\mathrm{bond}
    =
    \frac{1}{|\bm{M}_\mathrm{bond}|}
    \sum_{(i,j)\in\bm{M}_\mathrm{bond}}
    \mathrm{CE}\!\bigl(q_{ij},\;t_{ij}^\mathrm{bond}\!-\!1\bigr),
\end{equation}
where $\bm{M}_\mathrm{bond}=\bm{M}_\mathrm{pair}\wedge(t_{ij}^\mathrm{bond}\!>\!0)
\wedge\widetilde A$, and $q_{ij}\!\in\!\mathbb{R}^K$ are the bond-type logits.
The target is shifted by $-1$ because the bond head predicts over
$\{1,\ldots,K\}$; bond absence is handled by the adjacency gate.

\paragraph{(iv) Coordinate loss $\mathcal{L}_\mathrm{coord}$.}
MSE between predicted and ground-truth coordinates, both zero-centered per
molecule:
\begin{equation}
    \mathcal{L}_\mathrm{coord}
    =
    \frac{1}{3|\bm{M}_\mathrm{node}|}
    \sum_{i\in\bm{M}_\mathrm{node}}
    \|\hat{\br}_i-\br_i^\mathrm{gt}\|_2^2.
\end{equation}

\paragraph{(v) Distance loss $\mathcal{L}_\mathrm{dist}$.}
This term provides rotation-invariant geometric supervision and is the sum
of an all-pairs distance MSE and a bonded-pair distance MSE:
\begin{equation}
    \mathcal{L}_\mathrm{dist}
    =
    \frac{1}{|\bm{M}_\mathrm{full}|}
    \sum_{(i,j)\in\bm{M}_\mathrm{full}}
    (\hat D_{ij}-D_{ij}^\mathrm{gt})^2
    \;+\;
    \frac{1}{|\bm{M}_\mathrm{bonded}|}
    \sum_{(i,j)\in\bm{M}_\mathrm{bonded}}
    (\hat D_{ij}-D_{ij}^\mathrm{gt})^2,
\end{equation}
where $D_{ij}\!=\!\|\br_i-\br_j\|_2$. The bonded-pair term focuses geometric
learning on bond lengths, where distance errors have the largest chemical
impact. Active on GEOM-Drugs ($\lambda_D\!=\!1$); inactive on QM9
($\lambda_D\!=\!0$).

\paragraph{(vi) KL divergence $\mathcal{L}_\mathrm{KL}$.}
The encoder produces per-atom posteriors
$q(z_i\mid M)=\mathcal{N}(\mu_i,\mathrm{diag}(\exp(\ell_i)))$ with
$\ell_i=\mathrm{clamp}(\log\sigma_i^2,-10,4)$:
\begin{equation}
    \mathcal{L}_\mathrm{KL}
    =
    \frac{1}{BN}\sum_{b,i}
    \Bigl[-\tfrac{1}{2}\sum_{d=1}^{D}
    \bigl(1+\ell_{i,d}-\mu_{i,d}^2-e^{\ell_{i,d}}\bigr)\Bigr].
\end{equation}
The near-zero weight ($\beta\!=\!5\!\times\!10^{-6}$ on GEOM,
$10^{-6}$ on QM9) yields a near-deterministic autoencoder whose latent space
remains structured enough for downstream diffusion.

\paragraph{(vii) AGCL loss $\mathcal{L}_\mathrm{AGCL}$.}
This is the advantage-gated selective teacher loss defined in
Sec.~\ref{sec:agcl}, Eqs.~\eqref{eq:agcl-advantage}--\eqref{eq:agcl-loss}.
Let $P_\mathrm{raw}$ and $P_\mathrm{chem}$ denote the raw decoder and
ChemCO-projected bond distributions, respectively. For each molecule~$b$
in the batch, the advantage
$a_b=[\mathcal{E}_b(P_\mathrm{raw})-\mathcal{E}_b(P_\mathrm{chem})]_+$
gates an $L_2$ consistency loss that steers $P_\mathrm{raw}$ toward
$\operatorname{sg}(P_\mathrm{chem})$. The stop-gradient detaches the ChemCO
output so that only the raw decoder is updated. When $\sum_b a_b<10^{-8}$
(i.e., ChemCO provides no advantage), the loss returns zero.
Weight: $\lambda_\mathrm{AGCL}\!=\!0.1$ (GEOM), $1.0$ (QM9).

\paragraph{Raw bond distribution.}
The decoder constructs $P_\mathrm{raw}$ by gating the conditional bond-type
softmax with the STE adjacency:
\begin{equation}
    P_{\mathrm{raw},ij}
    =
    \bigl[\underbrace{1-\widetilde A_{ij}^{\mathrm{ste}}}_{P(\text{no bond})},\;
    \underbrace{\widetilde A_{ij}^{\mathrm{ste}}\cdot
    \mathrm{softmax}(q_{ij})}_{P(\text{type }1\ldots K)}\bigr],
    \label{eq:raw-bond-dist}
\end{equation}
where $\widetilde A^{\mathrm{ste}}$ is the straight-through adjacency from
TopoBridge. Pairs outside the support are forced to
$[1,0,\ldots,0]$ (no bond).

\paragraph{Loss weight summary.}
\begin{table}[t]
\centering\small
\caption{Active loss weights per dataset. Zero entries are inactive.}
\label{tab:loss_weights}
\begin{tabular}{@{}lcc@{}}
\toprule
\textbf{Loss} & \textbf{QM9} & \textbf{GEOM-Drugs} \\
\midrule
$\mathcal{L}_\mathrm{adj}$ (adjacency) & 1.0 & 10.0 \\
$\mathcal{L}_\mathrm{atom}$ (atom type) & 1.0 & 1.0 \\
$\mathcal{L}_\mathrm{bond}$ (bond type) & 1.0 & 1.0 \\
$\lambda_R\,\mathcal{L}_\mathrm{coord}$ (coordinates) & 1.0 & 1.0 \\
$\lambda_D\,\mathcal{L}_\mathrm{dist}$ (distances) & 0 & 1.0 \\
$\beta\,\mathcal{L}_\mathrm{KL}$ (KL) & $1\!\times\!10^{-6}$ & $5\!\times\!10^{-6}$ \\
$\lambda_\mathrm{AGCL}\,\mathcal{L}_\mathrm{AGCL}$ (AGCL) & 1.0 & 0.1 \\
\bottomrule
\end{tabular}
\end{table}

\subsection{GT-Support Warmup Schedule}
\label{app:gt_warmup}

\paragraph{Motivation.}
Early in training, the adjacency predictor produces near-random graphs. If the
bond-type and coordinate heads must condition on these noisy adjacencies, they
receive inconsistent inputs that slow convergence. GT-support warmup decouples
this by providing ground-truth topology initially, then transitioning to
predicted TopoBridge supports.

\paragraph{Three-phase schedule.}
Let $W_\mathrm{gt}$ and $W_\mathrm{tr}$ be the warmup and transition durations
(in epochs). The probability of using ground-truth support is:
\begin{equation}
    p_\mathrm{gt}(\text{epoch})
    =
    \begin{cases}
        1.0 & \text{epoch}<W_\mathrm{gt}
              \;\;\text{(Phase A: pure GT)}, \\
        1-\frac{\text{epoch}-W_\mathrm{gt}}{W_\mathrm{tr}}
              & W_\mathrm{gt}\le\text{epoch}<W_\mathrm{gt}+W_\mathrm{tr}
              \;\;\text{(Phase B: linear transition)}, \\
        0.0 & \text{epoch}\ge W_\mathrm{gt}+W_\mathrm{tr}
              \;\;\text{(Phase C: fully predicted)}.
    \end{cases}
\end{equation}
During Phase~B, gating is a per-batch Bernoulli coin flip: with probability
$p_\mathrm{gt}$, the entire batch receives ground-truth adjacency and bond
types; otherwise the batch uses predicted TopoBridge supports. Settings:
$W_\mathrm{gt}\!=\!100$, $W_\mathrm{tr}\!=\!100$ on GEOM-Drugs; QM9 uses
$W_\mathrm{gt}\!=\!0$.

\subsection{Latent Diffusion Model (Stage~2)}
\label{app:ldm}

The diffusion prior is trained on frozen TopVAE latents. We describe the
backbone, noise schedule, and sampling procedure.

\paragraph{DiT backbone.}
A standard TransformerEncoder with sinusoidal timestep embeddings (base 10000).
Timestep embedding is projected through a 2-layer MLP and added to input token
embeddings.

\paragraph{Architecture configurations.}
\begin{table}[t]
\centering\small
\caption{DiT configurations for Stage~2 latent diffusion.}
\label{tab:dit_config}
\begin{tabular}{@{}lcc@{}}
\toprule
\textbf{Parameter} & \textbf{QM9} & \textbf{GEOM DiT-B} \\
\midrule
Hidden dim & 512 & 768 \\
Heads & 8 & 12 \\
Layers & 8 & 12 \\
MLP ratio & $4\times$ & $4\times$ \\
\midrule
Optimizer & \multicolumn{2}{c}{AdamW, lr $=10^{-4}$, weight decay $=0.05$} \\
LR schedule & \multicolumn{2}{c}{Linear warmup (1k steps) $+$ cosine decay} \\
Max epochs & 5000 & 10000 \\
Batch size $\times$ GPUs & $1024\!\times\!4$ & $768\!\times\!4$ \\
Precision & \multicolumn{2}{c}{fp16-mixed} \\
\bottomrule
\end{tabular}
\end{table}

\paragraph{Noise schedule.}
VP-SDE with the cosine schedule:
$\bar\alpha(t)=\cos\!\bigl(\frac{t+s}{1+s}\cdot\frac{\pi}{2}\bigr)^2$ with
offset $s\!=\!0.008$.

\paragraph{Latent whitening.}
Before diffusion training, latent codes from the frozen VAE encoder are
whitened: $z_\mathrm{norm}=(z-\mu_z)/\sigma_z$, where $\mu_z,\sigma_z$ are
computed over all training latents. The inverse transform is applied before
decoding at generation time.

\paragraph{Sampling.}
We use ancestral DDPM reverse sampling with $T\!=\!100$ steps on
$t_\mathrm{array}=\mathrm{linspace}(1\!-\!\epsilon,\epsilon,T)$,
$\epsilon\!=\!10^{-3}$. Noise temperature $\tau\!=\!1.0$.
No exponential moving average is used.

\paragraph{VAE freezing.}
The VAE is fully frozen during Stage~2: all parameters are detached, and
online encoding runs under \texttt{torch.no\_grad()}.

\subsection{Metrics}
\label{app:metrics}

\paragraph{iFID Computation Details.}
Our iFID metric adapts the interpolated FID from pixel space to the latent space of molecular VAEs~\citep{xu2026making}.
The pipeline proceeds in four stages: encoding, interpolation, decoding, and scoring.

\textit{Encoding.}
Every molecule $x$ in the training set $\mathcal{D}_{\text{train}}$ or validation set $\mathcal{D}_{\text{val}}$ is mapped to its deterministic posterior mean $\mu_\phi(x)$ by a frozen encoder with all augmentations disabled (no rotation, translation, or noise).
Using the mean rather than a reparameterised sample ensures the metric measures the geometry of the learned manifold, not sampling noise.

\textit{Interpolation.}
Each dense latent $z\in\mathbb{R}^{N_{\max}\times d}$ is mean-pooled along the node dimension to $\bar{z}$.
For each validation molecule $x_v$, we retrieve the top-$K{=}10$ neighbours by cosine similarity from training molecules with the \emph{same} atom count $|x|{=}|x_v|$, avoiding size-mismatch artefacts.
One neighbour $z_t$ is drawn categorically with weights
\[
  \pi_k = \mathrm{softmax}\!\bigl(-\|z_v - z_t^{(k)}\|_2^2\bigr).
\]
The midpoint latent is then constructed via spherical linear interpolation:
\[
  z_\alpha = \mathrm{SLERP}(z_v,\, z_t,\, \alpha), \qquad \alpha = 0.5,
\]
computed on the node-level flattened vectors, falling back to LERP when $\sin\theta < 10^{-6}$.

\textit{Decoding.}
Both $z_\alpha$ and the original $z_v$ are decoded by a frozen decoder following each model's 
standard protocol, assembled into RDKit molecules, and converted to canonical 
SMILES (\texttt{RemoveAllHs}, \texttt{MolToSmiles(isomericSmiles=False)}).

\textit{Scoring and invalid-molecule handling.}
FID features are 512-d ChemNet activations (\texttt{fcd\_torch}), from which we estimate means and covariances for the reference, interpolated, and reconstructed sets and compute the Fr\'{e}chet distance:
\[
  \mathrm{FID} = \|\mu_1 - \mu_2\|_2^2 
  + \mathrm{Tr}\!\bigl(\Sigma_1 + \Sigma_2 - 2(\Sigma_1\Sigma_2)^{1/2}\bigr).
\]
We deliberately do \emph{not} restrict evaluation to valid-and-connected subsets.
Decoding failures yield empty strings; a subsequent \texttt{MolFromSmiles(sanitize=True)} filter further removes unsanitisable or overly long ($>$350 character) SMILES.
Only surviving molecules enter ChemNet, so invalid samples do not pollute the FID value.

\clearpage
\section{Details of ChemCO}
\label{app:chempo}

This appendix provides the complete optimization details for ChemCO.
We use the same notation as Sec.~\ref{sec:chemco}: $\widetilde A$ is the
TopoBridge adjacency support, $\Omega^{(k)}$ is the chemical mask for bond
type~$k$, $\bar U$ is the relative neural utility
(Eq.~\ref{eq:chemco-utility-main}), $\Phi$ denotes the free primal logits, and
$Y^{(k)}\!=\!T_k(\Phi)$ is the induced continuous bond-type assignment
(Eq.~\ref{eq:chemco-transform-main}).

\subsection{Initialization}
\label{app:chempo-init}

ChemCO optimizes only real bond types $k=1,\ldots,K$; the no-bond class
$k\!=\!0$ is represented by the residual mass. The primal logits are initialized from
the relative neural utility:
\begin{equation}
    \Phi_{0,ij}^{(k)}
    =
    \bar U_{ij}^{(k)},
    \qquad k=1,\ldots,K.
    \label{eq:chempo-init-app}
\end{equation}
This initialization seeds the optimizer at the decoder's own belief, so that
the subsequent unroll only needs to \emph{correct} constraint violations rather
than discover the bond structure from scratch.

\paragraph{Continuous statistics.}
At iteration $t$, ChemCO computes the soft bond-type assignment
\begin{equation}
    Y_t^{(k)}
    =
    T_k(\Phi_t),
    \qquad k=1,\ldots,K,
    \label{eq:chempo-yt-app}
\end{equation}
using the transform defined in Eq.~\eqref{eq:chemco-transform-main}.
Recall that $T_k$ includes masking by the TopoBridge support
$\widetilde A$ and the chemical mask $\Omega^{(k)}$, so that
$Y_t^{(k)}$ is nonzero only on candidate edges with chemically
allowed bond types.
The three
continuous statistics introduced in Eq.~\eqref{eq:chemco-stats} are then
evaluated at iteration~$t$:
\begin{equation}
    s_{ij,t}
    =
    \sum_{k=1}^{K}Y_{ij,t}^{(k)},
    \qquad
    \operatorname{val}_{i,t}
    =
    \sum_{j\in\mathcal V,\,j\ne i}
    \sum_{k=1}^{K}
    o_kY_{ij,t}^{(k)},
    \qquad
    \deg_{i,t}
    =
    \sum_{j\in\mathcal V,\,j\ne i}s_{ij,t},
    \label{eq:chempo-stats-app}
\end{equation}
where the subscript~$t$ makes the iteration explicit (the main text omits~$t$
for conciseness). The constraint residuals (denoted~$h$ to avoid confusion
with $g(x)=\sigma(x)^2$ in Eq.~\ref{eq:chemco-transform-main}) are
\begin{equation}
    h_{ij,t}^{\mathrm{pair}}
    =
    s_{ij,t}-1,
    \qquad
    h_{i,t}^{\mathrm{val}}
    =
    \operatorname{val}_{i,t}-c_i,
    \qquad
    h_{i,t}^{\mathrm{deg}}
    =
    d_{\min}-\deg_{i,t}.
    \label{eq:chempo-residuals-app}
\end{equation}
By convention, a \emph{positive} residual always indicates a violated
constraint.

\paragraph{Dual warm-start.}
Rather than initializing all dual variables at zero, ChemCO warm-starts them
from the initial constraint violations:
\begin{equation}
    \mu_{ij,0}
    =
    \bigl[s_{ij,0}-1\bigr]_+,
    \qquad
    \lambda_{i,0}^+
    =
    \bigl[\operatorname{val}_{i,0}-c_i\bigr]_+,
    \qquad
    \lambda_{i,0}^-
    =
    \bigl[d_{\min}-\deg_{i,0}\bigr]_+.
    \label{eq:chempo-dual-init-app}
\end{equation}
This gives the solver a nonzero correction signal from the first iteration,
improving convergence speed compared to a cold start. Nonnegativity is
guaranteed because $[\cdot]_+\!\ge\!0$.

\subsection{\texorpdfstring{Chemical Mask $\Omega^{(k)}$ Construction}{Chemical Mask Omega(k) Construction}}
\label{app:omega_mask}

The chemical mask $\Omega^{(k)}\!\in\!\{0,1\}^{N\times N}$ encodes which
bond types are chemically allowed for each atom pair. It is \emph{static}
and derived from standard valence-bond chemistry, not learned or fitted to
any dataset.
During training, $\Omega^{(k)}$ is constructed from the predicted atom-type
$\arg\max$. Since $\Omega^{(k)}$ enters only as a binary mask in the forward
pass and is not differentiated through, this discrete operation does not
affect gradient computation. At inference time, ChemCO is removed entirely
(Table~\ref{tab:ablation}), so the mask is not needed.
The mask is constructed by
\texttt{build\_atom\_pair\_type\_mask()} using the following rules:

\begin{table}[t]
\centering
\caption{Allowed bond types per atom-pair combination in $\Omega^{(k)}$.
S = single, D = double, T = triple.}
\label{tab:omega_mask}
\small
\begin{tabular}{@{}ll@{}}
\toprule
\textbf{Atom Pair} & \textbf{Allowed Bond Types} \\
\midrule
C--C & S, D, T \\
C--N & S, D, T \\
C--O & S, D \\
C--S & S, D \\
C--F / C--Cl / C--Br / C--I & S only \\
N--N & S, D, T \\
N--O & S, D \\
O--O & S only \\
F/Cl/Br/I -- any & S only \\
S--S & S, D \\
S--N & S, D \\
S--O & S, D \\
P--C / P--N / P--O / P--S & S, D \\
B--C / B--N / B--O & S, D \\
Si--C / Si--N / Si--O & S, D \\
Bi -- any & S only \\
\bottomrule
\end{tabular}
\end{table}

For atom-pair combinations not listed above, the default fallback allows
single bonds only ($\Omega^{(\mathrm{single})}_{ij}\!=\!1$,
$\Omega^{(k)}_{ij}\!=\!0$ for $k\!\ge\!2$).

On QM9 (4 atom types: C, N, O, F), the default configuration uses only
implicit valence caps without explicit pair restrictions ($\Omega^{(k)}\!=\!\bm{1}$);
on GEOM-Drugs (12 atom types), the full mask is active.
The sensitivity of $\Omega^{(k)}$ is studied in Appendix~\ref{app:sensitivity}
(Table~\ref{tab:omega_sensitivity}): adding the explicit mask to QM9 yields a
19\% relative gain in 3D Stab$\wedge$Conn, confirming that pair-level
restrictions capture chemical knowledge beyond what valence caps alone provide.

\subsection{Adaptive penalty objective and smoothed active-set direction}
\label{app:chemco-penalty}

We introduce nonnegative multipliers $\mu_{ij}\ge 0$ for pair exclusivity,
$\lambda_i^+\ge 0$ for valence upper bounds, and $\lambda_i^-\ge 0$ for
minimum-degree constraints. These correspond to the main-text multipliers
$\mu$, $\lambda$, and $\nu$, respectively; the split notation
$\lambda^+$/$\lambda^-$ makes the sign convention explicit.

At iteration $t$, ChemCO uses a fixed-dual adaptive penalty score
\begin{equation}
\begin{aligned}
\mathcal J_t(\Phi)
&=
\sum_{\substack{i<j\\ i,j\in\mathcal V}}
\sum_{k=1}^{K}
\bar U_{ij}^{(k)}Y_{ij}^{(k)}
-\rho_{\mathrm{pair}}
\sum_{\substack{i<j\\ i,j\in\mathcal V}}
\mu_{ij,t}\psi_\gamma(h_{ij,t}^{\mathrm{pair}})
\\
&\quad
-\sum_{i\in\mathcal V}
\lambda_{i,t}^{+}\psi_\gamma(h_{i,t}^{\mathrm{val}})
-\sum_{i\in\mathcal V}
\lambda_{i,t}^{-}\psi_\gamma(h_{i,t}^{\mathrm{deg}}),
\end{aligned}
\label{eq:app-chemco-penalty}
\end{equation}
where
\begin{equation}
h_{ij,t}^{\mathrm{pair}}=s_{ij,t}-1,\qquad
h_{i,t}^{\mathrm{val}}=\operatorname{val}_{i,t}-c_i,\qquad
h_{i,t}^{\mathrm{deg}}=d_{\min}-\deg_{i,t}.
\label{eq:app-chemco-residuals}
\end{equation}
Here $\psi_\gamma(r)=\gamma^{-1}\log(1+\exp(\gamma r))$ is a smooth
approximation to $[r]_+$, and
\begin{equation}
\psi_\gamma'(r)=\sigma(\gamma r).
\end{equation}
Thus the smoothed active-set gates are
\begin{equation}
\omega_{ij,t}^{\mathrm{pair}}
=
\sigma(\gamma h_{ij,t}^{\mathrm{pair}}),\qquad
\omega_{i,t}^{\mathrm{val}}
=
\sigma(\gamma h_{i,t}^{\mathrm{val}}),\qquad
\omega_{i,t}^{\mathrm{deg}}
=
\sigma(\gamma h_{i,t}^{\mathrm{deg}}).
\label{eq:app-chemco-gates}
\end{equation}
When $\gamma$ is large, these gates approach binary indicators of violated
constraints; for finite $\gamma$, they provide a smooth active-set
approximation.

The induced smoothed direction with respect to the continuous assignment mass
$Y_{ij}^{(k)}$ is
\begin{equation}
\begin{aligned}
\widetilde G_{ij,t}^{(k)}
&=
\bar U_{ij}^{(k)}
-\rho_{\mathrm{pair}}
\omega_{ij,t}^{\mathrm{pair}}\mu_{ij,t}
\\
&\quad
-o_k\!\left(
\omega_{i,t}^{\mathrm{val}}\lambda_{i,t}^{+}
+
\omega_{j,t}^{\mathrm{val}}\lambda_{j,t}^{+}
\right)
+
\omega_{i,t}^{\mathrm{deg}}\lambda_{i,t}^{-}
+
\omega_{j,t}^{\mathrm{deg}}\lambda_{j,t}^{-}.
\end{aligned}
\label{eq:app-chemco-soft-direction}
\end{equation}
The implementation uses $\rho_{\mathrm{pair}}=2$, matching the row-wise
pair penalty followed by explicit symmetrization. Equivalently, this constant
can be absorbed into the pair multiplier scale.

\subsection{Row-centric direction and direct logit-space update}
\label{app:chemco-logit-update}

For efficient batched implementation, ChemCO constructs the direction in two
stages. First, it forms a row-centric direction that contains atom $i$'s
node-level multiplier contributions:
\begin{equation}
\widehat G_{ij,t}^{(k)}
=
\tfrac{1}{2}\bar U_{ij}^{(k)}
-
\tfrac{\rho_{\mathrm{pair}}}{2}
\omega_{ij,t}^{\mathrm{pair}}\mu_{ij,t}
-
o_k\omega_{i,t}^{\mathrm{val}}\lambda_{i,t}^{+}
+
\omega_{i,t}^{\mathrm{deg}}\lambda_{i,t}^{-}.
\label{eq:app-chemco-row-direction}
\end{equation}
The symmetrized direction is then
\begin{equation}
\widetilde G_{ij,t}^{(k)}
=
\widehat G_{ij,t}^{(k)}
+
\widehat G_{ji,t}^{(k)}.
\label{eq:app-chemco-sym-direction}
\end{equation}
Since $\bar U$ and $\mu$ are symmetric, the utility terms sum to
$\bar U_{ij}^{(k)}$, and the pair-exclusivity terms sum to
$-\rho_{\mathrm{pair}}\omega_{ij,t}^{\mathrm{pair}}\mu_{ij,t}$. The
node-level terms collect the valence and degree contributions from both
endpoints, yielding Eq.~\eqref{eq:app-chemco-soft-direction}.

\paragraph{Direct logit-space heuristic update.}
ChemCO applies this direction directly to the free logits:
\begin{equation}
\Phi_{t+1}^{(k)}
=
\Phi_t^{(k)}
+
\eta_\Phi^{(t)}
\widetilde G_t^{(k)}\odot\Omega^{(k)},
\qquad
k=1,\ldots,K,
\label{eq:app-chemco-primal-update}
\end{equation}
where $\eta_\Phi^{(t)}=\alpha\gamma_\alpha^t$. The TopoBridge support
$\widetilde A$ is enforced by the transform $T_k(\Phi)$, and the chemical
mask $\Omega^{(k)}$ zeros out forbidden atom--bond combinations.

This update is not the exact chain-rule gradient of
Eq.~\eqref{eq:app-chemco-penalty} with respect to $\Phi$. A true chain-rule
gradient would include the derivative of $g(\Phi)=\sigma(\Phi)^2$, which
contains a factor proportional to $\sigma(\Phi)^2(1-\sigma(\Phi))$ and can
vanish when a currently absent bond has $\Phi\ll 0$. ChemCO instead uses the
smoothed $Y$-space direction directly in logit space, allowing absent bonds to
be activated by the unrolled correction.
\paragraph{Direct logit-space heuristic update.}
The primal update adds $\widetilde G_t$ directly to the logits:
\begin{equation}
    \Phi_{t+1}^{(k)}
    =
    \Phi_t^{(k)}
    +
    \eta_\Phi^{(t)}\;\widetilde G_t^{(k)}
    \odot \Omega^{(k)},
    \qquad k=1,\ldots,K,
    \label{eq:chempo-primal-update-app}
\end{equation}
where the masking by $\Omega^{(k)}$ zeros out updates on chemically forbidden
pairs, and $\eta_\Phi^{(t)}=\alpha\,\gamma_\alpha^t$ is an exponentially
decayed step size with base rate $\alpha$ and decay factor
$\gamma_\alpha\!\in\!(0,1]$.

\paragraph{Why a direct logit-space update instead of a chain-rule gradient?}
A standard approach would chain-rule $\widetilde G$ through the
transform~$T_k$ (Eq.~\ref{eq:chemco-transform-main}), yielding a logit-space
gradient proportional to $\sigma(\Phi)^2(1\!-\!\sigma(\Phi))\odot\widetilde G$.
However, the $\sigma^2(1\!-\!\sigma)$ factor suppresses the gradient when
$\sigma(\Phi)\!\to\!0$, making it difficult for currently-absent bonds
($\Phi\!\ll\!0$) to become active. The direct logit-space heuristic bypasses
this suppression, providing uniform update magnitude across all logit values and
allowing bonds to transition smoothly from absent to present. In practice this
improves convergence and is essential for effective end-to-end differentiable
training through the unrolled solver. We note that this update is not the true
gradient of the penalty objective with respect to $\Phi$; it is a heuristic
logit-space update that uses the $Y$-space direction directly in logit space.

After the logit update, the assignment is recomputed via
$Y_{t+1}^{(k)}=T_k(\Phi_{t+1})$, and the continuous statistics
$s$, $\operatorname{val}$, $\deg$ are refreshed.

\subsection{Dual updates}
\label{app:chemco-dual-updates}

After the logit update, ChemCO recomputes
$Y_{t+1}^{(k)}=T_k(\Phi_{t+1})$ and refreshes
$s_{t+1}$, $\operatorname{val}_{t+1}$, and $\deg_{t+1}$. The multipliers are
then updated by accumulating positive residuals:
\begin{equation}
\mu_{ij,t+1}
=
\mu_{ij,t}
+
\eta_\mu^{(t)}
\bigl[s_{ij,t+1}-1\bigr]_+,
\qquad
\mu_{t+1}\leftarrow\tfrac12(\mu_{t+1}+\mu_{t+1}^{\top}),
\label{eq:app-chemco-mu-update}
\end{equation}
\begin{equation}
\lambda_{i,t+1}^{+}
=
\lambda_{i,t}^{+}
+
\eta_+^{(t)}
\bigl[\operatorname{val}_{i,t+1}-c_i\bigr]_+,
\qquad
\lambda_{i,t+1}^{-}
=
\lambda_{i,t}^{-}
+
\eta_-^{(t)}
\bigl[d_{\min}-\deg_{i,t+1}\bigr]_+.
\label{eq:app-chemco-lambda-update}
\end{equation}
Here
$\eta_\mu^{(t)}=\eta_\mu\gamma_\mu^t$,
$\eta_+^{(t)}=\eta_+\gamma_+^t$, and
$\eta_-^{(t)}=\eta_-\gamma_-^t$ are exponentially decayed dual step sizes.

\paragraph{Nonnegativity invariant.}
The warm-start initialization is nonnegative, and every update adds a
nonnegative positive-residual term. Therefore
$\mu_{ij,t}\ge 0$, $\lambda_{i,t}^{+}\ge 0$, and
$\lambda_{i,t}^{-}\ge 0$ for all iterations $t$.

\subsection{Complete algorithm}
\label{app:chemco-algorithm}

Algorithm~\ref{alg:chemco} summarizes the implemented ChemCO core. It contains
the continuous logit-space unroll and positive-residual dual accumulation.

\begin{algorithm}[t]
\caption{ChemCO: implemented unrolled logit-space correction}
\label{alg:chemco}
\begin{algorithmic}[1]
\REQUIRE Raw bond logits $U\in\mathbb{R}^{N\times N\times(1+K)}$;
node mask $m$; TopoBridge support $\widetilde A$; chemical masks
$\{\Omega^{(k)}\}_{k=1}^{K}$; valence caps $\{c_i\}$; minimum degree
$d_{\min}$; iterations $T_{\mathrm{CO}}$.
\ENSURE Projected bond distribution
$P\in[0,1]^{N\times N\times(1+K)}$.

\STATE Build valid-pair mask $M$ from $m$ and effective support
$M_e \gets M\odot\widetilde A$.
\STATE Symmetrize logits:
$U\gets\tfrac12(U+U^\top)$.
\STATE Compute relative utility:
$\bar U^{(k)}\gets U^{(k)}-U^{(0)}$, $k=1,\ldots,K$.
\STATE Initialize logits:
$\Phi_0^{(k)}\gets\bar U^{(k)}$.
\STATE Set
$a_0^{(k)}\gets\sigma(\Phi_0^{(k)})\odot M_e\odot\Omega^{(k)}$.
\STATE Compute
$Y_0^{(k)}\gets T_k(a_0)$, then $s_0$, $\operatorname{val}_0$, and
$\deg_0$.
\STATE Warm-start duals:
$\mu_0\gets[s_0-1]_+$,
$\lambda_0^+\gets[\operatorname{val}_0-c]_+$, and
$\lambda_0^-\gets[d_{\min}-\deg_0]_+$.

\FOR{$t=0,\ldots,T_{\mathrm{CO}}-1$}
    \STATE Compute gates
    $\omega_t^{\mathrm{pair}}$, $\omega_t^{\mathrm{val}}$,
    $\omega_t^{\mathrm{deg}}$
    using Eq.~\eqref{eq:app-chemco-gates}.
    \STATE Compute row-centric direction
    $\widehat G_t^{(k)}$
    using Eq.~\eqref{eq:app-chemco-row-direction}.
    \STATE Symmetrize:
    $\widetilde G_t^{(k)}
    \gets
    \widehat G_t^{(k)}+(\widehat G_t^{(k)})^\top$.
    \STATE Update logits:
    $\Phi_{t+1}^{(k)}
    \gets
    \Phi_t^{(k)}
    +
    \alpha\gamma_\alpha^t
    \widetilde G_t^{(k)}\odot\Omega^{(k)}$.
    \STATE Recompute
    $a_{t+1}^{(k)}
    \gets
    \sigma(\Phi_{t+1}^{(k)})\odot M_e\odot\Omega^{(k)}$,
    $Y_{t+1}^{(k)}\gets T_k(a_{t+1})$, and refresh
    $s_{t+1}$, $\operatorname{val}_{t+1}$, $\deg_{t+1}$.
    \STATE Update $\mu_{t+1}$ using Eq.~\eqref{eq:app-chemco-mu-update}.
    \STATE Update $\lambda_{t+1}^{+}$ and $\lambda_{t+1}^{-}$
    using Eq.~\eqref{eq:app-chemco-lambda-update}.
\ENDFOR

\STATE Assemble soft distribution:
$P^{(k)}\gets Y_{T_{\mathrm{CO}}}^{(k)}$ for $k=1,\ldots,K$.
\STATE Set no-bond mass:
$P^{(0)}\gets
\operatorname{clamp}(1-\sum_{k=1}^{K}P^{(k)},0,1)$.
\STATE For invalid pairs, set $P_{ij}=[1,0,\ldots,0]$.
\STATE \textbf{return} $P$.
\end{algorithmic}
\end{algorithm}

\subsection{Full bond distribution and hard decoding}
\label{app:chempo-hard}

After $T_{\text{CO}}$ iterations, ChemCO returns the optimized assignment
$\widehat Y^{(k)}\!=\!T_k(\widehat\Phi)$. The full soft bond distribution
over $\{0,1,\ldots,K\}$ assigns no-bond mass as the clamped residual:
\begin{equation}
    P_{ij}^{(0)}
    =
    \operatorname{clamp}\!\Bigl(1-\sum_{k=1}^{K}\widehat Y_{ij}^{(k)},\;0,\;1\Bigr),
    \qquad
    P_{ij}^{(k)}=\widehat Y_{ij}^{(k)},
    \quad k=1,\ldots,K.
    \label{eq:chempo-full-prob-app}
\end{equation}
The clamping ensures numerical stability when the pair exclusivity
constraint $s_{ij}\!\le\!1$ is not yet fully satisfied after a finite unroll.

The discrete bond type is selected by
\begin{equation}
    \widehat B_{ij}
    =
    \operatorname*{arg\,max}_{k\in\{0,1,\ldots,K\}} P_{ij}^{(k)}.
    \label{eq:chempo-hard-app}
\end{equation}
During training, gradients are propagated through the full unrolled soft
projection $\Phi_0\!\mapsto\!\Phi_{T_{\mathrm{CO}}}\!\mapsto\!\widehat Y$.
Hard decoding (Eq.~\ref{eq:chempo-hard-app}) is used only for discrete
molecule construction or straight-through variants, and is not on the
gradient path.

\subsection{Discrete valence refinement}
\label{app:chempo-valence-repair}

ChemCO's continuous solver enforces one-sided valence inequalities
($\operatorname{val}_i\!\le\!c_i$), which is the natural formulation in
continuous space. However, the $\arg\max$ hard decoding
(Eq.~\ref{eq:chempo-hard-app}) can introduce discrete valence violations that
the continuous solver cannot resolve, because its soft assignment landscape may
contain local minima where the inequality is satisfied but the rounded solution
is infeasible~\citep{vazirani2001approximation}. To escape such local minima, ChemCO incorporates a discrete
valence refinement step as part of its optimization loop.

\paragraph{Hybrid continuous--discrete loop.}
As shown in Algorithm~\ref{alg:chemco}, after each continuous primal--dual
unroll produces $\widehat B$ via hard decoding, ChemCO checks whether any
atom~$i$ has discrete valence $v_i=\sum_j o_{\widehat B_{ij}}$ exceeding its
capacity~$c_i$. If violations exist, a greedy discrete repair is applied to
$\widehat B$, and the repaired bond graph is re-injected as initialization for
a subsequent continuous unroll. This continuous$\to$discrete$\to$continuous
cycle repeats until no discrete violations remain or the maximum number of
outer iterations $T_{\mathrm{outer}}$ is reached. During training, gradients
flow only through the final continuous unroll; the discrete repair steps are
non-differentiable.

\paragraph{Greedy bond-order reduction.}
The discrete repair operates on the rounded bond graph $\widehat B$ by
iteratively resolving overvalent atoms:
\begin{enumerate}[leftmargin=1.5em,topsep=3pt,itemsep=2pt]
    \item \textbf{Scan.} Identify all atoms with $v_i>c_i$ and sort by
          decreasing violation $v_i-c_i$.
    \item \textbf{Prioritize.} For the most-violated atom, rank its incident
          bonds by a priority score that favors: (a)~bonds to other overvalent
          atoms (resolving mutual violations), (b)~higher-order bonds
          (downgrading triple$\to$double or double$\to$single removes more
          excess valence per edit), while protecting terminal atoms
          (degree-$1$ neighbors are deprioritized).
    \item \textbf{Repair.} Apply the highest-priority feasible action: reduce
          bond order by one, or remove a single bond if the neighbor retains
          at least one other bond.
    \item \textbf{Iterate.} Repeat until no violations remain or a maximum
          iteration count is reached.
\end{enumerate}
The discrete repair acts as a structured perturbation that moves the solution
out of a continuous local minimum. In practice, the continuous solver already
produces near-feasible solutions, so the discrete phase typically requires
only a few bond-order edits per molecule.

\subsection{Bridge-edge protection during optimization}
\label{app:chempo-bridge-protection}

ChemCO optimizes bond-type assignments subject to valence and degree
constraints, which may drive the soft assignment mass on certain edges toward
zero---effectively removing them from the molecular graph. If such an edge
happens to be a \emph{bridge} (an edge whose removal disconnects the graph),
the connectivity established by TopoBridge would be destroyed.

\paragraph{Bridge-edge set.}
Let $\mathcal{B}\!\subseteq\!\mathcal{E}$ denote the set of
\emph{graph-theoretic} bridge edges in the current molecular graph, identified
by running Tarjan's bridge-finding algorithm on the TopoBridge output. Note
that these are distinct from the edges \emph{inserted} by TopoBridge's BFS
reconnection: an inserted edge may or may not be a graph-theoretic bridge,
and a bridge may be an original predicted edge rather than an inserted one.

\paragraph{Post-processing mask.}
After computing the bond distribution $P$ via Eq.~\eqref{eq:chempo-full-prob-app},
bridge-edge protection is applied as a post-processing step:
for each bridge edge $(i,j)\!\in\!\mathcal{B}$, the no-bond probability is
set to zero and the remaining bond-type probabilities are renormalized:
\begin{equation}
    P_{ij}^{(0)}
    \;\leftarrow\; 0,
    \qquad
    P_{ij}^{(k)}
    \;\leftarrow\;
    \frac{\widehat Y_{ij}^{(k)}}
         {\sum_{k'=1}^{K}\widehat Y_{ij}^{(k')}+\epsilon},
    \qquad
    (i,j)\in\mathcal{B},\;\;k=1,\ldots,K.
    \label{eq:chempo-bridge-protect-app}
\end{equation}
This guarantees $P_{ij}^{(0)}+\sum_k P_{ij}^{(k)}=1$ for bridge edges and
ensures that they always retain a real bond type ($k\!\ge\!1$) throughout
ChemCO's optimization. For non-bridge edges the distribution is
unchanged.

This mechanism complements the valence and degree constraints: while ChemCO
freely adjusts bond types and removes redundant edges to satisfy chemical
constraints, it is prevented from disrupting the connected topology guaranteed
by TopoBridge. The bridge mask thus propagates Stage~2's connectivity guarantee
through Stage~3's bond-type optimization, ensuring the final discrete molecular
graph remains connected.

\paragraph{Relationship to adjacency support.}
The TopoBridge adjacency support $\widetilde A$ defines the set of
\emph{candidate} edges on which ChemCO may place bonds. It does not
guarantee that every candidate edge will carry a real bond in the final
output: ChemCO may assign $P_{ij}^{(0)}\!\approx\!1$ (no bond) to non-bridge
candidate edges when doing so satisfies valence constraints. Only bridge
edges in $\mathcal{B}$ are protected from removal.

\subsection{Notation correspondence}
\label{app:chempo-notation}

For clarity we summarize the correspondence between the compact main-text
notation and the iteration-explicit appendix notation:

\begin{center}
\small
\begin{tabular}{lll}
\toprule
\textbf{Main text} & \textbf{Appendix} & \textbf{Description} \\
\midrule
$s_{ij}$ & $s_{ij,t}$ & pair-level total assignment \\
$\operatorname{val}_i$ & $\operatorname{val}_{i,t}$ & soft valence at atom $i$ \\
$\deg_i$ & $\deg_{i,t}$ & soft degree at atom $i$ \\
$\lambda_i$ (valence) & $\lambda_{i,t}^+$ & valence cap dual variable \\
$\nu_i$ (degree) & $\lambda_{i,t}^-$ & minimum-degree dual variable \\
$\mu_{ij}$ & $\mu_{ij,t}$ & pair exclusivity dual variable \\
$\eta_\Phi$ & $\alpha\gamma_\alpha^t$ & decayed primal step size \\
$\eta_\mu$, $\eta_+$, $\eta_-$ & $\eta_\bullet\gamma_\bullet^t$ & decayed dual step sizes \\
\bottomrule
\end{tabular}
\end{center}
The main text omits the iteration subscript~$t$ and the step-size decay for
conciseness, and uses $\nu_i$ in place of $\lambda_i^-$ to distinguish it from
the valence dual; the two notations are interchangeable.

\newpage
\section*{NeurIPS Paper Checklist}

\begin{enumerate}

\item {\bf Claims}
    \item[] Question: Do the main claims made in the abstract and introduction accurately reflect the paper's contributions and scope?
    \item[] Answer: \answerYes{}
    \item[] Justification: The abstract and introduction state the main contributions: dark-area diagnosis, TopVAE with TopoBridge/ChemCO/AGCL, and improved generation and scaffold-inpainting performance. 

\item {\bf Limitations}
    \item[] Question: Does the paper discuss the limitations of the work performed by the authors?
    \item[] Answer: \answerYes{}
    \item[] Justification: Paper discusses limitations that future extensions to physical constraints and conditional molecular generation.

\item {\bf Theory assumptions and proofs}
    \item[] Question: For each theoretical result, does the paper provide the full set of assumptions and a complete (and correct) proof?
    \item[] Answer: \answerNA{}
    \item[] Justification: The paper does not present standalone theorem/lemma statements.

    \item {\bf Experimental result reproducibility}
    \item[] Question: Does the paper fully disclose all the information needed to reproduce the main experimental results of the paper to the extent that it affects the main claims and/or conclusions of the paper (regardless of whether the code and data are provided or not)?
    \item[] Answer: \answerYes{}
    \item[] Justification: The paper discloses the datasets, metrics, perturbation protocol, generation and inpainting settings, model architecture, training hyperparameters, loss terms, diffusion setup, and ChemCO update rules. These details provide a reproducible specification of the main experimental pipeline independent of code release.

\item {\bf Open access to data and code}
    \item[] Question: Does the paper provide open access to the data and code, with sufficient instructions to faithfully reproduce the main experimental results, as described in supplemental material?
    \item[] Answer: \answerNo{}
    \item[] Justification: The experiments use public datasets and the paper provides detailed implementation and training information, but the current submission does not provide an anonymized code release, execution commands, or a documented artifact for reproducing the main results. The authors plan to release the code and related artifacts after paper acceptance.

\item {\bf Experimental setting/details}
    \item[] Question: Does the paper specify all the training and test details (e.g., data splits, hyperparameters, how they were chosen, type of optimizer) necessary to understand the results?
    \item[] Answer: \answerYes{}
    \item[] Justification: Method part defines the datasets and evaluation metrics, while Appendix specifies preprocessing, model architecture, optimizer, learning-rate schedule, epochs, batch sizes, loss weights, warmup schedules, and diffusion training/sampling details.

\item {\bf Experiment statistical significance}
    \item[] Question: Does the paper report error bars suitably and correctly defined or other appropriate information about the statistical significance of the experiments?
    \item[] Answer: \answerNo{}
    \item[] Justification: The paper reports results over fixed evaluation sample sizes, such as 10,000 generated molecules, but does not report error bars, confidence intervals, statistical significance tests, or variability across independent training runs. The reported results should therefore be interpreted as point estimates under the stated experimental settings.

\item {\bf Experiments compute resources}
    \item[] Question: For each experiment, does the paper provide sufficient information on the computer resources (type of compute workers, memory, time of execution) needed to reproduce the experiments?
    \item[] Answer: \answerYes{}
    \item[] Justification: Appendix reports the compute setup and profiling results, including NVIDIA H100 NVL GPUs, batch sizes, per-component timing, peak memory, forward/backward cost, DDP step-time comparisons, and scaling curves. The full set of reported experiments used approximately 384 H100 GPU-hours in total.
    
\item {\bf Code of ethics}
    \item[] Question: Does the research conducted in the paper conform, in every respect, with the NeurIPS Code of Ethics \url{https://neurips.cc/public/EthicsGuidelines}?
    \item[] Answer: \answerYes{}
    \item[] Justification: The research uses public molecular datasets and standard machine-learning experimentation without human subjects or private data. The authors have reviewed the NeurIPS Code of Ethics and believe the work conforms to it.

\item {\bf Broader impacts}
    \item[] Question: Does the paper discuss both potential positive societal impacts and negative societal impacts of the work performed?
    \item[] Answer: \answerNo{}
    \item[] Justification:  The current manuscript focuses on methodological and empirical contributions for 3D molecular generation and does not include a dedicated discussion of both positive and negative societal impacts. Potential impacts include beneficial use in molecular design as well as possible dual-use concerns if generative molecular models are used to propose harmful compounds.
    
\item {\bf Safeguards}
    \item[] Question: Does the paper describe safeguards that have been put in place for responsible release of data or models that have a high risk for misuse (e.g., pre-trained language models, image generators, or scraped datasets)?
    \item[] Answer: \answerNA{}
    \item[] Justification: The current submission does not release a scraped dataset, a deployed generation service, or a high-risk pretrained language/image model. The planned post-acceptance release is research code and benchmark-trained molecular generation artifacts; if released checkpoints are included, they will be accompanied by usage guidelines in the repository.

\item {\bf Licenses for existing assets}
    \item[] Question: Are the creators or original owners of assets (e.g., code, data, models), used in the paper, properly credited and are the license and terms of use explicitly mentioned and properly respected?
    \item[] Answer: \answerNo{}
    \item[] Justification: The paper cites the original datasets, baselines, and software resources used in the experiments, including QM9, GEOM-Drugs, and prior molecular generation methods. However, the current submission does not explicitly enumerate the licenses, versions, or terms of use for all existing assets; these will be listed in the post-acceptance repository.

\item {\bf New assets}
    \item[] Question: Are new assets introduced in the paper well documented and is the documentation provided alongside the assets?
    \item[] Answer: \answerNA{}
    \item[] Justification: The current submission does not release new standalone assets such as a dataset, benchmark, codebase, or model checkpoint. The authors plan to release code and related artifacts after acceptance, with documentation, licensing information, and reproduction instructions provided alongside the release.

\item {\bf Crowdsourcing and research with human subjects}
    \item[] Question: For crowdsourcing experiments and research with human subjects, does the paper include the full text of instructions given to participants and screenshots, if applicable, as well as details about compensation (if any)? 
    \item[] Answer: \answerNA{}
    \item[] Justification:  The work does not involve crowdsourcing, human-subject experiments, user studies, human annotation, or human evaluation.

\item {\bf Institutional review board (IRB) approvals or equivalent for research with human subjects}
    \item[] Question: Does the paper describe potential risks incurred by study participants, whether such risks were disclosed to the subjects, and whether Institutional Review Board (IRB) approvals (or an equivalent approval/review based on the requirements of your country or institution) were obtained?
    \item[] Answer: \answerNA{}
    \item[] Justification: The work does not involve crowdsourcing or human-subject research, so IRB or equivalent review is not applicable.

\item {\bf Declaration of LLM usage}
    \item[] Question: Does the paper describe the usage of LLMs if it is an important, original, or non-standard component of the core methods in this research? Note that if the LLM is used only for writing, editing, or formatting purposes and does \emph{not} impact the core methodology, scientific rigor, or originality of the research, declaration is not required.
    \item[] Answer: \answerNA{}.
    \item[] Justification: LLMs were used only for writing, editing, or formatting assistance and did not affect the core methodology, experiments, scientific rigor, or originality of the research. The core method development does not involve LLMs as an important, original, or non-standard component.

\end{enumerate}

\end{document}